\documentclass[11pt]{article}

\usepackage[utf8]{inputenc}   
\usepackage[T1]{fontenc}
\usepackage[margin=1in]{geometry}
\usepackage{amsmath,amssymb}
\usepackage{subcaption}
\usepackage{bm}
\usepackage{xcolor}
\usepackage{enumitem}
\usepackage{graphicx}
\usepackage{tikz}
\usetikzlibrary{spy, arrows.meta, positioning, matrix, calc}

\usepackage{booktabs}
\usepackage{multirow}
\usepackage{array}

\definecolor{darkpurple}{HTML}{9013FE}
\definecolor{vividorange}{HTML}{ED702F}
\definecolor{richblue}{HTML}{4A90E2}
\definecolor{purple1}{HTML}{4B0082}
\definecolor{purple2}{HTML}{9013FE}
\definecolor{purple3}{HTML}{5E239D}
\definecolor{purple4}{HTML}{B580FF}
\definecolor{purple5}{HTML}{E5D5FF}

\definecolor{blue1}{HTML}{0033A0}
\definecolor{blue2}{HTML}{1E90FF}
\definecolor{blue3}{HTML}{6699CC}
\definecolor{blue4}{HTML}{A4C8E1}
\definecolor{blue5}{HTML}{D6EAF8}
\definecolor{blue5_2}{HTML}{4A90E2}

\definecolor{orange1}{HTML}{8B2500}
\definecolor{orange2}{HTML}{FF7F0E}
\definecolor{orange3}{HTML}{FDB863}
\definecolor{orange4}{HTML}{FDD9B5}
\definecolor{orange5}{HTML}{FFF1E0}

\definecolor{lightgray}{HTML}{D9D9D9}
\definecolor{darkgray}{HTML}{595959}
\definecolor{paleblue}{HTML}{A7C7E7}

\usepackage{nicematrix}
\usepackage[hidelinks]{hyperref}
\usepackage{titletoc}
\usepackage{float}
\usepackage[ruled,vlined]{algorithm2e}

\usepackage[
  style=numeric-comp,
  minbibnames=3,   
  maxbibnames=4,   
  maxcitenames=1,  
  mincitenames=1,  
  giveninits=true,
  doi=true,
  url=true,          
  sorting=none,
]{biblatex}

\AtEveryBibitem{%
  \clearfield{issn}%
  \clearfield{month}%
  \iffieldundef{doi}{}{\clearfield{url}\clearfield{urlyear}}%
  \iffieldundef{eprint}{}{\clearfield{url}\clearfield{urlyear}}%
}
\newcommand{\cornerlabelimgshift}[5][]{%
  \begin{tikzpicture}
    \node[inner sep=0] (img) {\includegraphics[#1]{#2}};
    \node[
      anchor=#4,
      font=\bfseries\normalsize,
      fill=white,
      fill opacity=0,
      text opacity=1,
      inner sep=1pt,
      #5
    ] at (img.#4) {#3};
  \end{tikzpicture}%
}

\newlength{\schematicwidth}
\newlength{\twocolwidthDiag}
\newlength{\boxheaderawidthtwoD}
\newlength{\boxheaderbwidthtwoD}
\newlength{\leftfigwidth}
\newlength{\rightfigwidth}
\newlength{\rightpanelplotwidth}
\newlength{\rightpanelheaderwidth}
\newlength{\rightpaneldrop}
\newlength{\rightpaneldropf}
\newlength{\rightheaderraise}
\newlength{\rowheaderwidth}
\newlength{\panelwidthLHS}
\newlength{\panelwidthRHS}
\newlength{\lhscolsep}
\newlength{\rhscolsep}
\newlength{\lhsrowsep}
\newlength{\rhsrowsep}
\newlength{\rowheaderraiseA}
\newlength{\rowheaderraiseB}
\newlength{\rowheaderraiseC}
\newlength{\boxheaderleftwidth}
\newlength{\boxheaderrightwidth}
\newlength{\fourcolwidth}
\newlength{\boxheadertopwidth}
\newlength{\boxheaderbottomwidth}
\newlength{\panelwidthtwoD}
\newlength{\boxheaderleftwidthtwoD}
\newlength{\boxheaderrightwidthtwoD}
\newlength{\leftfigwidthdepth}
\newlength{\rightfigwidthdepth}
\newlength{\rightpanelplotwidthdepth}
\newlength{\rightpanelheaderwidthdepth}
\newlength{\rightpaneldropdepth}
\newlength{\rightpaneldropdepthb}
\newlength{\rightheaderraisedepth}
\newlength{\groupboxwidth}

\title{Hyperparameter selection for equation learning with biologically-informed neural networks}
\author{
  \begin{tabular}{c}
    William Lavery\textsuperscript{1}, 
    Jodie A. Cochrane\textsuperscript{1}, 
    John T. Nardini\textsuperscript{2}, 
    Sara Hamis\textsuperscript{1} \\[1ex]
    \small \textsuperscript{1}Department of Information Technology, Uppsala University, Sweden \\
    \small \textsuperscript{2}Department of Mathematics and Statistics, The College of New Jersey, USA \\[1ex]
    \small \texttt{william.lavery@it.uu.se, jodie.cochrane@it.uu.se, nardinij@tcnj.edu} \\
    \small \texttt{sara.hamis@it.uu.se}
  \end{tabular}
}
\date{}

\begin{document}

\maketitle
\section*{Abstract}
Biologically-informed neural networks (BINNs) have emerged as a flexible subclass of physics-informed neural networks (PINNs) for learning terms in partial differential equations from data. BINNs are particularly suited for biological systems, where the governing equations are highly nonlinear and only partially known \textit{a priori}, and where data observations are often sparse, noisy, and incomplete. 
However, applying BINNs effectively in practice depends critically on hyperparameter selection, which remains a central challenge in equation-learning frameworks. 
Hyperparameters are often chosen heuristically and only cursorily documented, which limits the reproducibility of results and the transferability of methods. We present a diagnostic workflow for hyperparameter selection that can be used when the ground-truth equations are not known. 
The workflow is guided by three main questions:
(1) Are the benefits of greater network capacity worth the cost? (2) Do more training epochs keep reducing the validation loss? (3) Do the learned terms stop changing as network capacity and training increase? 
We apply our workflow to synthetic systems of varying complexity with known ground truth, spanning diffusion and growth right-hand side terms and data ranging from $1\text{D}{+}t$ to $2\text{D}{+}t$. We demonstrate that the validation loss generally follows the true error in the learned terms and distil practical rules of thumb for selecting hyperparameters in the BINN architecture. By providing a structured workflow, practical guidelines, and suggested starting values for hyperparameter selection, this work lowers the barrier to BINN-based equation learning.

\clearpage
\newpage

\section{Introduction}

\subsection{Equation learning with neural networks}

Biological dynamics are often modelled with differential equations whose
functional forms are only partially known from theory. Equation learning (EQL)
seeks to infer such governing equations directly from data \cite{campsvalls2023discoveringcausalrelationsequations}. Prominent EQL approaches include symbolic-regression methods that search over analytic expressions to fit the data \cite{Schmidt2009,Cranmer2020,Cranmer2023}, sparse-regression frameworks such as Sparse Identification of Nonlinear Dynamics (SINDy) and its partial differential equation (PDE) extensions that select active terms from a candidate library \cite{Brunton2016,Rudy2017}, and physics-informed neural networks (PINNs) \cite{Raissi2019}. In their inverse form, PINNs represent the solution of a differential equation with a neural network and infer the equation's unknown parameters or terms directly from data.
Building on the PINN approach, biologically-informed neural networks (BINNs) learn the unknown coefficient functions of the right-hand side (RHS) terms of a differential equation while retaining its known operator template \cite{Lagergren2020}. This combination of known structure and learned coefficient functions makes BINNs well suited to mechanistic learning in biology: the structure anchors the model when data are sparse, noisy, and only partially observed, while the networks supply the coefficient functions that theory leaves unspecified \cite{Baker2018}.

BINNs were introduced by \textcite{Lagergren2020} to model cell migration as a
$1\text{D}{+}t$ reaction--diffusion process. We state the governing equation here in
the general $d$-dimensional form,
\begin{align}
    \frac{\partial u(\mathbf{x},t)}{\partial t}
    =
    \nabla \cdot \left[D(u(\mathbf{x},t)) \nabla u(\mathbf{x},t)\right]
    + G(u(\mathbf{x},t))\,u(\mathbf{x},t),
    \quad \mathbf{x} \in \Omega \subset \mathbb{R}^d,\ t \in [0,t_f],
\label{eq:reaction_diff_eq_2d}
\end{align}
where $u(\mathbf{x},t)$ is the density at position $\mathbf{x}$ and time $t$, $\Omega$ is a bounded spatial domain of dimension $d$, and $t_f$ is the final time. The EQL problem then becomes learning the density $u(\mathbf{x},t)$ on the left-hand side of Eq.~\eqref{eq:reaction_diff_eq_2d} with a state network, and the diffusion and growth coefficient functions $D(u(\mathbf{x},t))$ and $G(u(\mathbf{x},t))$ with separate RHS networks. All networks are fully connected feedforward neural networks, also known as multilayer perceptrons (MLPs), interconnected within the BINN.
Since their introduction, BINNs have been extended, for example to jointly learn observation noise~\cite{crossley_likelihood-based_2026} 
and to $2\text{D}{+}t$ systems with symbolic-regression equation discovery~\cite{lavery_physics-informed_2026}, 
and applied to tasks such as forecasting stochastic agent-based model data~\cite{nardini2024forecasting}.

\subsection{Hyperparameter selection}
Hyperparameter selection is a challenge faced by every practitioner of neural-network-based EQL, and the settings chosen 
strongly affect both the accuracy of the recovered functions and the cost of training. State networks in PINNs are notoriously sensitive to architectural and training choices, motivating a body of work on principled hyperparameter and architecture selection that ranges from tuning dominant hyperparameters individually \cite{wang_auto-pinn_2023}, to joint Bayesian optimisation \cite{escapil-inchauspe_hyper-parameter_2023}, to gradient-based and evolutionary architecture-search methods \cite{wang_nas-pinn_2024,zhang_discovering_2024,wong_evolutionary_2026}, and to multi-objective treatments of the data--physics loss balance~\cite{rohrhofer_data_2023}. 
The networks for the unknown RHS terms, in contrast, have received less systematic study. Nonetheless, they are usually kept small, as they typically represent low-complexity functions 
\cite{Podina2023UPINN,philipps_current_2025,bolibar_universal_2023,kuwahara_predicting_2024,philipps_non-negative_2024}. 
Reflecting both the difficulty and importance of these choices, systematic hyperparameter-selection procedures have begun to emerge for particular subclasses of PINNs \cite{rosa_homo-pinn_2025, sharma_hyperparameter_2024, zhou_automated_2025}.

Compared to standard PINNs, BINNs face an additional challenge: because the state and RHS coefficient functions are learned simultaneously, their hyperparameters interact and cannot necessarily be tuned in isolation. 
Moreover, the coefficient functions being learned are never observed, so selection cannot fall back on held-out accuracy against a ground truth. On top of this, hyperparameter selection for BINNs remains largely understudied and underreported, which raises the barrier to using them and limits the reproducibility of results.

To address this, we present a practical, ground-truth-free workflow for hyperparameter selection in BINNs. The workflow focuses on selecting the capacities (primarily width and depth) of the state and RHS MLPs and the early-stopping patience, which controls how long training continues without sufficient improvement in the validation loss. We:  
\begin{itemize}[leftmargin=*, itemsep=0.2em]
\item Present a diagnostic workflow for BINN hyperparameter selection that requires no ground truth and fixes the hyperparameters in an order dictated by the architecture.
\item Evaluate the workflow on a suite of synthetic systems with known ground truth, spanning a range of diffusion and growth terms 
and extending to noisy, $2\text{D}{+}t$ data. 
\item Distil transferable rules of thumb for hyperparameter selection: first, use narrow MLPs for the RHS terms, with their sizes chosen according to the expected complexity of each term, and employ a substantially wider state MLP. Second, set the early-stopping patience from the validation loss and the consistency of the learned functions.
\end{itemize}

The BINN pipeline implementation for hyperparameter selection is available in an open-access \href{https://github.com/williamlavery/binn-hyperparam-selection}{GitHub} repository. 
The repository also includes user-friendly interactive notebooks for loading trained BINN models and reproducing all plots in this article. The pipeline can be readily adapted to new hyperparameters and other PDE-governed systems.

\section{Background}\label{sec:bg}

\subsection{The biologically-informed neural network architecture}
\begin{figure}[h!]
    \centering

    \begin{tikzpicture}
        \node[draw=lightgray, line width=1pt, rounded corners=5pt, inner sep=10pt] (topbox) {
            \begin{minipage}{0.95\textwidth}
                \centering
                \begin{minipage}[c]{0.365\textwidth}
                    \centering
                    \raisebox{0.1\height}{%
                        \includegraphics[width=\linewidth]{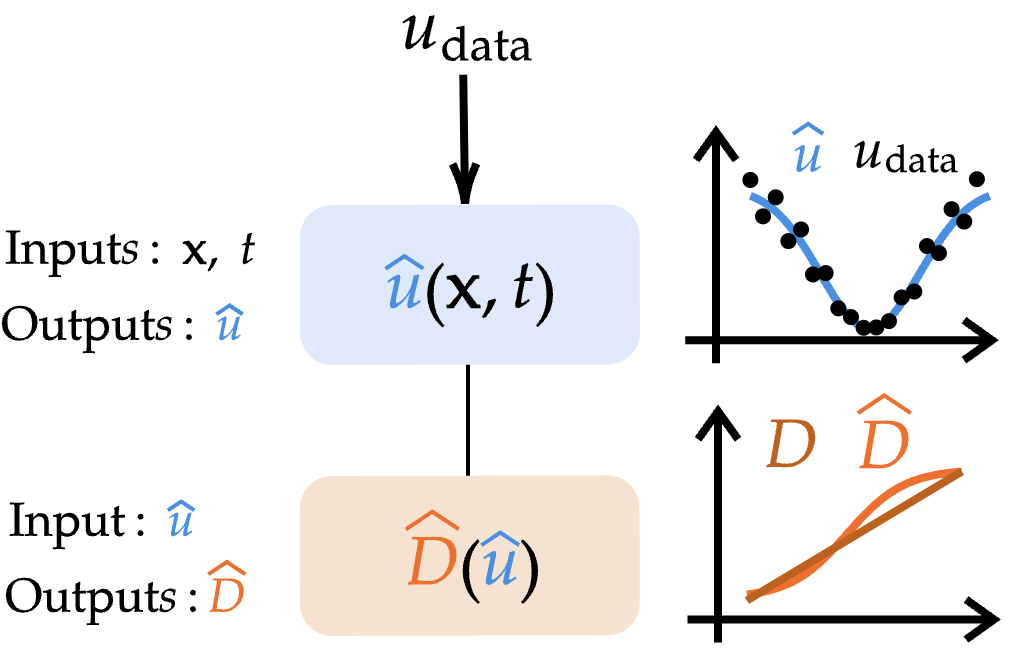}}
                \end{minipage}%
                \hspace{1cm}%
                \begin{minipage}[c]{0.4\textwidth}
                    \centering
                    \includegraphics[width=\linewidth]{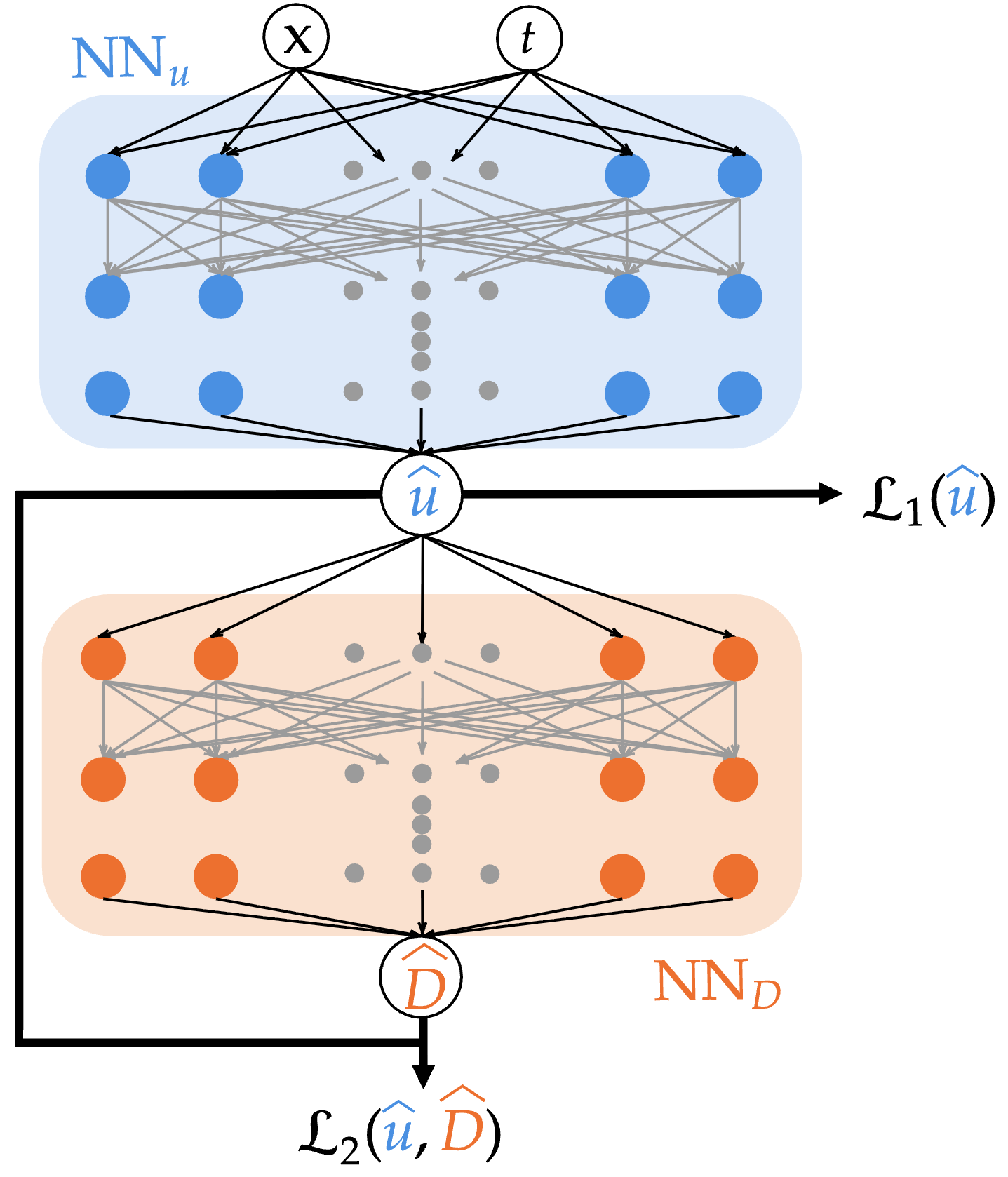}
                \end{minipage}
            \end{minipage}
        };
        \node[fill=white, text=black] at (topbox.north) {\textbf{BINN architecture for diffusion equations}};
    \end{tikzpicture}

    \vspace{0.25cm}

    \begin{tikzpicture}
        \node[draw=gray, line width=1pt, rounded corners=5pt, inner sep=10pt] (bottombox) {
            \begin{minipage}{0.95\textwidth}
                \centering
                \begin{minipage}[c]{0.4\textwidth}
                    \centering
                    \raisebox{0.1\height}{%
                        \includegraphics[width=\linewidth]{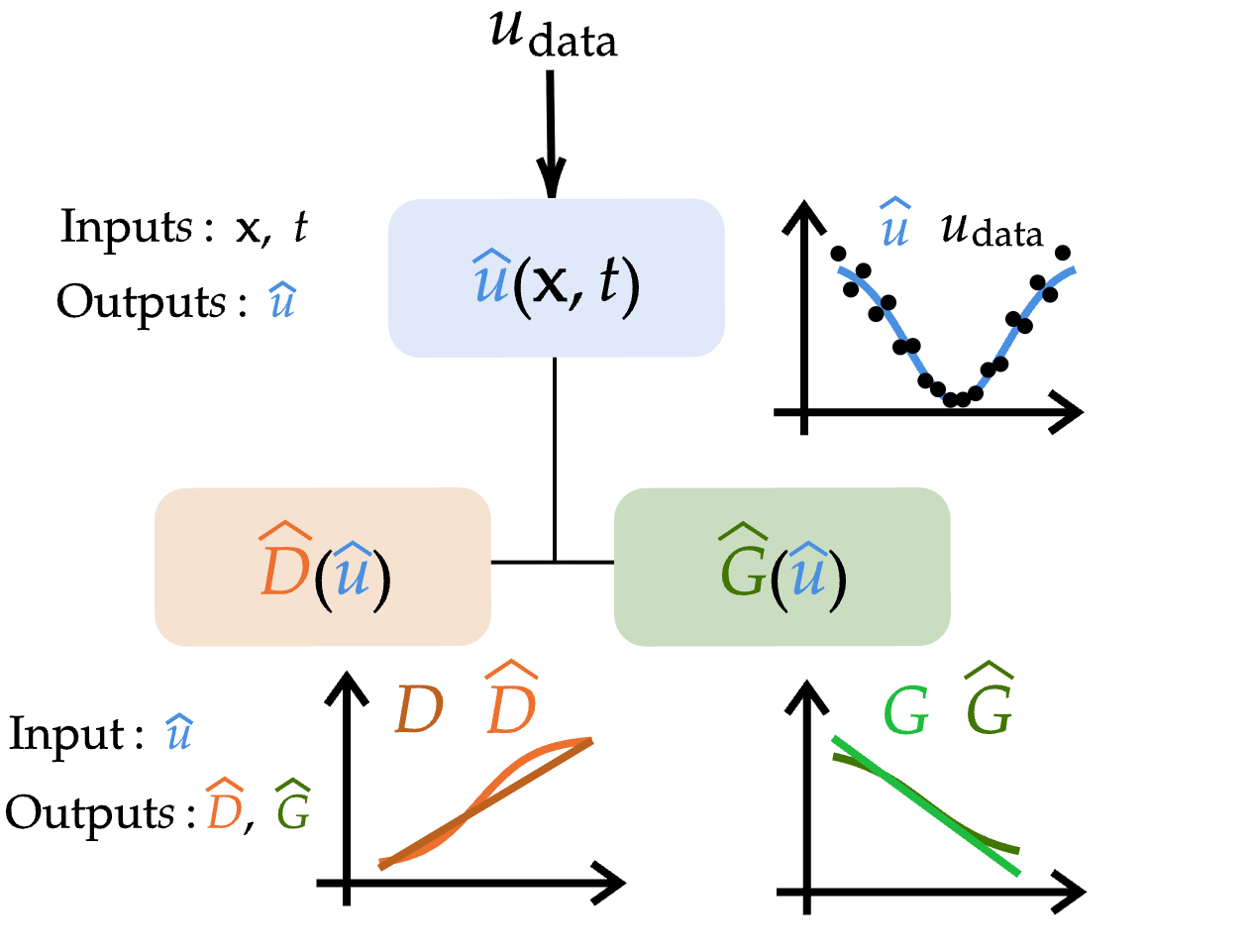}}
                \end{minipage}%
                \hfill
                \begin{minipage}[c]{0.58\textwidth}
                    \centering
                    \includegraphics[width=\linewidth]{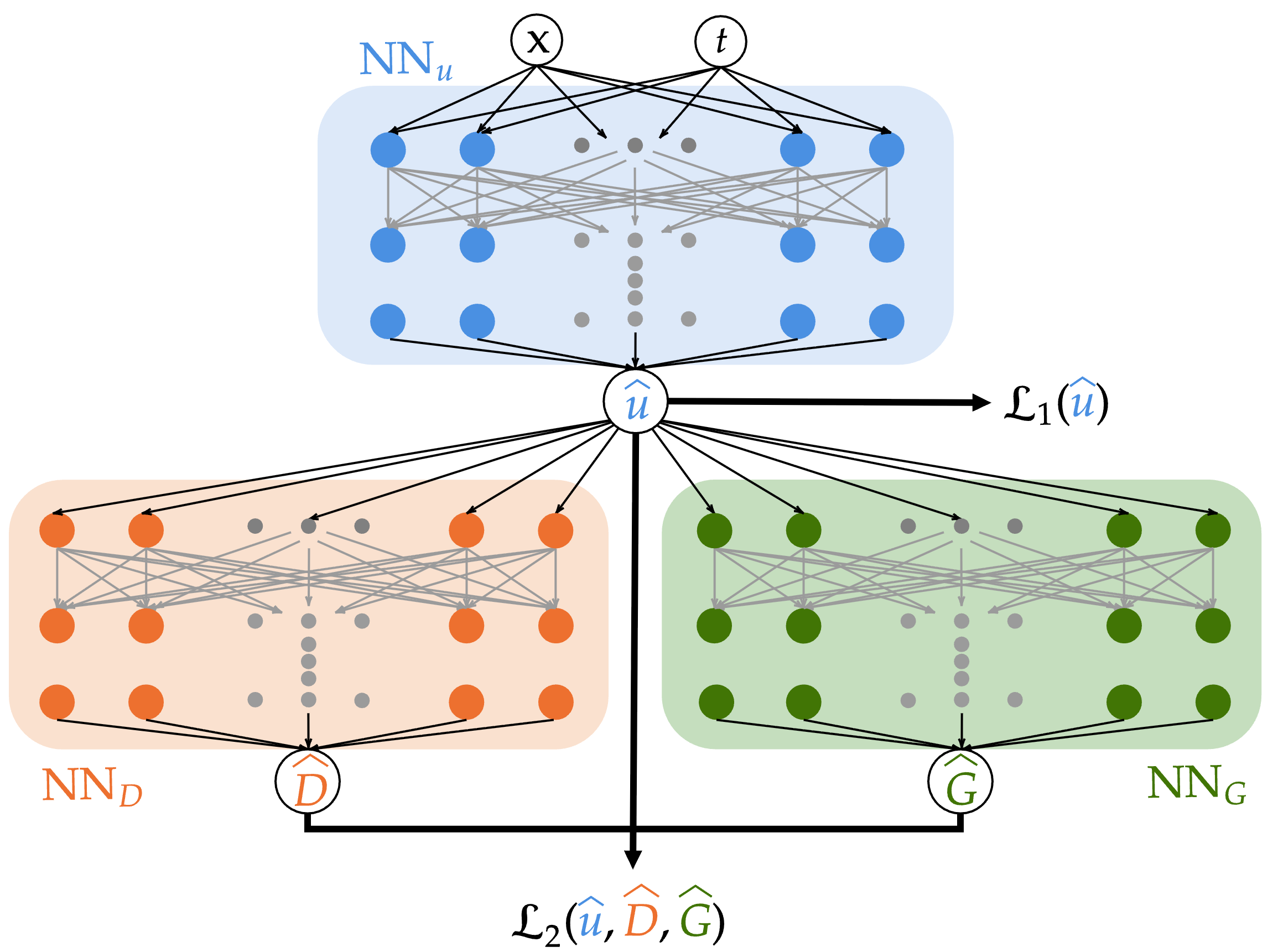}
                \end{minipage}
            \end{minipage}
        };
        \node[fill=white, text=black] at (bottombox.north) {\textbf{BINN architecture for reaction--diffusion equations}};
    \end{tikzpicture}

\caption{
\textbf{Overview of the biologically-informed neural network (BINN) architecture.} 
Schematic overviews of the network flow (\textit{left}) and the corresponding network architectures (\textit{right}) are shown for the diffusion equation and the reaction--diffusion equation. The ground truth is the diffusion function $D$ for the former, and the diffusion function $D$ together with the growth function $G$ for the latter. Discretised state data $u_{\text{data}}$ and their spatiotemporal coordinates $(\mathbf{x},t)$ are input to the state multilayer perceptron (MLP) $\mathrm{NN}_u$, which outputs the predicted state $\hat u$ (blue). 
The right-hand side MLPs are trained jointly to recover the underlying biological mechanisms: the diffusion MLP ($\mathrm{NN}_D$, orange) learns the diffusion function $\hat D$ and the growth MLP ($\mathrm{NN}_G$, green) learns the growth function $\hat G$.
Training minimises a total loss comprising a data fitting term $\mathcal{L}_1(\hat{u})$ and a physics-informed term $\mathcal{L}_2(\hat{u}, \hat{D}, \hat{G})$, the latter enforcing the governing partial differential equation and optionally incorporating user-specified biological constraints. 
}
    \label{fig:fig1}
\end{figure}

The BINN architecture \cite{Lagergren2020} is illustrated in Fig.~\ref{fig:fig1}. 
It comprises a state MLP $\mathrm{NN}_u$ and one or two RHS MLPs ($\mathrm{NN}_D$ for diffusion and $\mathrm{NN}_G$ for growth). The networks output learned surrogate functions $\hat{u}$, $\hat{D}$, and $\hat{G}$ for $u(\cdot)$, $D(\cdot)$, and $G(\cdot)$ in Eq.~\eqref{eq:reaction_diff_eq_2d}, respectively. 
Without loss of generality, the state $u(\mathbf{x},t)$ can be considered a density evolving in space and time. During training, this density is evaluated on a gridded domain $\mathcal{G}$ using bin-averaged values at grid points $(\mathbf{x},t)$. 
In the full three-dimensional case ($d=3$), the domain is discretised as
\[
(x_{1,i},x_{2,j},x_{3,k},t_s) \in \mathcal{G},
\qquad
(i,j,k,s) \in
\{1,\ldots,n_{x_1}\}
\times
\{1,\ldots,n_{x_2}\}
\times
\{1,\ldots,n_{x_3}\}
\times
\{1,\ldots,n_t\},
\]
where $n_{x_1},n_{x_2},n_{x_3}$ denote the fixed numbers of bins in each spatial direction, and $n_{t}$ the number of observed time points. This yields discretised density data arranged as a tensor
\[
u_{\text{data}} \in \mathbb{R}^{\: n_{x_1}\times n_{x_2}\times n_{x_3}\times n_t},
\]
which is then randomly split into a training set and a validation set.
Subsequently, these two sets are used to train and validate the density MLP 
$\mathrm{NN}_u(\,\cdot\,;\theta_u): \mathbb{R}^{d+1}\to\mathbb{R}$, with trainable parameters $\theta_u$,
which maps spatiotemporal coordinates $(\mathbf{x},t)$
to the predicted density surrogate $\hat{u}(\mathbf{x},t) := \mathrm{NN}_u(\mathbf{x},t;\theta_u)$ for $\mathbf{x} \in \mathbb{R}^d, \ t \in \mathbb{R}$. 
The surrogate $\hat{u}$ is in turn passed into 
$\mathrm{NN}_{D}(\,\cdot\,;\theta_D): \mathbb{R} \to \mathbb{R}$ and $\mathrm{NN}_{G}(\,\cdot\,;\theta_G): \mathbb{R} \to \mathbb{R}$,
mapping a density $u \in \mathbb{R}$ to the learned diffusion $\hat{D}(u) := \mathrm{NN}_{D}(u;\theta_D)$ and growth $\hat{G}(u) := \mathrm{NN}_{G}(u;\theta_G)$ coefficient functions.
All trainable MLP parameters $\theta_{u}$, $\theta_{D}$, and $\theta_{G}$ are jointly trained on the data while enforcing that the predictions adhere to the governing PDE and any user-defined biological constraints via corresponding loss terms. 

\subsection{Biologically-informed learning}
To infer $D(\cdot)$ and $G(\cdot)$ in Eq.~\eqref{eq:reaction_diff_eq_2d}, the BINN's MLPs are jointly trained by minimising the total objective function
\begin{equation}
\mathcal{L}_{\text{Total}}(\hat{u}, \hat{D}, \hat{G}) = 
\underbrace{\lambda_{\text{data}}\,\mathcal{L}_{\text{data}}(\hat{u})}_{\mathcal{L}_1} 
+ 
\underbrace{\lambda_{\text{PDE}}\,\mathcal{L}_{\text{PDE}}(\hat{u}, \hat{D}, \hat{G}) + \lambda_{\text{bio}}\,\mathcal{L}_{\text{bio}}(\hat{D}, \hat{G})}_{\mathcal{L}_2},
\label{eq:total_loss}
\end{equation}
with $\lambda_{\text{data}}, \lambda_{\text{PDE}}, \lambda_{\text{bio}} \geq 0$ denoting the weights applied to the least-squares data loss, the PDE loss, and the biological constraint loss, respectively. This is schematically shown in Fig.~\ref{fig:fig1} 
with $\mathcal{L}_1(\hat{u})=\lambda_{\text{data}}\mathcal{L}_{\text{data}}(\hat{u})$ and $\mathcal{L}_2(\hat{u}, \hat{D}, \hat{G})= \lambda_{\text{PDE}} \mathcal{L}_{\text{PDE}}(\hat{u}, \hat{D}, \hat{G}) + \lambda_{\text{bio}} \mathcal{L}_{\text{bio}}(\hat{D}, \hat{G})$. For compactness, each loss is written in terms of the surrogate predictions $\hat{u}$, $\hat{D}$, and $\hat{G}$ alone; the observed data $u_{\text{data}}$, the loss weights, and the biological bounds enter as fixed inputs. Optimisation is performed over the network parameters $\theta_u$, $\theta_D$, and $\theta_G$ that generate these surrogates. 
Weights $\lambda_{\text{data}}=\lambda_{\text{PDE}}=1$ follow \textcite{Lagergren2020}, and we set 
$\lambda_{\text{bio}}=1$, as discussed in Supplementary Material~\ref{sec:SM:pde_data_loss}--\ref{sec:SM:bio_loss}. The three loss terms in Eq.~\eqref{eq:total_loss} are detailed below for the
$2\text{D}{+}t$ case ($d=2$); their non-dimensionalised forms used in the
implementation are derived in Supplementary Material~\ref{sec:supmat:losses}.

\paragraph{Data loss.}
The data loss enforces agreement between the density predictions and the observed data.
Observations are assumed to be corrupted by additive Gaussian noise with non-constant variance, consistent with statistical error models used in related studies \cite{Banks2011, Lagergren2020,Lagergren2020_b}. 
Specifically, 
\begin{equation}
u_{\text{data}}^{i,j,s} = u(x_{1,i}, x_{2,j}, t_s) + \omega_{i,j,s} \, u^{\gamma}(x_{1,i}, x_{2,j}, t_s)\, \varepsilon_{i,j,s},
\label{eq:err_model}
\end{equation}
where $u(\cdot)$ denotes the unknown PDE solution for the density, $\varepsilon_{i,j,s} \overset{\text{iid}}{\sim} \mathcal{N}(0,1)$, $\gamma \in \mathbb{R}^+$ controls how the noise magnitude scales with density, and $\omega_{i,j,s} \in \mathbb{R}^+$ are additional variance scaling coefficients.

Under this model, the negative log-likelihood weights each squared data residual $(u_{\text{data}}-\hat{u})^2$ by the inverse of its variance, leading to a generalised least-squares (GLS) formulation in which these discrepancies are divided by $\hat{u}^\gamma$ (see Supplementary Material~\ref{sec:supmat:losses} for more information). Intuitively, this down-weights residuals in high-density regions and up-weights those in low-density regions. However, biological datasets frequently contain extended regions of very low population density, for which $\hat{u} \approx 0$. To ensure robust optimisation, we therefore drop the density-dependent weighting, which corresponds to setting $\gamma = 0$ in the loss, yielding a standard least-squares objective,
\begin{equation}
\mathcal{L}_{\text{data}}(\hat{u})
=
\frac{1}{n_{x_1}n_{x_2}n_t}
\sum_{i=1}^{n_{x_1}}
\sum_{j=1}^{n_{x_2}}
\sum_{s=1}^{n_t}
\left(
u_{\text{data}}^{i,j,s}
-
\hat{u}(x_{1,i},x_{2,j},t_s)
\right)^2.
\label{eq:data_loss}
\end{equation}
In Supplementary Material~\ref{sec:SM2}, the loss function in Eq.~\eqref{eq:data_loss} is shown to outperform the GLS loss used in \textcite{Lagergren2020}, even when the ground-truth noise scaling parameter $\gamma=1$ is used.

\begin{figure}[h!]
    \centering
    \begin{minipage}[c]{0.32\textwidth}
        \centering
        {\scriptsize\textbf{\hspace{0.6cm} Data points}}
        \vspace{0.2em}
        \includegraphics[width=\linewidth]{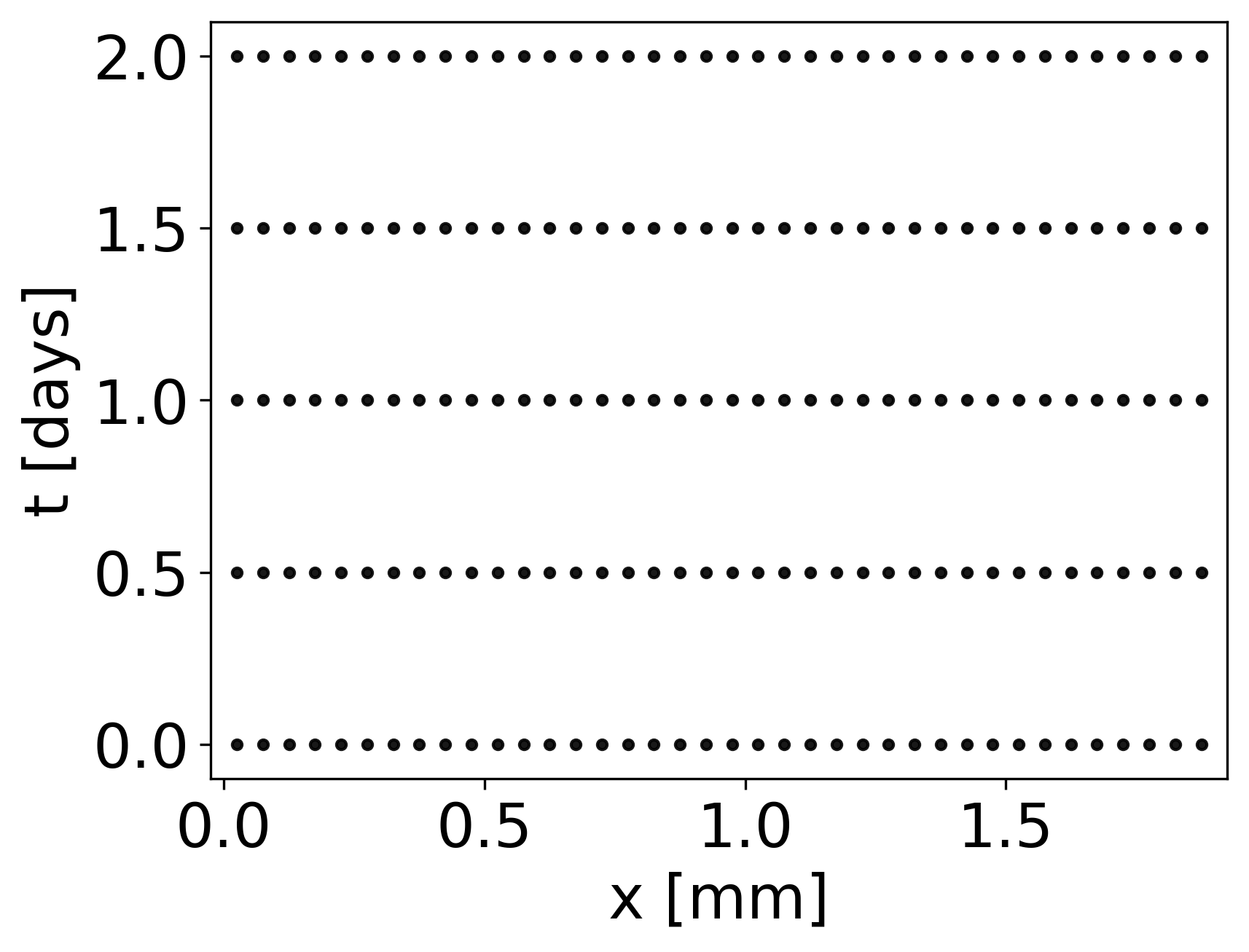}
    \end{minipage}
    \hfill
    \begin{minipage}[c]{0.62\textwidth}
        \centering
        \begin{tikzpicture}
            \node[
                draw=lightgray,
                line width=1pt,
                rounded corners=5pt,
                inner sep=0pt,
                text width=\linewidth,
                align=center
            ] (collocationbox) {
                \begin{minipage}{0.95\linewidth}
                    \centering
                    \vspace{0.45cm}
                    \begin{minipage}[c]{0.47\linewidth}
                        \centering
                        {\scriptsize\text{\hspace{0.3cm} Minibatch = 1}}
                        \vspace{0.2em}
                        \includegraphics[width=\linewidth]{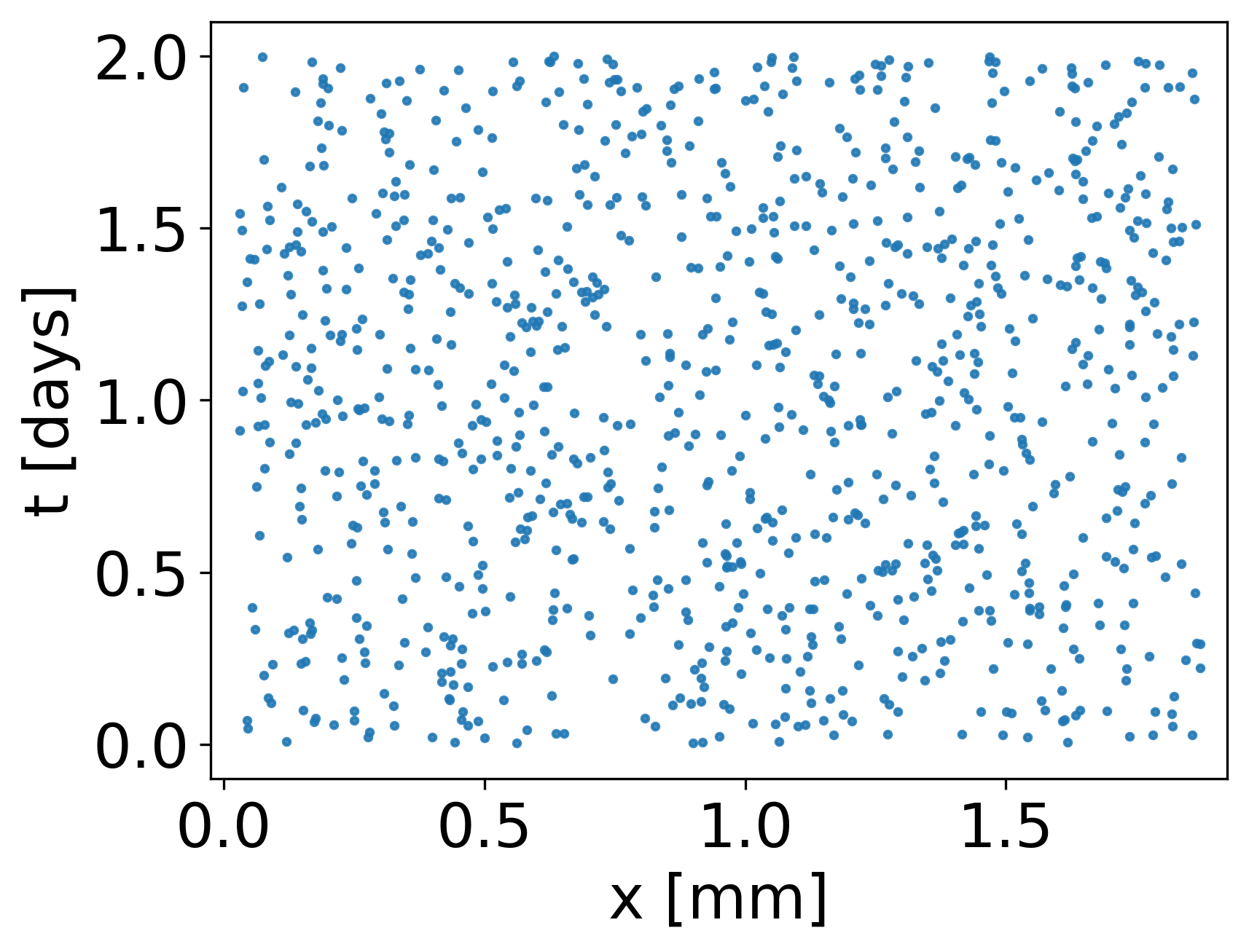}
                    \end{minipage}
                    \hfill
                    \begin{minipage}[c]{0.47\linewidth}
                        \centering
                        {\scriptsize\text{\hspace{0.3cm} Minibatch = 10}}
                        \vspace{0.2em}
                        \includegraphics[width=\linewidth]{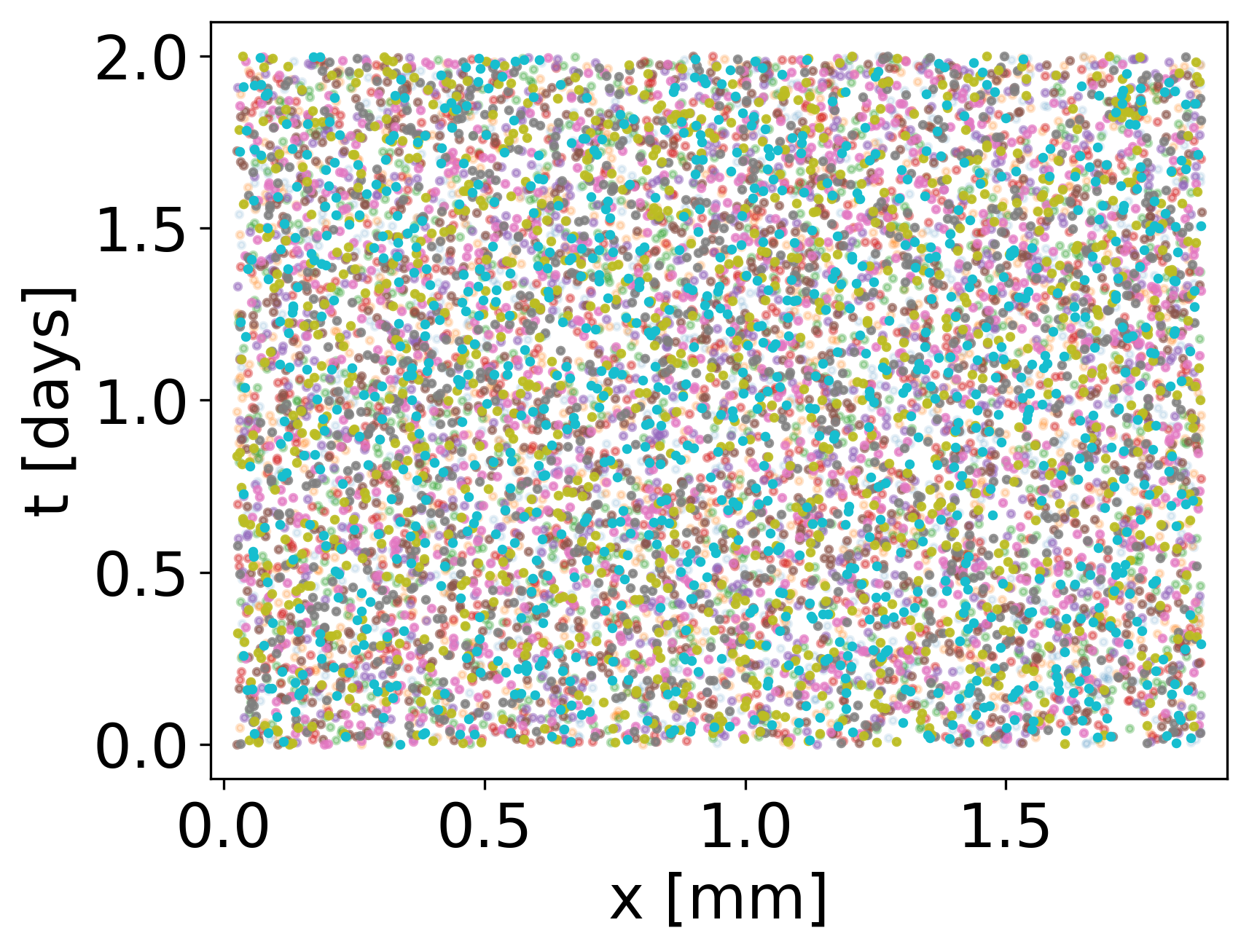}
                    \end{minipage}
                \end{minipage}
            };
            \node[
                fill=white,
                text=black,
                inner xsep=5pt,
                inner ysep=1pt,
                font=\scriptsize\bfseries
            ] at (collocationbox.north) {Collocation points};
        \end{tikzpicture}
    \end{minipage}
\caption{
\textbf{Data points versus collocation points in $\boldsymbol{1\mathrm{D}{+}t}$.}
\textit{Left}: data grid for $u_{\mathrm{data}}$, comprising $38\times5$ points.
\textit{Right}: collocation points sampled over $1$ and $10$ training minibatches, coloured by minibatch.
For $1$ minibatch, a single set $\mathcal{C}_1=\{(x_k^{(c)},t_k^{(c)})\}_{k=1}^{n_c}$ is sampled; for $10$ minibatches, the sets $\mathcal{C}_1,\ldots,\mathcal{C}_{10}$ illustrate how repeated resampling of random collocation points progressively fills the spatiotemporal domain.
}
\label{fig:collocation-points}
\end{figure}

\paragraph{PDE loss.}
The PDE objective function penalises deviations of the learned solution from the governing equation 
at a set of $n_c$ randomised collocation points 
$\{(x_{1,k}^{(c)}, x_{2,k}^{(c)}, t_k^{(c)})\}_{k=1}^{n_c}.$
During learning, these points are repeatedly and independently sampled 
from the uniform distribution defined by the coordinate-wise minima and maxima of the observed data. The PDE loss is defined as the mean-squared residual evaluated at the collocation points,
\begin{align}
\mathcal{L}_{\text{PDE}}(\hat{u}, \hat{D}, \hat{G})
&=
\frac{1}{n_c}
\sum_{k=1}^{n_c}
\left[
\left.\frac{\partial \hat{u}}{\partial t}\right|_{k}
-
\left.\nabla\!\cdot\!\left(\hat{D}(\hat{u})\,\nabla \hat{u}\right)\right|_{k}
-
\hat{G}_k\,\hat{u}_k
\right]^2,
\label{eq:pde_loss}
\end{align}
where $\hat{u}_k = \hat{u}(x_{1,k}^{(c)}, x_{2,k}^{(c)}, t_k^{(c)})$, $\hat{D}_k = \hat{D}(\hat{u}_k)$, and
$\hat{G}_k = \hat{G}(\hat{u}_k)$. 
The notation 
$\left.f\right|_{k}$ denotes evaluation of a function $f$ and its derivatives
at the $k$-th collocation point. 
The distinction between data points and collocation points is illustrated in Fig.~\ref{fig:collocation-points} for a \(1\text{D}{+}t\) case.

\paragraph{Biological constraint loss.}
Biological plausibility is enforced through an additional loss term imposing explicit constraints on diffusion and population growth. For example, in many cell-based experiments, the diffusion function $D(\cdot)$ is assumed to be non-decreasing with respect to density and bounded as $0\le d_{\min} \leq D(\cdot)\leq d_{\max}$, while the per-capita growth rate $G(\cdot)$ is assumed to be non-increasing with respect to density and bounded as $g_{\min}\leq G(\cdot) \leq g_{\max}$.
In this setting, the biologically motivated penalty is imposed on a fixed, equally spaced grid of $n_{\text{bio}}$ density values
$
u_q \in [u_{\min}, u_{\max}], q=1,\ldots,n_{\text{bio}},
$
where \(u_{\min}\) and \(u_{\max}\) denote the minimum and maximum density values observed in the data, producing the biological loss
\begin{align}
\mathcal{L}_{\mathrm{bio}}(\hat{D},\hat{G})
&=
\sum_{q=1}^{n_{\text{bio}}}
\Bigg[
\lambda_{D_{\lim}}
\Big(
\mathbb{I}_{\{\hat{D}_q<d_{\min}\}}(\hat{D}_q-d_{\min})^2
+
\mathbb{I}_{\{\hat{D}_q>d_{\max}\}}(\hat{D}_q-d_{\max})^2
\Big)
\nonumber \\[3pt]
&\quad
+
\lambda_{G_{\lim}}
\Big(
\mathbb{I}_{\{\hat{G}_q<g_{\min}\}}(\hat{G}_q-g_{\min})^2
+
\mathbb{I}_{\{\hat{G}_q>g_{\max}\}}(\hat{G}_q-g_{\max})^2
\Big)
\nonumber \\[3pt]
&\quad
+
\lambda_{\partial D}
\mathbb{I}_{\left\{
\left.\frac{\partial \hat{D}}{\partial u}\right|_{u_q}<0
\right\}}
\left(
\left.\frac{\partial \hat{D}}{\partial u}\right|_{u_q}
\right)^2
+
\lambda_{\partial G}
\mathbb{I}_{\left\{
\left.\frac{\partial \hat{G}}{\partial u}\right|_{u_q}>0
\right\}}
\left(
\left.\frac{\partial \hat{G}}{\partial u}\right|_{u_q}
\right)^2
\Bigg],
\label{eq:bio_loss}
\end{align}
where the indicator function $\mathbb{I}_{\{\text{condition}\}}$ equals $1$ if the condition holds and $0$ otherwise. The coefficients $\lambda_{D_{\lim}}$ and $\lambda_{G_{\lim}}$ weight the penalties enforcing the prescribed bounds on $\hat{D}$ and $\hat{G}$, respectively, while $\lambda_{\partial D}$ and $\lambda_{\partial G}$ weight the corresponding derivative-constraint penalties.

\subsection{Training--validation split}
In BINNs, state data are randomly partitioned into training and
validation sets by assigning grid points independently of space
and time, with the point selection varying across random seeds
\cite{Lagergren2020}. The training set is used to fit the BINN parameters, while the validation set is used to monitor overfitting via early stopping (Supplementary Material~\ref{sec:SM:ES}); throughout, we use an $80/20$ training--validation (TV) split. 
Predictions of the state and RHS MLPs are then performed on the
combined training and validation data. This reflects our aim to learn accurate RHS terms over the observed range of the state quantity, rather than extrapolating to unseen state-quantity regimes. 

\begin{figure}[h!]
    \centering
    \begin{minipage}[c]{0.66\textwidth}
        \centering
        \begin{tikzpicture}
            \node[
                draw=lightgray,
                line width=1pt,
                rounded corners=5pt,
                inner sep=0pt,
                text width=\linewidth,
                align=center
            ] (iconebox) {
                \begin{minipage}{0.95\linewidth}
                    \centering
                    \vspace{0.45cm}
                    \begin{minipage}[c]{0.47\linewidth}
                        \centering
                        {\scriptsize\text{\hspace{0.3cm} \textbf{TV split 1}}}
                        \vspace{0.2em}
                        \includegraphics[width=\linewidth]{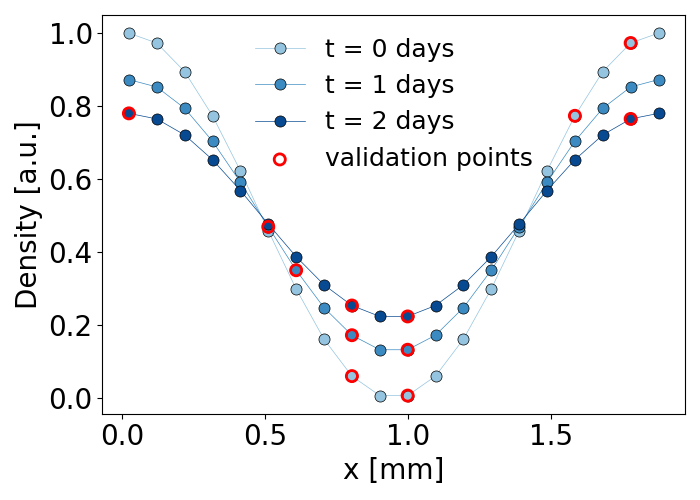}
                        \vspace{0.4em}
                        \includegraphics[width=\linewidth]{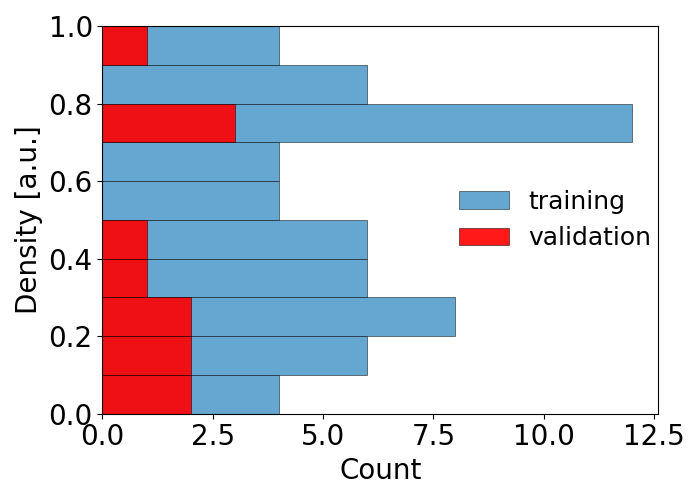}
                    \end{minipage}
                    \hfill
                    \begin{minipage}[c]{0.47\linewidth}
                        \centering
                        {\scriptsize\text{\hspace{0.3cm} \textbf{TV split 2}}}
                        \vspace{0.2em}
                        \includegraphics[width=\linewidth]{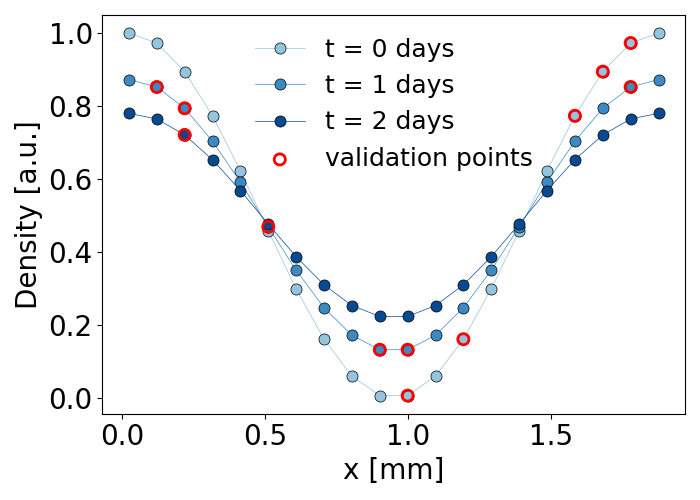}
                        \vspace{0.4em}
                        \includegraphics[width=\linewidth]{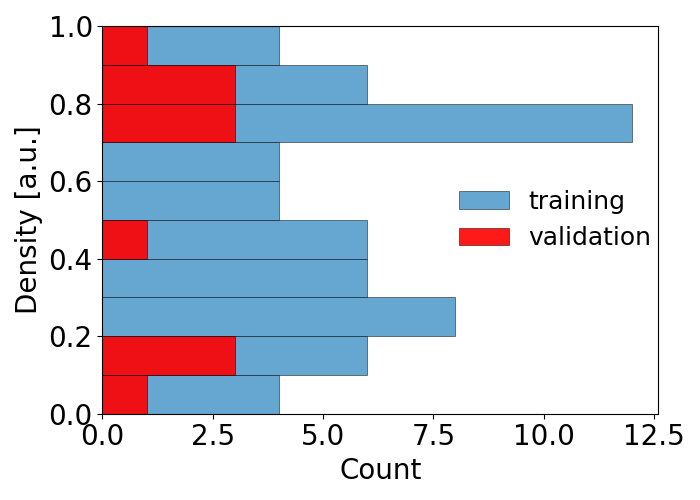}
                    \end{minipage}
                    \vspace{0.2cm}
                \end{minipage}
            };
            \node[
                fill=white,
                text=black,
                inner xsep=5pt,
                inner ysep=1pt,
                font=\scriptsize\bfseries
            ] at (iconebox.north) {\textbf{Training--validation (TV) splits for first initial condition}};
        \end{tikzpicture}
    \end{minipage}
    \hfill
    \begin{minipage}[c]{0.30\textwidth}
        \centering
        {\scriptsize\textbf{\hspace{0.3cm} Training--validation split 1 for second initial condition}}
        \vspace{0.2em}
        \includegraphics[width=\linewidth]{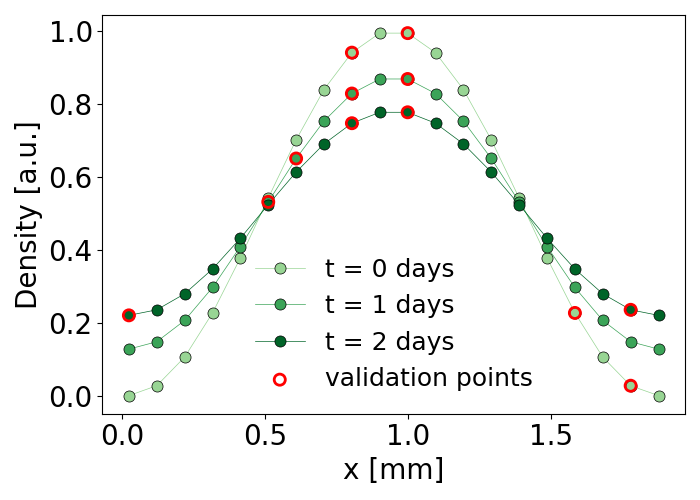}
        \vspace{0.4em}
        \includegraphics[width=\linewidth]{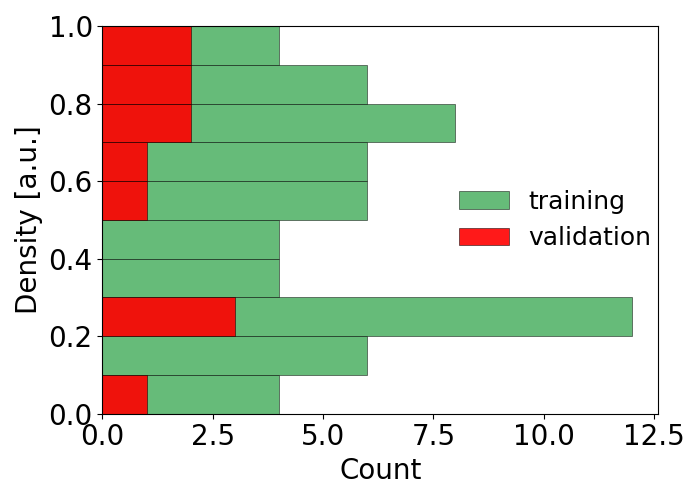}
    \end{minipage}
    \caption{
    \textbf{Training and validation sets under different training--validation (TV) splits and initial conditions.}
    Panels are generated using $1\text{D}{+}t$ data sampled at 38 spatial and 5 temporal points with an 80/20 TV split.
    \textit{Top row} shows spatial density profiles at 3 time points. \textit{Bottom row} shows the corresponding training and validation distributions over density. The two boxed panels use a cosine initial condition and differ only in the random seed defining the TV split, while the stand-alone panel uses an inverted cosine initial condition with the same seed as TV split 1.}
    \label{fig:TV_split}
\end{figure}

Partitioning at the level of individual grid points, rather than
whole trajectories or repeats, is particularly suited to the
sparse-data settings that motivate BINNs \cite{Lagergren2020}. 
In many biological experiments only a handful of repeats are
available, and in fields such as ecology the observed dynamics may
correspond to a real-world process that unfolds only once, so that
training on one complete run and validating on others is not
possible. 
Randomly holding out individual grid points allows a
TV split to be formed even from a single
realisation.
Illustrations of TV splits, and how the contents of the training and validation sets depend on the random seed, are shown in Fig.~\ref{fig:TV_split}.

\section{Methods: hyperparameter-selection workflow} \label{sec:pipeline}

\subsection{Hyperparameters in BINNs} \label{sec:methods_intro}
We categorise BINN hyperparameters into three groups based on how practitioners set them. The first group comprises generic neural-network settings that practitioners typically leave at standard defaults rather than tune: the optimiser, the batch size, the TV split, and the early-stopping threshold (the relative improvement in validation loss that counts as sufficient, and so resets the early-stopping patience). The second group comprises BINN-specific settings that practitioners typically set from domain knowledge and practical considerations rather than tune: the loss weights, the number of collocation points, and the activation functions. The third group comprises the BINN hyperparameters that must be tuned for each new combination of data and RHS structure: the MLP capacities (width and depth) and the early-stopping patience.

In this work we present a workflow for selecting the third group, controlling capacity primarily through width. For completeness, we also demonstrate how to evaluate the second group's choices in Supplementary Material~\ref{sec:SM1}. The first group we set once and leave untouched, though one could readily carry out checks analogous to those for the second group.

\subsection{Diagnostic hyperparameter-selection workflow}
In a BINN, the MLP capacities and early-stopping patience are interdependent: their effects on accuracy and cost interact. Therefore, tuning each hyperparameter in isolation risks a selection that is optimal only for the particular settings of the others, while a full joint sweep over all hyperparameters is usually infeasible. As a practical middle ground, we design a semi-sequential workflow whose order is dictated by the BINN architecture, so that it follows the network's data flow: the density predicted by the state network is the input to the RHS networks, and the coefficient functions can only be recovered once the density is adequately represented. Accordingly, we fix the state network capacity first, then the RHS capacities, and finally the early-stopping patience. Each choice is then confirmed to hold as the downstream hyperparameters are selected. To mitigate the stochasticity introduced by the TV split, every configuration is run across several random seeds (here, 3), and each diagnostic is read from the resulting spread rather than from a single training run.
The workflow is illustrated in Fig.~\ref{fig:pipeline_flow} and reads its decisions from three ground-truth-free diagnostics:

\begin{itemize}[leftmargin=*, itemsep=0.2em]
\item \textbf{Cost-performance trade-off charts} show the final validation loss and runtime, with a subset of the hyperparameters varied and the rest held fixed. Reading across a swept hyperparameter, a saturation point marks where further increases cease to be worth their cost. 
\item \textbf{Training dynamics trace plots} show the validation loss over epochs. These reveal whether performance differences between hyperparameter configurations are systematic rather than artefacts of where training stopped, and whether further training is expected to reduce the loss.
\item \textbf{Consistency plots} show the learned RHS coefficient functions over the data-supported range of the density $u$, overlaid across hyperparameter configurations. The hyperparameter value beyond which these functions stop changing signals convergence.
\end{itemize}

Runtimes are measured on an Apple M3 Pro (CPU-only) up to the saved model
($e_{\mathrm{sav}}$; Algorithm~\ref{alg:early_stopping}, Supplementary
Material~\ref{sec:SM:ES}). Because machine load varies, runtimes should be
compared only within individual case studies.

\definecolor{c1}{RGB}{ 66,110,158}
\definecolor{c1f}{RGB}{226,236,246}
\definecolor{c2}{RGB}{ 63,133, 94}
\definecolor{c2f}{RGB}{228,241,232}
\definecolor{c3}{RGB}{190,124, 48}
\definecolor{c3f}{RGB}{251,240,225}
\definecolor{c4}{RGB}{ 96,102,110}
\definecolor{c4f}{RGB}{240,241,243}

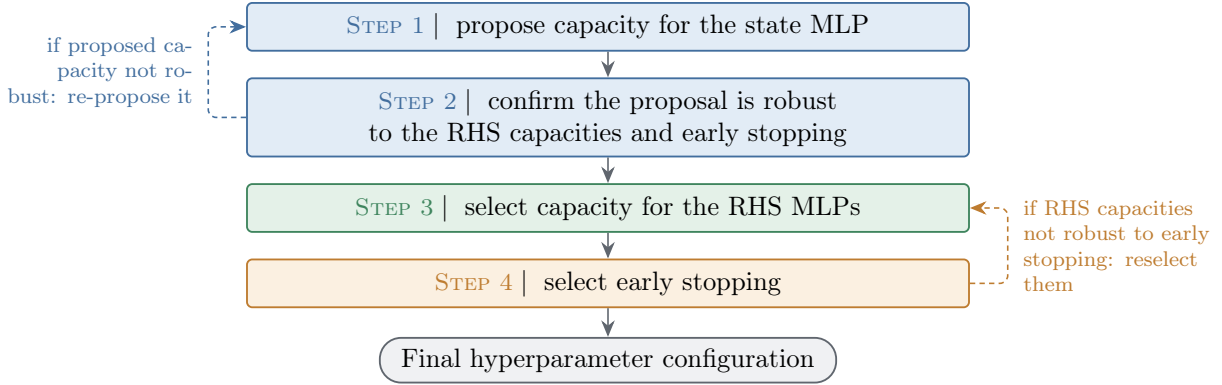
\begin{figure}[t]
\centering
\begin{tikzpicture}[
  font=\small,
  node distance=3.4mm and 0mm,
  proc/.style={rectangle, rounded corners=2pt, draw=#1, line width=0.7pt,
               align=center, text width=9.2cm, inner xsep=5pt, inner ysep=4pt},
  proc/.default=c4,
  term/.style={rectangle, rounded corners=8pt, draw=c4, line width=0.7pt,
               fill=c4f, align=center, inner xsep=8pt, inner ysep=4pt},
  arr/.style={-{Stealth[length=2.2mm,width=1.8mm]}, draw=c4, line width=0.6pt},
  loop/.style={-{Stealth[length=2.2mm,width=1.8mm]}, draw=#1, line width=0.6pt,
               dashed, dash pattern=on 2pt off 1.6pt, rounded corners=3pt},
  loop/.default=c4,
  lab/.style={font=\scriptsize, inner sep=0pt, text width=2.7cm}
]

\node[proc=c1, fill=c1f] (s1)
  {{\scshape\footnotesize\color{c1}Step 1}\;\textbar\; propose capacity for the state MLP};
\node[proc=c1, fill=c1f, below=of s1] (s2)
  {{\scshape\footnotesize\color{c1}Step 2}\;\textbar\; confirm the proposal is robust to the RHS capacities and early stopping};
\node[proc=c2, fill=c2f, below=of s2] (s3)
  {{\scshape\footnotesize\color{c2}Step 3}\;\textbar\; select capacity for the RHS MLPs};
\node[proc=c3, fill=c3f, below=of s3] (s4)
  {{\scshape\footnotesize\color{c3}Step 4}\;\textbar\; select early stopping};
\node[term, below=3.8mm of s4] (end)
  {Final hyperparameter configuration};

\foreach \a/\b in {s1/s2, s2/s3, s3/s4, s4/end} \draw[arr] (\a) -- (\b);

\draw[loop=c1] (s2.west) -- ++(-5mm,0) coordinate (lc) |- (s1.west);
\node[lab, anchor=east, align=right, text=c1]
  at ([xshift=-2mm]$(lc)!0.5!(lc|-s1.west)$)
  {if proposed capacity not robust: re-propose it};

\draw[loop=c3] (s4.east) -- ++(5mm,0) coordinate (rc) |- (s3.east);
\node[lab, anchor=west, align=left, text=c3]
  at ([xshift=2mm]$(rc)!0.5!(rc|-s3.east)$)
  {if RHS capacities not robust to early stopping: reselect them};

\end{tikzpicture}
\caption{\textbf{The hyperparameter-selection workflow.} A semi-sequential procedure: the hyperparameters are decided in an order determined by the biologically-informed neural network architecture. The capacity---network width and/or depth---of the state multilayer perceptron (MLP), which must learn the data, is proposed first, and is confirmed and selected once it is robust to the downstream MLP capacities and early stopping, and otherwise re-proposed (dashed loop). The capacities of the RHS MLPs are then selected, followed by the early-stopping patience; if the RHS capacities prove not robust to that patience they are reselected (second dashed loop).}
\label{fig:pipeline_flow}
\end{figure}

\subsection{Synthetic data generation}
To ensure suitability for biological applications, we evaluate the BINN architecture on biologically relevant simulation data generated from known ground-truth functions so that the learned functions can be assessed against a known reference.
The ground-truth coefficient functions are restricted to a biologically admissible space, with bounds $(d_{\min}, d_{\max}) = (0, 0.1)\ \mathrm{mm}^2\,\mathrm{days}^{-1}$ for diffusion
and $(g_{\min}, g_{\max}) = (-0.48, 2.4)\ \mathrm{days}^{-1}$ for growth. These values are set from observations in scratch-assay experiments \cite{Jin2016} with some additional flexibility following \textcite{Lagergren2020}, yielding a suite of physically plausible diffusion and reaction--diffusion equations that are summarised in Table~\ref{tab:expressions} and visualised in Fig.~\ref{fig:REqs_schematic} of Appendix~\ref{app:data_gen}.

All simulations are performed subject to no-flux boundary conditions at the domain boundaries. 
The governing equations are first solved on a fine spatiotemporal grid. To match realistic experimental resolution, spatial data are then obtained by binning and averaging the solution, and temporal data are sampled at discrete time points. 
The space and time resolutions are chosen to reflect previous BINN studies with microscopy data in $1\text{D}{+}t$ \cite{Lagergren2020} and $2\text{D}{+}t$ \cite{lavery_physics-informed_2026} settings.
Initial conditions resemble a smooth scratch in the $1\text{D}{+}t$ case and a low-amplitude cosine-squared profile matching a localised seeding of cells on a plate in the $2\text{D}{+}t$ case. The data-generating settings are summarised in Table~\ref{tab:datagen}.
Biological constraint weights are set to $\lambda_{D_{\lim}}=\lambda_{G_{\lim}}=0$ and $\lambda_{\partial D}=\lambda_{\partial G}=10$, with $n_{\text{bio}}=100$ in Eq.~\eqref{eq:bio_loss}. 
The biological constraint weights are selected to enforce the constraints without impeding training during early epochs, as shown in Supplementary Material~\ref{sec:SM:bio_loss}. For enforcing the PDE constraint, we use $n_c=1000$ collocation points in $1\text{D}{+}t$ and $n_c=100$ in $2\text{D}{+}t$ (Eq.~\eqref{eq:pde_loss}). The value of $n_c$ is chosen to balance the increased runtime associated with larger $n_c$ against the greater stochasticity arising from evaluating the PDE at fewer points when $n_c$ is smaller, as demonstrated in Supplementary Material~\ref{sec:SM:collocation_pts}.
\begin{table}[h!]
\caption{
\textbf{Ground-truth diffusion and growth functions used to generate the simulated data.} Functions are defined on the dimensionless density $\tilde{u} = u / K$; $K = 1700\text{ cells mm}^{-2}$. }
\footnotesize
\centering
\begin{tabular}{@{}ccll@{}}
\toprule
\textbf{PDE term} & \textbf{Function} & \textbf{Expression} & \textbf{Parameters (units)} \\
\midrule
\multirow{4}{*}{Diffusion}
& $D_1$
& $d_{1,1}$
& $d_{1,1}=0.02472~\mathrm{mm}^2\,\mathrm{days}^{-1}$ \\
& $D_2$
& $d_{2,1} + d_{2,2}\,\tilde{u}$
& $(d_{2,1},d_{2,2})=(0.015,\,0.06)~\mathrm{mm}^2\,\mathrm{days}^{-1}$ \\
& $D_3$
& $d_{3,1} + d_{3,2}\,\tilde{u}^2$
& $(d_{3,1},d_{3,2})=(0.01,\,0.044)~\mathrm{mm}^2\,\mathrm{days}^{-1}$ \\
& $D_4$
& $d_{4,1} + d_{4,2}\!\left(1 - e^{-d_{4,3} \tilde{u}}\right)$
& $\begin{aligned}
(d_{4,1},d_{4,2}) &= (0.003,\,0.095)~\mathrm{mm}^2\,\mathrm{days}^{-1},~d_{4,3} = 2.5
\end{aligned}$ 
\\ \cmidrule{1-4}
\multirow{4}{*}{Growth}
& $G_1$
& $g_{1,1}$
& $g_{1,1}=0.65~\mathrm{days}^{-1}$ \\
& $G_2$
& $g_{2,1} + g_{2,2}\,\tilde{u}$
& $(g_{2,1},g_{2,2})=(1.2,\,-1.5)~\mathrm{days}^{-1}$ \\
& $G_3$
& $g_{3,1} + g_{3,2}\, \tilde{u}^2$
& $(g_{3,1},g_{3,2})=(1.05,\,-0.145)~\mathrm{days}^{-1}$ \\
& $G_4$
& $g_{4,1} + g_{4,2}\!\left(1 - e^{-g_{4,3} \tilde{u}}\right)$
& $(g_{4,1},g_{4,2})=(0.35,\,-0.65)~\mathrm{days}^{-1}$,~$g_{4,3}=4$ \\
\bottomrule
\end{tabular}
\label{tab:expressions}
\end{table}
\begin{table}[h!]
\caption{\textbf{Data-generation settings.}
Normalised spatial coordinates are defined as
$\xi=(x-x_{\min})/\mathrm{L}$ in $1\mathrm{D}{+}t$ and
$\xi_i=(x_i-x_{i,\min})/\mathrm{L}_i$ in $2\mathrm{D}{+}t$.
MPE denotes mean percentage error.}
\footnotesize
\centering
\begin{tabular}{@{}lcc@{}}
\toprule
& $\boldsymbol{1\mathrm{D}{+}t}$ & $\boldsymbol{2\mathrm{D}{+}t}$ \\
\midrule
Spatial grid points
& $38$ & $11\times 11$ \\
Spatial domain
& $\mathrm{L}=1.85\,\mathrm{mm}$
& $\mathrm{L}_1=\mathrm{L}_2=1.85\,\mathrm{mm}$ \\
Time points
& $5$ & $9$ \\
Duration
& $2\,\mathrm{days}$ & $2\,\mathrm{days}$ \\
Noise (MPE, $\gamma$)
& — & $(5\%,~1)$ \\
Initial condition
& $\tfrac{1}{2}\left[1+\cos(2\pi\xi)\right]$
& $\tfrac{1}{4}
   \cos^{2}\!\bigl(\pi(\xi_1-\tfrac12)\bigr)
   \cos^{2}\!\bigl(\pi(\xi_2-\tfrac12)\bigr)$ \\
\bottomrule
\end{tabular}
\label{tab:datagen}
\end{table}

\section{Results}
We apply the workflow to four cases of increasing complexity, each highlighting a different aspect of hyperparameter selection.
We first demonstrate the full workflow on a system with a single RHS term: a constant diffusion coefficient function (Section~\ref{sec:res-case-1}).
Subsequent cases then isolate harder sub-problems: a non-trivial diffusion coefficient function that may require increased MLP capacity (Section~\ref{sec:res-case-2}); two coupled RHS MLPs---diffusion and growth---requiring joint selection (Section~\ref{sec:res-case-3}); and noisy $2\text{D}{+}t$ data that tests the workflow's robustness to measurement noise in a second spatial dimension (Section~\ref{sec:res-case-4}).

\subsection{Case~I: demonstrating the full workflow (constant diffusion in \texorpdfstring{$\boldsymbol{1\mathrm{D}{+}t}$}{1D+t})}
\label{sec:res-case-1}

We first consider a system with constant diffusion $D_1$ and no growth (Table~\ref{tab:expressions}), and use it to demonstrate the full hyperparameter-selection workflow, following the steps of Fig.~\ref{fig:pipeline_flow}. Here, the state (density) capacity is controlled through $W_u$, and the RHS (diffusion) capacity through $W_D$. 
To streamline the workflow, we hold the MLP depths---the numbers of hidden layers $L_u$, $L_D$ and $L_G$---fixed at three, a decision justified in the note on MLP depth.
\paragraph{Step 1: propose $W_u$.}
We first fix the diffusion network at a deliberately small width, $W_D = 2$, to limit its flexibility and ensure that the learning is driven primarily by the density network width, $W_u$. We then sweep $W_u \in \{8,16,32,64,128\}$ across early-stopping patiences $ES\in\{1000,2000,3000\}$ and visualise the cost-performance trade-off (Fig.~\ref{fig:res1}a).
Here, across $ES$ values, the validation loss improves consistently with $W_u$, and the runtime per unit of validation loss continues to favour wider networks. However, by $W_u = 128$ the absolute cost becomes excessive, with a single run at $ES = 2000$ taking roughly $2 \times 10^{3}\,\mathrm{s}$. We propose $W_u = 64$, the largest tested width whose runtime remains reasonable while the validation loss still improves appreciably.

\input{figures/results_fig_1}
\paragraph{Step 2: confirm and select $W_u$.}
Using two separate plots, we investigate whether the proposed $W_u$ is robust to different choices of $ES$ and $W_D$.
 For inspecting the effect of early stopping, we trace the validation-loss trajectories over epochs (Fig.~\ref{fig:res1}b). Here, the curves stay separated by $W_u$ throughout training for every tested $ES \in \{1000, 2000, 3000\}$, confirming that the ranking reflects $W_u$ itself, not when training happened to stop.
 For $W_D$-inspection, we plot the cost-performance trade-off across $W_D\in\{1,2,4\}$ (Fig.~\ref{fig:res1}c), i.e., the $W_D$-based counterpart of the early-stopping-focused sweep in Step~1. The trade-off shows that $W_u=64$ performs well across the tested $W_D$. On both counts, the $W_u=64$ proposed in Step~1 therefore stands. Had a different $W_u$ been preferable at some $W_D$, we would loop back to Step~1 (dashed loop in Fig.~\ref{fig:pipeline_flow}) and re-plot the cost-performance trade-off over $W_u$ at that $W_D$.
\paragraph{Step 3: select $W_D$.}
With $W_u=64$ now locked in, we select the diffusion width $W_D$. Here, since Step~2 already varied $W_D$, we can reuse those cost-performance plots (Fig.~\ref{fig:res1}c). Varying $W_D$ between 1 and 4 has little impact on either validation loss or computational cost. Since the choice is therefore largely inconsequential, we select $W_D = 2$ as a low-cost safeguard, retaining slightly more flexibility than $W_D = 1$ affords.

\paragraph{Step 4: select early stopping.}
Finally, we select the early-stopping patience by reading the validation loss and the consistency of the learned functions together. The validation loss (Fig.~\ref{fig:res1}a) continues to drop reasonably systematically with increasing early-stopping patience, so on its own it gives no clear stopping point. We therefore turn to the consistency of the diffusion predictions (Fig.~\ref{fig:res1}d): the learned diffusion function $\hat D(u)$ varies only mildly across $ES$; beyond $ES=1000$ the continuing fall in validation loss no longer changes the learned diffusion appreciably. We select $ES=2000$, a conservative choice that captures the gains that matter for function recovery at modest extra training time.

\paragraph{A note on MLP depth.} The depth of each network is set empirically, but the choices are consistent with a simple expectation. The RHS coefficient functions, such as the diffusion \(D(u)\), are low-dimensional and smooth, so we expect them to require little depth: universal-approximation results guarantee that even a single hidden layer suffices in principle \cite{cybenko_approximation_1989, hornik_approximation_1991}, and the targets lack the complex, compositional structure for which added depth is known to give substantial gains \cite{telgarsky_benefits_2016, eldan_power_2016}. The same approximation guarantees apply to \(\mathrm{NN}_u\), but its target is the richer density field, so whether depth helps there is a genuinely empirical question. We resolve both by sweeping depth and width. Figure~\ref{fig:depth} compares validation loss and runtime across widths and
depths for \(\mathrm{NN}_u\) (panel a) and \(\mathrm{NN}_D\) (panel b), and shows
the consistency of the recovered diffusion function across depths of both
networks (panels c,d). Two observations stand out: (i) for the density network
(Fig.~\ref{fig:depth}a), validation loss drops markedly from depth two to
three but plateaus from three to four for widths \(W_u = 64, 128\); and (ii)
validation loss is largely insensitive to the depth of the diffusion network, as
depth two already suffices for the simple constant-diffusion ground truth.
Balancing runtime against performance, and retaining headroom for harder RHS forms, we use depth $3$ for both.

\begin{figure}[H]
\centering

\begin{minipage}{0.285\linewidth}
  \centering
  \cornerlabelimgshift[width=\linewidth]
    {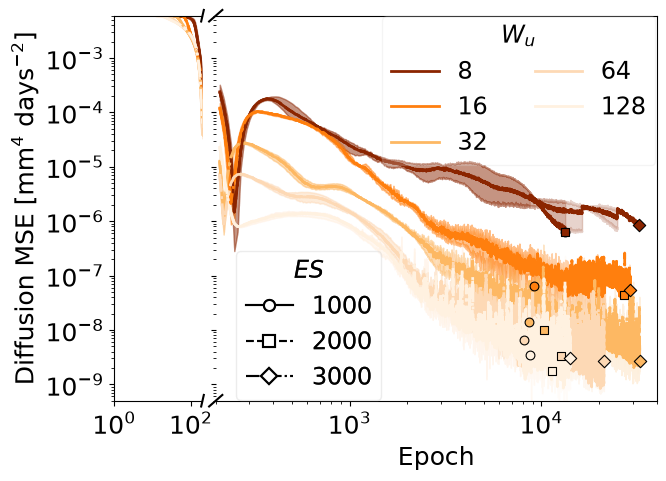}
    {(a)}{north}{xshift=10pt,yshift=10pt}
\end{minipage}
\hspace{0.2cm}
\begin{minipage}{0.285\linewidth}
  \centering
  \cornerlabelimgshift[width=\linewidth]
    {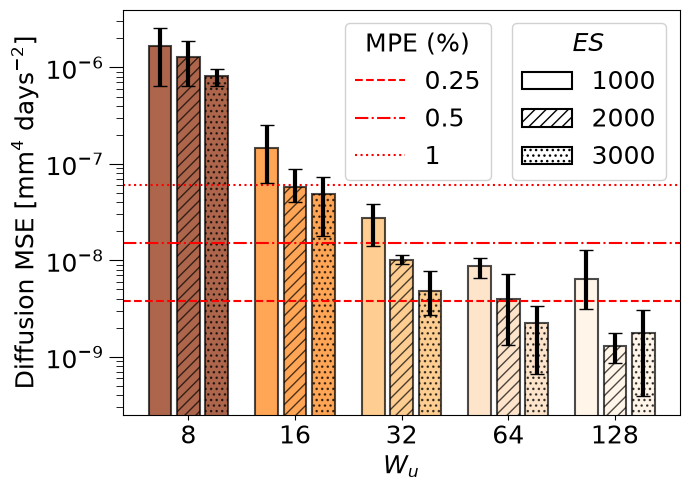}
    {(b)}{north}{xshift=10pt,yshift=10pt}
\end{minipage}
\hspace{0.2cm}
\begin{minipage}{0.285\linewidth}
  \centering
  \cornerlabelimgshift[width=\linewidth]
    {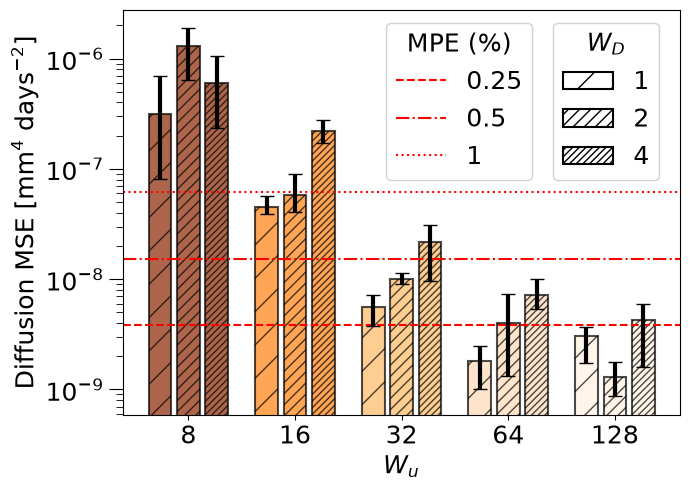}
    {(c)}{north}{xshift=10pt,yshift=10pt}
\end{minipage}
\vspace{-0.1cm}
\caption{\textbf{Ground-truth verification of the learned function in the synthetic Case~I study.} 
Data generated from a simple PDE with one term---constant diffusion---on the right-hand side (RHS). Mean-squared error (MSE) of the learned diffusion coefficient function $\hat D(u)$ against the ground-truth $D_1$. 
\textbf{(a)} MSE over training epochs, for different $W_u$ and early-stopping patiences $ES$ with $W_D=2$. Filled bands show min--max ranges across $3$ repeats with different training--validation (TV) splits.
\textbf{(b)} MSE across varying widths ($W_u$) of the state network and the early-stopping patience ($ES$), with the RHS network width fixed (here at $W_D=2$).
\textbf{(c)} The same MSE across varying values for $W_u$ and $W_D$, with fixed $ES=2000$. 
In \textbf{(b)} and \textbf{(c)}, the bars and whiskers respectively show mean values and min--max ranges across the different TV splits, and mean percentage error (MPE) levels of $0.25\%$, $0.5\%$, and $1\%$ are indicated by horizontal lines.
}
\label{fig:res2}
\end{figure}

\input{figures/results_fig_depth}
\paragraph{Verification.}
Because the data are synthetic, we can verify the ground-truth-free selections resulting from the workflow against the known diffusion function. Revisiting the $W_u$ sweep, we see that the diffusion-function mean-squared error (MSE) saturates with $W_u$ in the same way as the validation loss, and that its trajectory over epochs mirrors that of the validation loss (Fig.~\ref{fig:res2}a,b). From this we gather that the validation loss, which needs no ground truth, serves as an observable proxy for the otherwise-unobservable MSE. The $W_D$ sweep shows the same agreement. Notably, for all $W_u\in\{32,64,128\}$ and $W_D\in\{1,2,4\}$, the diffusion-function MSE stays below the level that a uniform $1\%$ error in the learned function would produce (Fig.~\ref{fig:res2}b,c); this is the sense in which all mean percentage error (MPE) reference lines in this work are defined (Supplementary Material~\ref{sec:SM:metrics}). This illustrates that a small diffusion network recovers the true function; taken together with the runtimes in Fig.~\ref{fig:res1}c, it further shows that larger $W_D$ carries little additional cost.

\subsection{Case~II: selecting \texorpdfstring{$W_D$}{W\_D} by consistency (non-constant diffusion in \texorpdfstring{$\boldsymbol{1\mathrm{D}{+}t}$}{1D+t})}
\label{sec:res-case-2}

\begin{figure}[H]
\centering

\setlength{\rowheaderwidth}{0.010\textwidth}

\setlength{\panelwidthLHS}{0.19\textwidth}
\setlength{\panelwidthRHS}{0.2\textwidth}

\setlength{\lhscolsep}{-1pt}
\setlength{\rhscolsep}{-20pt}
\setlength{\lhsrowsep}{0pt}
\setlength{\rhsrowsep}{-5pt}

\setlength{\boxheaderleftwidth}{0.36\textwidth}
\setlength{\boxheaderrightwidth}{0.54\textwidth}
\setlength{\rowheaderraiseA}{0.50cm}
\setlength{\rowheaderraiseB}{0.30cm}
\setlength{\rowheaderraiseC}{0.20cm}

\newcommand{\rowheaderfont}{\tiny}
\newcommand{\figheaderfont}{\scriptsize}

\newcommand{\esheaderpanel}[1]{%
\begin{minipage}[t]{\panelwidthRHS}
\centering
{\figheaderfont\boldmath #1}
\end{minipage}%
}
\newcommand{\emptypanelfive}{%
\begin{minipage}[t]{\panelwidthLHS}~\end{minipage}%
}
\newcommand{\emptyrowheader}{%
\begin{minipage}[t]{\rowheaderwidth}~\end{minipage}%
}
\newcommand{\rowheaderpanel}[2]{%
\begin{minipage}[t]{\rowheaderwidth}
\centering
\raisebox{#1}{\rotatebox{90}{\rowheaderfont\textbf{#2}}}
\end{minipage}%
}

\newcommand{\plainpanelfive}[1]{%
\begin{minipage}[t]{\panelwidthLHS}
\centering
\cornerlabelimgshift[width=\linewidth]
  {#1}{}{north}{xshift=5pt,yshift=-5pt}
\end{minipage}%
}

\newcommand{\linearspypanelfive}[2]{%
\begin{minipage}[t]{\panelwidthRHS}
\centering
\begin{tikzpicture}[
  spy using outlines={
    magnification=2,
    size=0.9cm,
    connect spies,
      rounded corners=2pt
  },
  thick
]
  \node[inner sep=0] (img) at (0,0)
    {\cornerlabelimgshift[width=\linewidth]
      {#1}{#2}{north}{xshift=5pt,yshift=-5pt}};
  \spy on (1.1,0.7)
    in node[anchor=north east] at (1.5,0.25);
\end{tikzpicture}
\end{minipage}%
}
\newcommand{\quadspypanelfive}[2]{%
\begin{minipage}[t]{\panelwidthRHS}
\centering
\begin{tikzpicture}[
  spy using outlines={
    magnification=2,
    size=0.9cm,
    connect spies,
      rounded corners=2pt
  },
  thick
]
  \node[inner sep=0] (img) at (0,0)
    {\cornerlabelimgshift[width=\linewidth]
      {#1}{#2}{north}{xshift=5pt,yshift=-5pt}};
  \spy on (1.2,0.7)
    in node[anchor=north east] at (0.1,1);
\end{tikzpicture}
\end{minipage}%
}
\newcommand{\expspypanelfive}[2]{%
\begin{minipage}[t]{\panelwidthRHS}
\centering
\begin{tikzpicture}[
  spy using outlines={
    magnification=2,
    size=0.9cm,
    connect spies,
      rounded corners=2pt
  },
  thick
]
  \node[inner sep=0] (img) at (0,0)
    {\cornerlabelimgshift[width=\linewidth]
      {#1}{#2}{north}{xshift=5pt,yshift=-5pt}};
  \spy on (1.1,0.7)
    in node[anchor=north east] at (1.5,0.25);
\end{tikzpicture}
\end{minipage}%
}

\begin{tikzpicture}
\matrix (m) [
    matrix of nodes,
    column sep=6pt,
    row sep=\lhsrowsep,
    nodes={inner sep=0pt, outer sep=0pt, anchor=center}
]{
    \emptyrowheader
    &
    \emptypanelfive
    &[\lhscolsep]
    \emptypanelfive
    &
    \esheaderpanel{$ES=2000$}
    &[\rhscolsep]
    \esheaderpanel{$ES=3000$}
    &[\rhscolsep]
    \esheaderpanel{$ES=5000$}
    \\[3pt]
    
    \rowheaderpanel{\rowheaderraiseA}{Linear, $D_2$}
    &
    \plainpanelfive{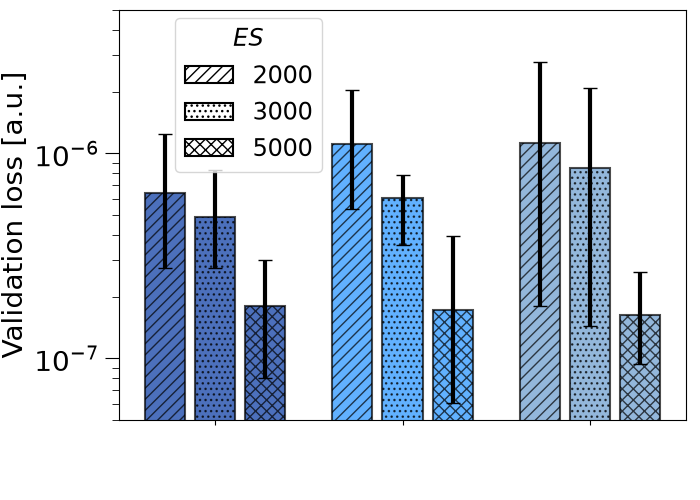}
    &[\lhscolsep]
    \plainpanelfive{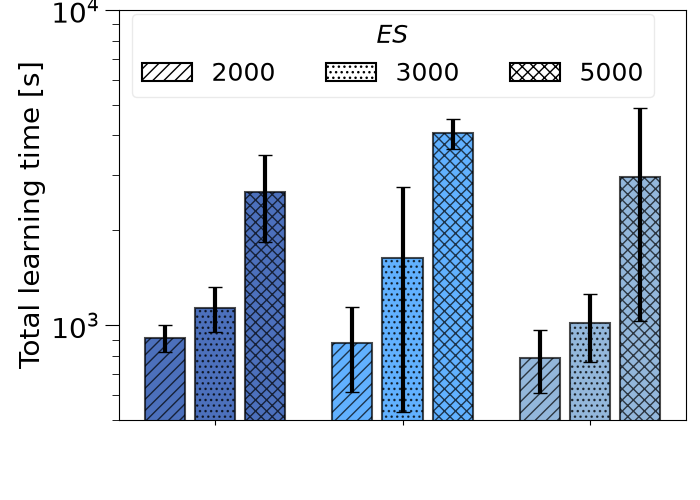}
    &
    \linearspypanelfive
      {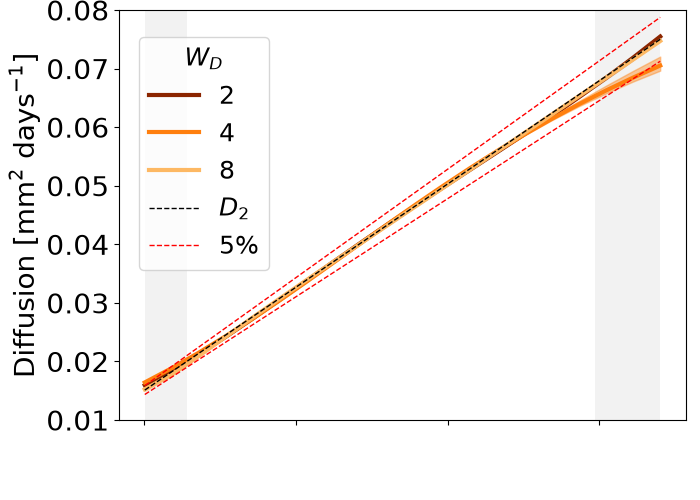}
      {}
    &[\rhscolsep]
    \linearspypanelfive
      {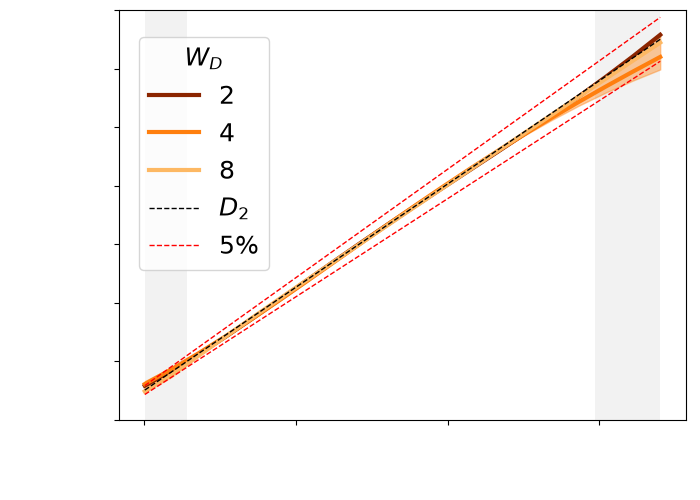}
      {}
    &[\rhscolsep]
    \linearspypanelfive
      {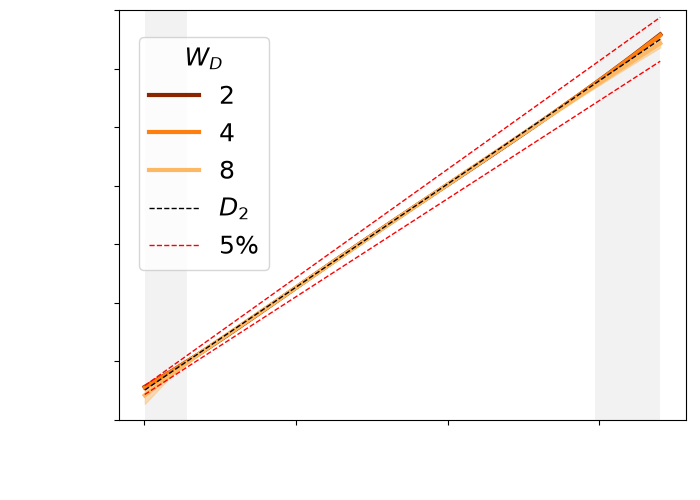}
      {}
    \\[\rhsrowsep]
    
    \rowheaderpanel{\rowheaderraiseB}{Quadratic, $D_3$}
    &
    \plainpanelfive{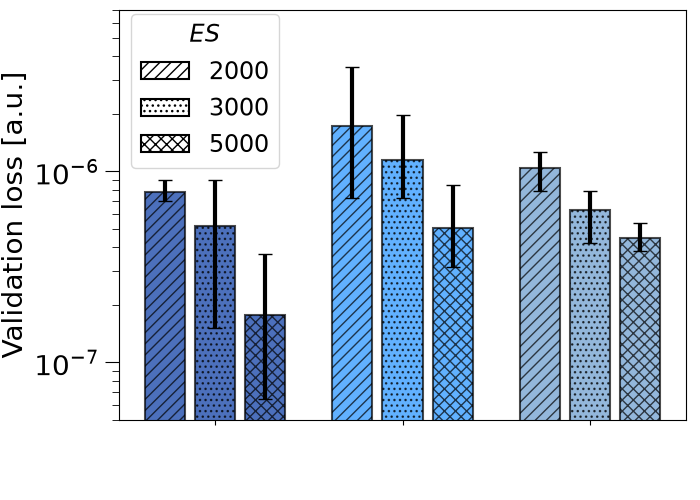}
    &[\lhscolsep]
    \plainpanelfive{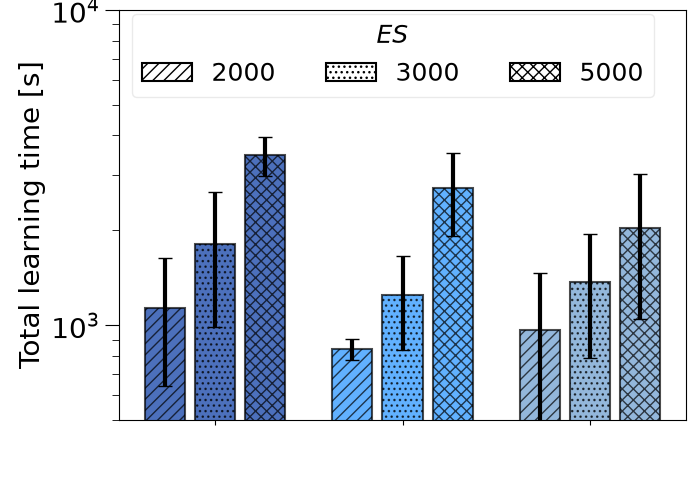}
    &
    \quadspypanelfive
      {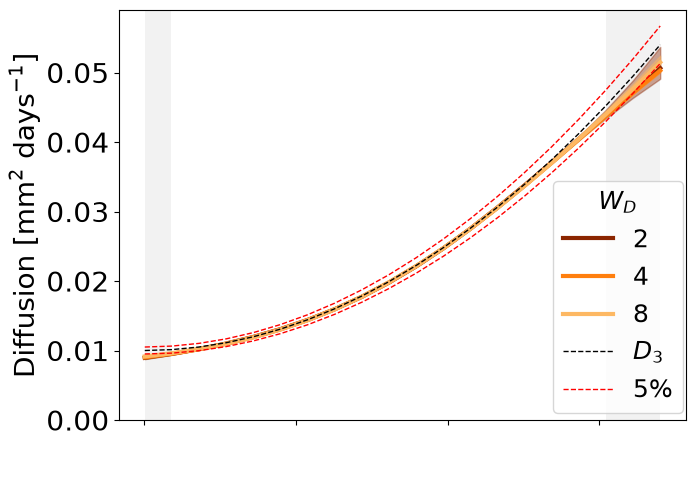}
      {}
    &[\rhscolsep]
    \quadspypanelfive
      {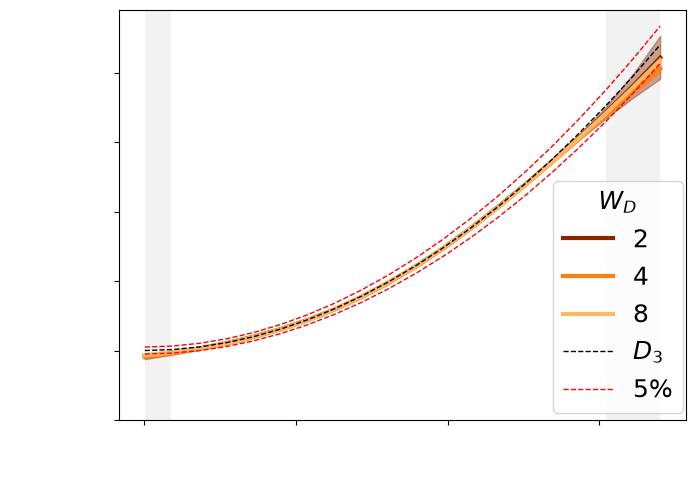}
      {}
    &[\rhscolsep]
    \quadspypanelfive
      {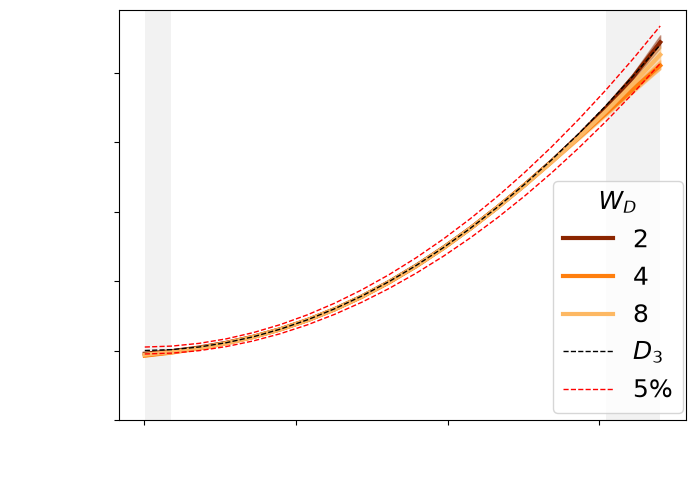}
      {}
    \\[\rhsrowsep]
    
    \rowheaderpanel{\rowheaderraiseC}{Exponential, $D_4$}
    &
    \plainpanelfive{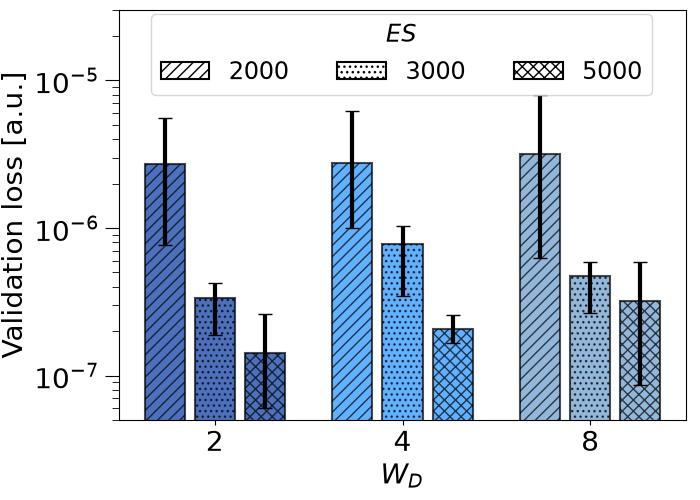}
    &[\lhscolsep]
    \plainpanelfive{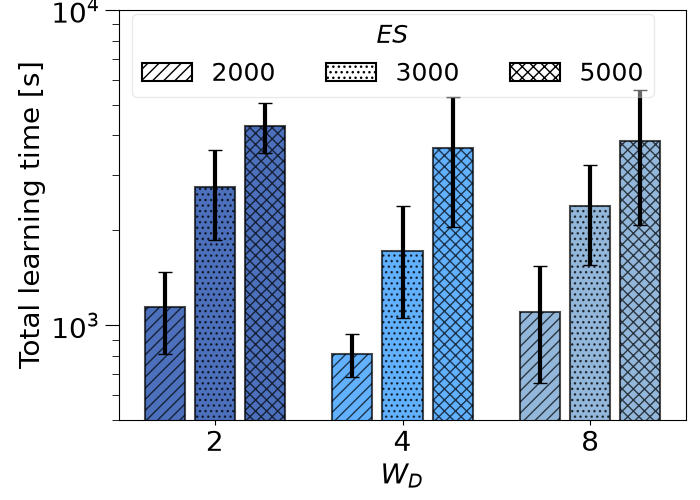}
    &
    \expspypanelfive
      {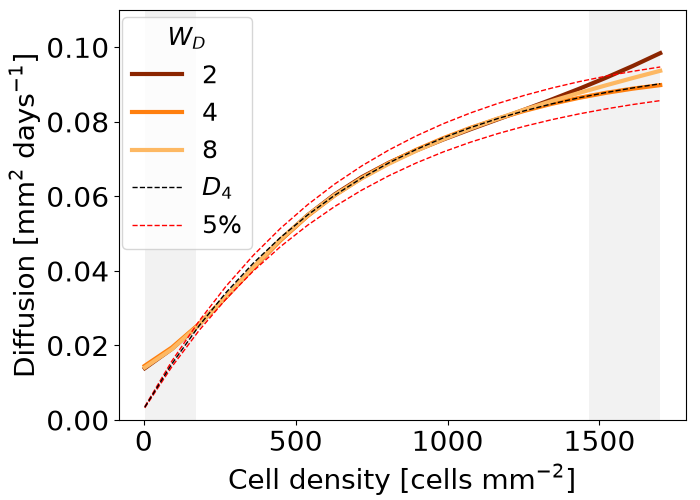}
      {}
    &[\rhscolsep]
    \expspypanelfive
      {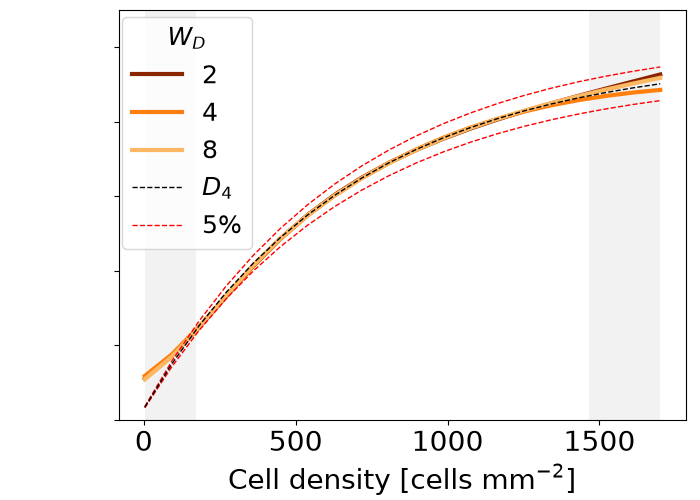}
      {}
    &[\rhscolsep]
    \expspypanelfive
      {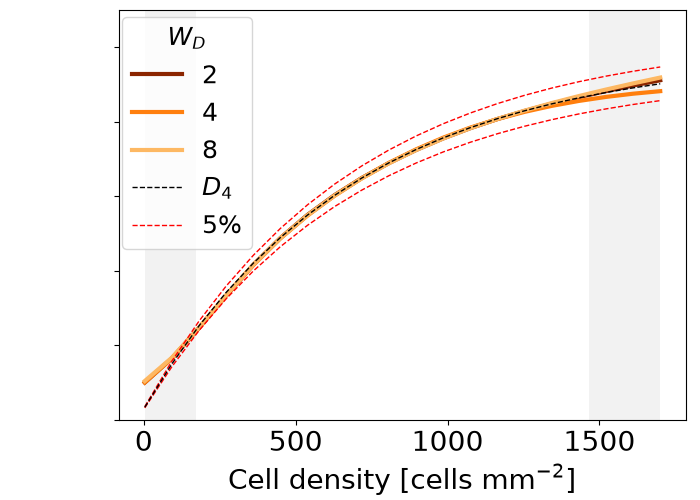}
      {}
    \\
};
\coordinate (boxNW) at ([xshift=-3pt,yshift=20pt]m.north west);
\coordinate (boxSE) at ([xshift= 3pt,yshift=2pt]m.south east);
\draw[
    lightgray,
    line width=1pt,
    rounded corners=5pt
]
(boxNW) rectangle (boxSE);
\coordinate (vDiv) at ($(m-1-3.east)!0.5!(m-1-4.west)$);
\draw[
    lightgray,
    line width=1pt
]
(vDiv |- boxNW) -- (vDiv |- boxSE);
\coordinate (leftHeaderC) at
  ($([xshift=-2pt]m-1-2.north west)!0.5!([xshift=2pt]m-1-3.north east)$);
\coordinate (rightHeaderC) at
  ($([xshift=-2pt]m-1-4.north west)!0.5!([xshift=2pt]m-1-6.north east)$);
\node[
    fill=white,
    text=black,
    font=\figheaderfont,
    anchor=north,
    inner xsep=1pt,
    inner ysep=0.5pt,
    align=center,
    text width=\boxheaderleftwidth
] at ([yshift=10pt]leftHeaderC)
{\textbf{(a)} Cost-performance across $W_D$};
\node[
    fill=white,
    text=black,
    font=\figheaderfont,
    anchor=north,
    inner xsep=1pt,
    inner ysep=0.5pt,
    align=center,
    text width=\boxheaderrightwidth
] at ([yshift=16pt]rightHeaderC)
{\textbf{(b)} Learned diffusion functions across $ES$};
\end{tikzpicture}
\vspace{-0.1cm}
\caption{\textbf{Selecting $W_D$ for non-constant diffusion.} Data are generated from three diffusion functions of increasing complexity: linear, quadratic, and exponential. 
\textbf{(a)} Trade-off between validation loss and runtime across varying widths ($W_D$) of the diffusion network and the early-stopping patience ($ES$), with the state network width fixed (here at $W_u=64$).
Bars and whiskers respectively show mean values and min--max ranges across $3$ repeats with different training--validation (TV) splits. 
\textbf{(b)} The learned diffusion function $\hat D(u)$ across these widths, shown at $ES=2000$, $3000$ and $5000$. 
Filled bands show min--max ranges across the $3$ 
TV splits. 
Regions outside the central $90\%$ of the density support are shown in grey.
The dashed ground-truth line and $\pm5\%$ bands play no role in the selection workflow and are shown only for reference.}
\label{fig:res3}
\end{figure}

We now apply the hyperparameter-selection workflow to diffusion functions that are more complicated than constant, and thus may require more capacity in the diffusion MLP. Specifically, we consider linear ($D_2$), quadratic ($D_3$), and exponential ($D_4$) diffusion (Table~\ref{tab:expressions}).
The state network, however, fits data of comparable complexity to Case~I, so the same $W_u$ remains appropriate. We can therefore reuse $W_u=64$ and enter the pipeline directly at Step~3, producing cost-performance plots at fixed $W_u$ over varying $W_D$ and $ES$, whose smallest values are those selected in Case~I. Here we test $W_D\in\{2,4,8\}$ and $ES\in\{2000,3000,5000\}$. Doing so, we find that the cost-performance trade-off charts alone do not pin down a preferred $W_D$ for any of the three diffusion functions (Fig.~\ref{fig:res3}a). Across the tested widths, neither the validation loss nor the runtime changes systematically with $W_D$: for each diffusion function, the differences between widths are small relative to the spread across TV splits, and the wider networks $W_D\in\{4,8\}$ do not improve on $W_D=2$. Both metrics instead depend mainly on the early-stopping patience, with the validation loss decreasing and the runtime increasing as $ES$ grows. Since the charts show no clear saturation point in $W_D$, we must also inspect the consistency of the learned diffusion function across configurations of $W_D$ and early-stopping patiences (Fig.~\ref{fig:res3}b).

Here, the consistency plots show that the predictions for the different widths agree closely, with the degree of agreement across $ES$ depending on the form of the true diffusion function. 
For linear diffusion, $W_D=2$ and $W_D=8$ result in near-identical predictions that are stable across $ES$. The $W_D=4$ predictions initially deviate from the $W_D=2$ and $8$ predictions outside the central $90\%$ density support but become consistent at $ES=5000$. Hence, $W_D=2$ and $ES=2000$ are favoured.

For quadratic diffusion, the predictions are predominantly stable across all tested $(W_D,ES)$ configurations, so the cheapest setting, $W_D=2$ at $ES=2000$, is preferred.
For exponential diffusion, the predictions stabilise across $W_D$ values for $ES\geq 3000$.
Taken together, consistency selects ($W_D=2$, $ES=3000$) as the smallest configuration at which the learned functions have converged across all three diffusion forms. Although the validation loss keeps falling at larger settings, the recovered function no longer changes significantly, while the runtime continues to rise. Furthermore, the ground truth confirms that this consistency-based selection is accurate: the recovered functions match the true diffusion within $5\%$ MPE across all three forms (Fig.~\ref{fig:A2}, Appendix~\ref{app:case2-mse}).

 \subsection{Case~III: selecting early stopping for coupled functions (diffusion and growth in \texorpdfstring{$\boldsymbol{1\mathrm{D}{+}t}$}{1D+t})}
\label{sec:res-case-3}

We next apply the workflow to a PDE with two RHS terms, diffusion and growth, so that hyperparameter selection must account for learning both functions at once. Specifically, we consider constant diffusion ($D_1$) coupled with one of constant ($G_1$), linear ($G_2$), quadratic ($G_3$), or exponential ($G_4$) growth (Table~\ref{tab:expressions}). The data are of comparable complexity to Cases~I and~II, so the same $W_u = 64$ remains appropriate. The growth functions are no more complex than the diffusion functions of Case~II, so the RHS widths carry over as well, giving $W_D = W_G = 2$. In Appendix~\ref{app:case3-mse} (Fig.~\ref{fig:A3}a,b), we show that increasing $W_D=W_G$ beyond $2$ does not consistently improve performance across growth functions. We can therefore enter the pipeline directly at Step~4 to select early stopping. As in Case~II, the smallest tested value is the one selected in Case~I, giving $ES\in\{2000,3000,5000\}$.

\begin{figure}[h]
\centering

\setlength{\rowheaderwidth}{0.018\textwidth}
\setlength{\fourcolwidth}{0.228\textwidth}

\setlength{\boxheadertopwidth}{0.86\textwidth}
\setlength{\boxheaderbottomwidth}{0.86\textwidth}

\newcommand{\rowheaderfont}{\tiny}
\newcommand{\colheaderfont}{\scriptsize}
\newcommand{\figheaderfont}{\scriptsize}

\newcommand{\rowheaderpanel}[1]{%
\begin{minipage}[c]{\rowheaderwidth}
\centering
\rotatebox{90}{\rowheaderfont\textbf{#1}}
\end{minipage}%
}

\begin{tikzpicture}

\matrix (m) [
    matrix of nodes,
    column sep=4pt,
    row sep=13pt,
    nodes={
      inner sep=0pt,
      outer sep=0pt,
      anchor=center
    }
]{

    \rowheaderpanel{Diffusion}
    &
    \begin{minipage}[c]{\fourcolwidth}
      \centering
      \vspace*{1cm}
      \cornerlabelimgshift[width=\linewidth]{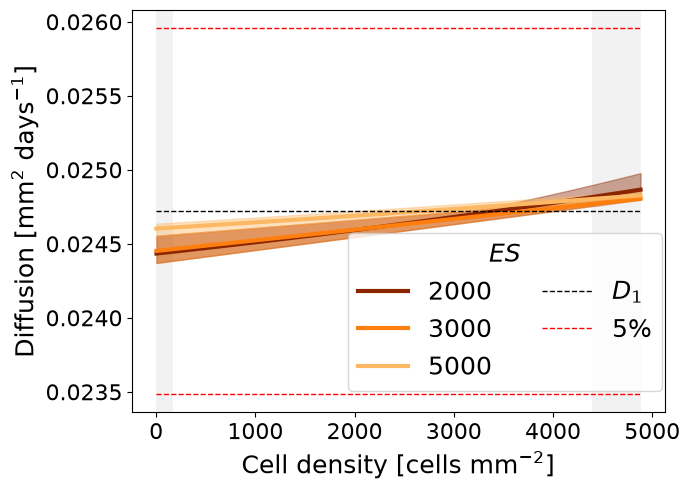}{}{north}{xshift=5pt,yshift=-5pt}
    \end{minipage}
    &
    \begin{minipage}[c]{\fourcolwidth}
      \centering
      \vspace*{1cm}
      \cornerlabelimgshift[width=\linewidth]
        {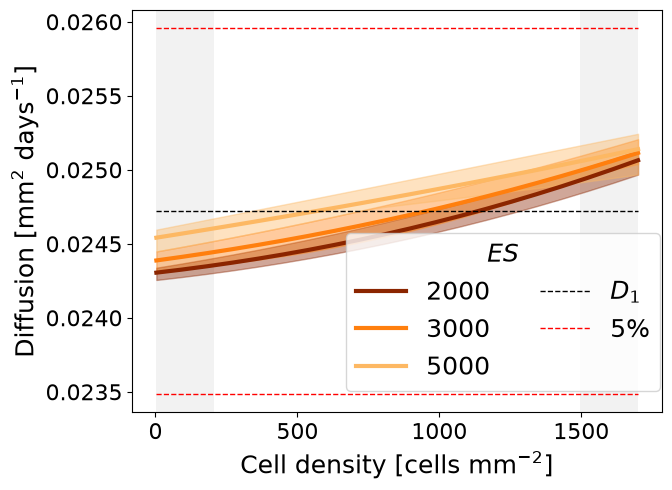}
        {}{north}{xshift=5pt,yshift=-5pt}
    \end{minipage}
    &
    \begin{minipage}[c]{\fourcolwidth}
      \centering
      \vspace*{1cm}
      \cornerlabelimgshift[width=\linewidth]
        {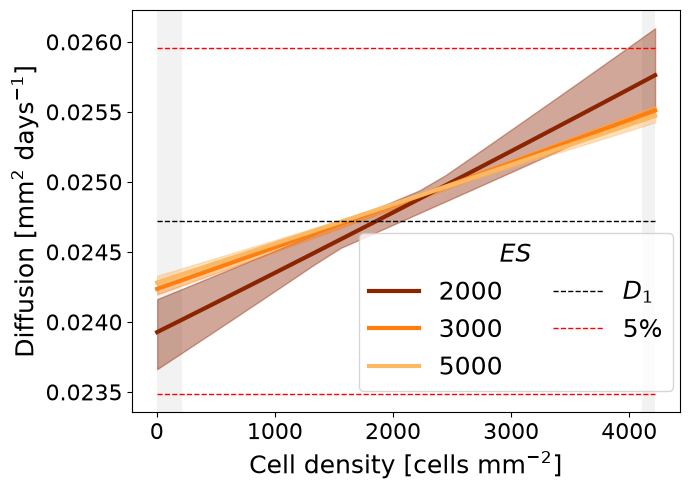}
        {}{north}{xshift=5pt,yshift=-5pt}
    \end{minipage}
    &
    \begin{minipage}[c]{\fourcolwidth}
      \centering
      \vspace*{1cm}
      \cornerlabelimgshift[width=\linewidth]
        {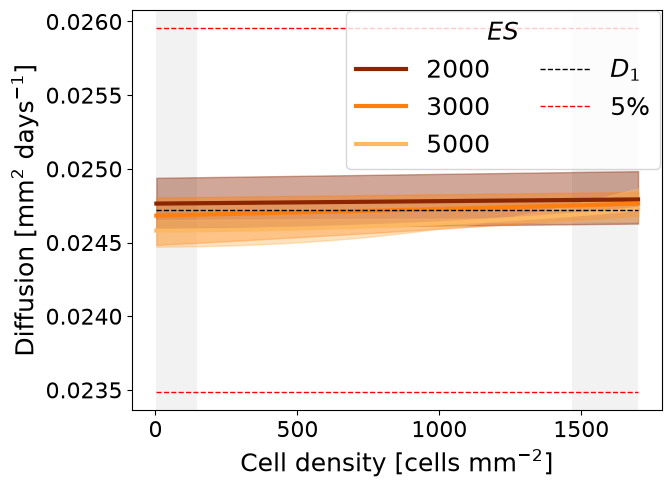}
        {}{north}{xshift=5pt,yshift=-5pt}
    \end{minipage}
    \\

    \rowheaderpanel{Growth}
    &
    \begin{minipage}[c]{\fourcolwidth}
      \centering
      \vspace*{-0.2cm}
      \cornerlabelimgshift[width=\linewidth]
        {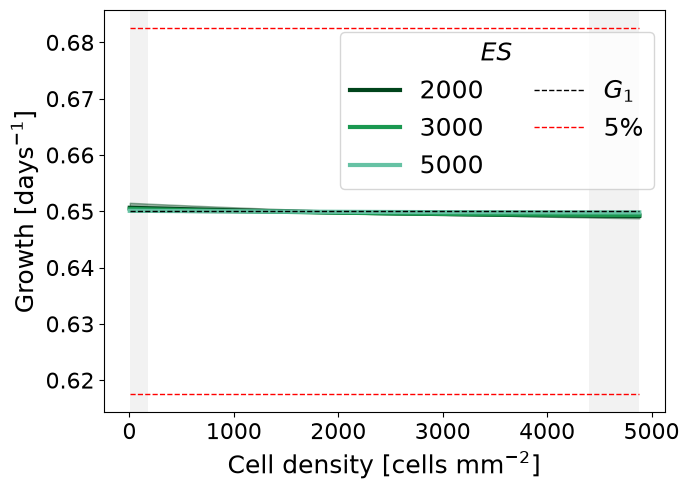}
        {}{north}{xshift=5pt,yshift=-5pt}
    \end{minipage}
    &
    \begin{minipage}[c]{\fourcolwidth}
      \centering
      \vspace*{-0.2cm}
      \cornerlabelimgshift[width=\linewidth]
        {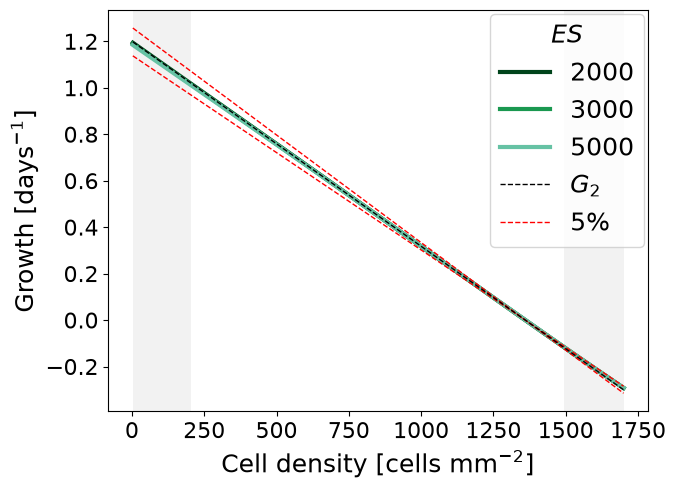}
        {}{north}{xshift=5pt,yshift=-5pt}
    \end{minipage}
    &
    \begin{minipage}[c]{\fourcolwidth}
      \centering
      \vspace*{-0.2cm}
      \cornerlabelimgshift[width=\linewidth]
        {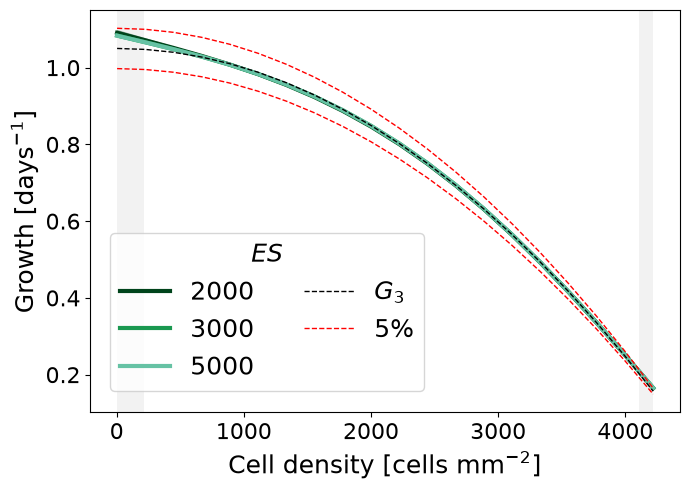}
        {}{north}{xshift=5pt,yshift=-5pt}
    \end{minipage}
    &
    \begin{minipage}[c]{\fourcolwidth}
      \centering
      \vspace*{-0.2cm}
      \cornerlabelimgshift[width=\linewidth]
        {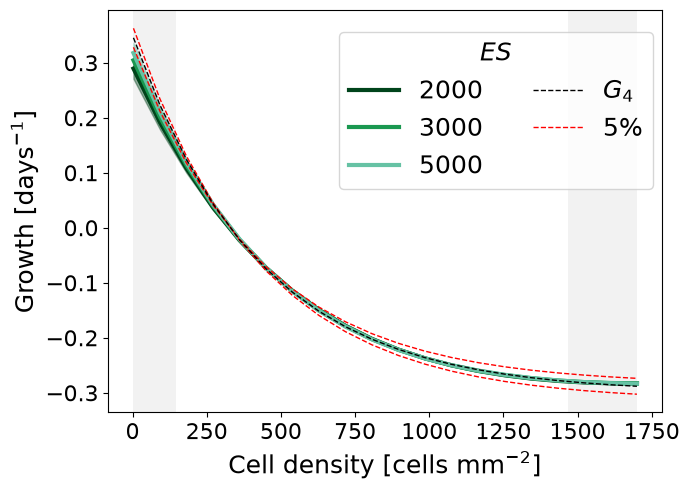}
        {}{north}{xshift=5pt,yshift=-5pt}
    \end{minipage}
    \\
};

\coordinate (outerNW) at ([xshift=-5pt,yshift=7pt]m.north west);
\coordinate (outerSE) at ([xshift= 5pt,yshift=-5pt]m.south east);

\draw[
    lightgray,
    line width=1pt,
    rounded corners=5pt
]
(outerNW) rectangle (outerSE);

\coordinate (bottomHeaderC) at
  ($([xshift=-2pt]m-1-2.north west)!0.5!([xshift=2pt]m-1-5.north east)$);

\node[
    fill=white,
    text=black,
    font=\figheaderfont,
    anchor=north,
    inner xsep=1pt,
    inner ysep=0.5pt,
    align=center,
    text width=\boxheaderbottomwidth
] at ([yshift=0pt]bottomHeaderC)
{\textbf{Learned diffusion and growth functions}};

\node[
    fill=white,
    text=black,
    font=\colheaderfont,
    anchor=north,
    inner xsep=1pt,
    inner ysep=0.5pt,
    align=center
] at ([xshift=0.15cm,yshift=-16pt]m-1-2.north)
{Constant growth};

\node[
    fill=white,
    text=black,
    font=\colheaderfont,
    anchor=north,
    inner xsep=1pt,
    inner ysep=0.5pt,
    align=center
] at ([xshift=0.15cm,yshift=-16pt]m-1-3.north)
{Linear growth};

\node[
    fill=white,
    text=black,
    font=\colheaderfont,
    anchor=north,
    inner xsep=1pt,
    inner ysep=0.5pt,
    align=center
] at ([xshift=0.15cm,yshift=-16pt]m-1-4.north)
{Quadratic growth};

\node[
    fill=white,
    text=black,
    font=\colheaderfont,
    anchor=north,
    inner xsep=1pt,
    inner ysep=0.5pt,
    align=center
] at ([xshift=0.15cm,yshift=-16pt]m-1-5.north)
{Exponential growth};

\end{tikzpicture}

\vspace{-0.2cm}
\caption{\textbf{Selecting hyperparameters under coupled diffusion and growth.} Data are generated from constant diffusion ($D_1$) paired with four growth forms---constant ($G_1$), linear ($G_2$), quadratic ($G_3$), and exponential ($G_4$).
Consistency of the learned diffusion and growth functions across early-stopping patiences $ES \in \{2000,3000, 5000\}$ with fixed $W_u=64$ and $W_D=W_G=2$ is shown. 
Filled bands show min--max ranges across $3$ repeats with different training--validation splits.
Regions
outside the central $90\%$ of the density support are shown in grey.
The dashed ground-truth line and $\pm5\%$ bands play no role in the selection workflow and are shown only for reference. }
\label{fig:res4}
\end{figure}

Also, as in Case~II, the cost-performance trade-off alone does not pin down a preferred $ES$ (Fig.~\ref{fig:A4}, Appendix~\ref{app:case3-mse}), so we turn to the consistency plots (Fig.~\ref{fig:res4}). With two coupled functions to recover, neither can be judged alone because an $ES$ at which one function appears converged may still leave the other changing. Consistency must therefore be read on $D(u)$ and $G(u)$ simultaneously. Read jointly (Fig.~\ref{fig:res4}), these RHS functions are generally consistent across all four combinations and all $ES$. By $ES=2000$, the learned RHS functions have largely stabilised, except for quadratic growth, where the diffusion prediction only becomes consistent from $ES = 3000$ onwards. We therefore adopt the more conservative $ES = 3000$: the larger patience guards against premature stopping, at little additional cost for the constant, linear and exponential growth functions, where it is not needed. This selection is supported by comparing the learned diffusion and growth functions against the ground truth, as shown visually in Fig.~\ref{fig:res4} and quantitatively in Appendix~\ref{app:case3-mse}.

\subsection{Case~IV: selecting \texorpdfstring{$ES$}{ES} on noisy data (diffusion and growth in \texorpdfstring{$\boldsymbol{2\mathrm{D}{+}t}$}{2D+t})}
\label{sec:res-case-4}

Finally, we ask whether the workflow holds up under noisy and higher-dimensional data. To this end, we apply it to $2\text{D}{+}t$ data generated with constant diffusion ($D_1$) and linear growth ($G_2$), corrupted by $5\%$ MPE noise with $\gamma = 1$.
Moving to $2\text{D}{+}t$, we carry the density MLP width forward at $W_u = 64$ and re-run the pipeline over the RHS widths only, testing $W_D = W_G \in \{2, 4, 8, 16\}$. We select $W_D = W_G = 4$ as the smallest width whose learned $\hat{D}(u)$ and $\hat{G}(u)$ agree with those of the next width tested, $W_D = W_G = 8$, while $W_D = W_G = 2$ and $16$ each deviate significantly (Fig.~\ref{fig:A5}, Appendix~\ref{app:case4-mse}). We enter the pipeline at Step~4 to select $ES$ from $\{500,1000,2000,3000\}$. 
Here, the smallest tested $ES$ value is lower than in the previous cases. This is because we expect noise to make the validation loss plateau earlier. The PDE regularisation discourages the network from fitting the noise, so it recovers only a smoothed density fit, and once that fit is reached, which happens sooner at higher noise levels, the validation loss stops improving.

\begin{figure}[h]
\centering
\setlength{\panelwidthtwoD}{0.2275\textwidth}
\setlength{\boxheaderleftwidthtwoD}{0.44\textwidth}
\setlength{\boxheaderrightwidthtwoD}{0.44\textwidth}
\newcommand{\figheaderfonttwoD}{\scriptsize}
\newcommand{\plainpaneltwoD}[1]{%
\begin{minipage}[t]{\panelwidthtwoD}
\centering
\cornerlabelimgshift[width=\linewidth]
  {#1}{}{north}{xshift=5pt,yshift=-5pt}
\end{minipage}%
}
\begin{tikzpicture}
\matrix (m) [
    matrix of nodes,
    column sep=8pt,
    row sep=0pt,
    nodes={inner sep=0pt, outer sep=0pt, anchor=center}
]{
    \plainpaneltwoD{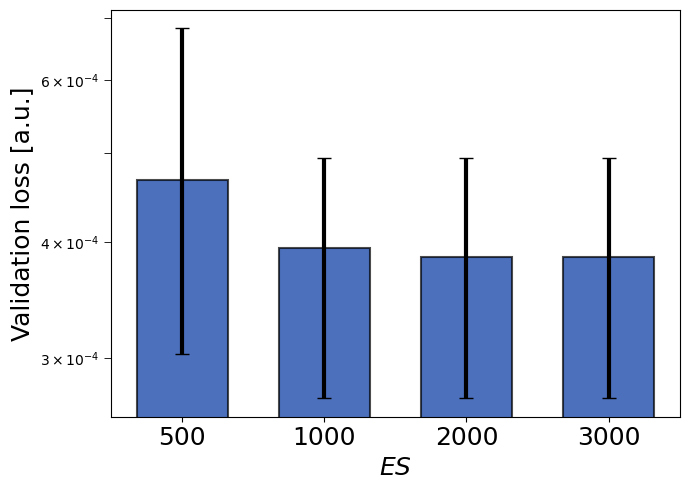}
    &

    \plainpaneltwoD{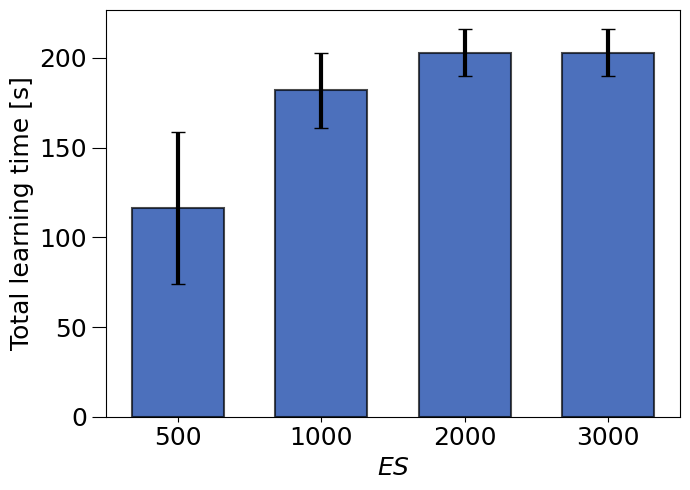}
    &

    \plainpaneltwoD{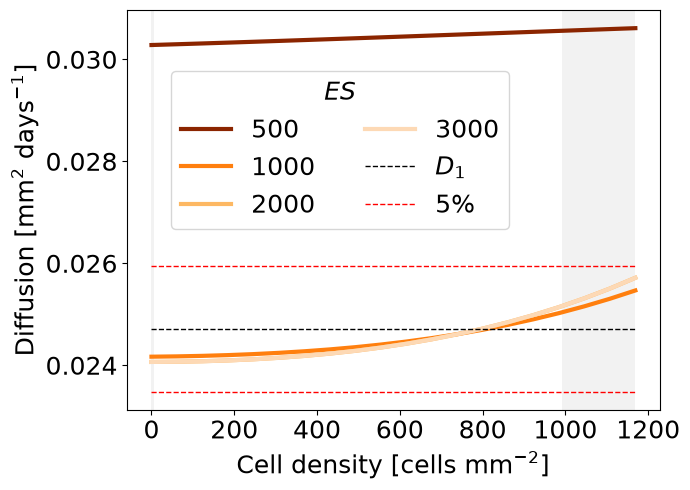}
    &
    \plainpaneltwoD{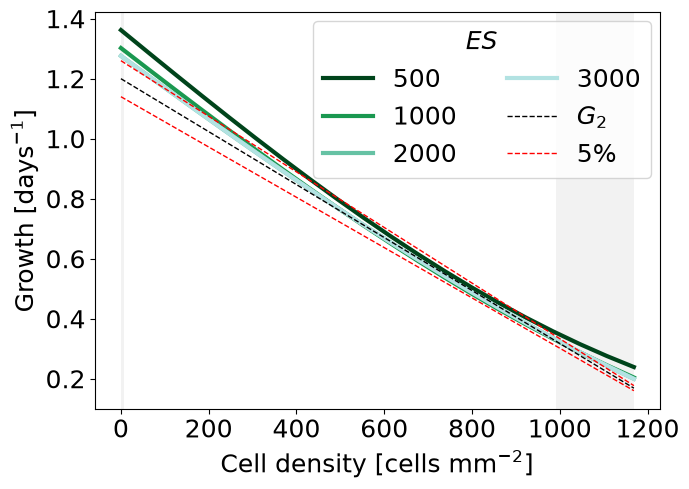}
    \\
};
\coordinate (boxNW) at ([xshift=-4pt,yshift=18pt]m.north west);
\coordinate (boxSE) at ([xshift= 4pt,yshift= 2pt]m.south east);
\draw[
    lightgray,
    line width=1pt,
    rounded corners=5pt
]
(boxNW) rectangle (boxSE);
\coordinate (vDiv) at ($(m-1-2.east)!0.5!(m-1-3.west)$);
\draw[
    lightgray,
    line width=1pt
]
(vDiv |- boxNW) -- (vDiv |- boxSE);
\coordinate (leftHeaderC) at
  ($([xshift=-2pt]m-1-1.north west)!0.5!([xshift=2pt]m-1-2.north east)$);
\coordinate (rightHeaderC) at
  ($([xshift=-2pt]m-1-3.north west)!0.5!([xshift=2pt]m-1-4.north east)$);
\node[
    fill=white,
    text=black,
    font=\figheaderfonttwoD,
    anchor=north,
    inner xsep=1pt,
    inner ysep=0.5pt,
    align=center,
    text width=\boxheaderleftwidthtwoD
] at ([yshift=15pt]leftHeaderC)
{\textbf{(a)} Cost-performance across $ES$};
\node[
    fill=white,
    text=black,
    font=\figheaderfonttwoD,
    anchor=north,
    inner xsep=1pt,
    inner ysep=0.5pt,
    align=center,
    text width=\boxheaderrightwidthtwoD
] at ([yshift=15pt]rightHeaderC)
{\textbf{(b)} Learned functions across $ES$};
\end{tikzpicture}
\vspace{-0.6cm}
\caption{\textbf{Selecting the early-stopping patience on noisy $\boldsymbol{2\mathrm{D}{+}t}$ data.}
Data generated from
constant diffusion ($D_1$) and linear growth ($G_2$), with $5\%$ MPE noise added, at fixed widths ($W_u=64$, $W_D=W_G=4$). 
\textbf{(a)} Cost-performance trade-off between validation loss and runtime across the early-stopping patience ($ES$).
Bars and whiskers respectively show mean values and min--max ranges across $3$ repeats with different training--validation (TV) splits.
\textbf{(b)} Consistency of the learned diffusion and growth functions, $\hat D(u)$ and $\hat G(u)$, across varying $ES$. 
Regions outside the central $90\%$ of the density support are shown in grey. The dashed ground-truth line and $\pm5\%$ bands play no role in the selection workflow and are shown only for reference.} 
\label{fig:res5}
\end{figure}

In two spatial dimensions we use a coarse spatial grid to keep training computationally tractable. Even so, compared with the equivalent $1\text{D}{+}t$ problem in Case~III, the data volume grows by a factor of $5.7$ (Table~\ref{tab:datagen}), making this a data-richer but noisy regime. Despite the added noise, the cost-performance trade-off alone can motivate the choice of $ES$.
The validation loss drops between $ES=500$ and $1000$ and then plateaus (Fig.~\ref{fig:res5}a), motivating $ES=1000$. This plateau is also reflected in the learned diffusion and growth functions (Fig.~\ref{fig:res5}b), confirming the choice. Furthermore, MSE plots against the ground-truth functions $D_1(u)$ and $G_2(u)$ support this selection and, by extension, the workflow's applicability in noisy $2\text{D}{+}t$ settings (Fig.~\ref{fig:A6}, Appendix~\ref{app:case4-mse}).

\section{Discussion}
We have presented a diagnostic workflow that selects BINN hyperparameters from training-time signals alone, with no access to the ground truth. We applied the workflow to four synthetic systems of increasing complexity. In the noise-free cases, the validation loss and the consistency of the learned functions selected hyperparameters whose recovered diffusion and growth functions came within $5\%$ of the ground truth, at reasonable training cost. Because these tests are carried out in a controlled setting where the truth is known, they lend credence not only to the workflow but to BINNs themselves.

The workflow yields four transferable rules of thumb, each grounded in the case studies. 
First, the state (density) MLP should be made substantially wider than the RHS MLPs,
since the density MLP must capture the full spatiotemporal dynamics of the data,
whereas the RHS MLPs need only represent the comparatively simple coefficient functions considered here.
Second, keep the RHS MLPs narrow and scale their widths with the expected complexity of each term: $W_D=2$ was sufficient from non-constant diffusion through coupled diffusion and growth (Cases~II--III, Figs.~\ref{fig:res3}--\ref{fig:res4}), increasing only modestly to $W_D=W_G=4$ in the higher-dimensional $2\text{D}{+}t$ setting (Case~IV, Fig.~\ref{fig:res5}). 
Third, capacity is better spent on width than on depth:
three hidden layers sufficed for all networks, with the validation loss and recovered functions largely insensitive to further depth,
consistent with the simple targets of the RHS (Case~I, Fig.~\ref{fig:depth}). 
Fourth, the early-stopping patience should be selected using both the validation-loss curve and the consistency of the learned terms,
judging all coupled terms simultaneously (Cases~III and~IV, Figs.~\ref{fig:res4} and~\ref{fig:res5}). 
Underpinning all four rules of thumb, the trend in the validation loss, which needs no ground truth, tracks the otherwise-unobservable joint error in the recovered functions (Case~I, Figs.~\ref{fig:res1} and~\ref{fig:res2}),
and this is what makes these ground-truth-free selections trustworthy.

Recent work by 
\textcite{crossley_reliable_2026}
studies how architecture and optimisation choices affect BINN performance on 
$1\text{D}{+}t$ advection--diffusion--reaction models, 
finding that moderately expressive architectures, intermediate learning rates, and balanced data and PDE losses
are key to reliable recovery of the RHS functions. 
Whereas their study surveys the influence of individual optimisation hyperparameters, 
our contribution is an architecture-ordered workflow for jointly selecting network capacity and early-stopping patience, extended here to noisy two-dimensional data. 
The two studies' conclusions agree in the areas where they overlap, namely network capacity and loss weighting (the latter considered in our Supplementary Material~\ref{sec:SM:pde_data_loss}--\ref{sec:SM:bio_loss}).

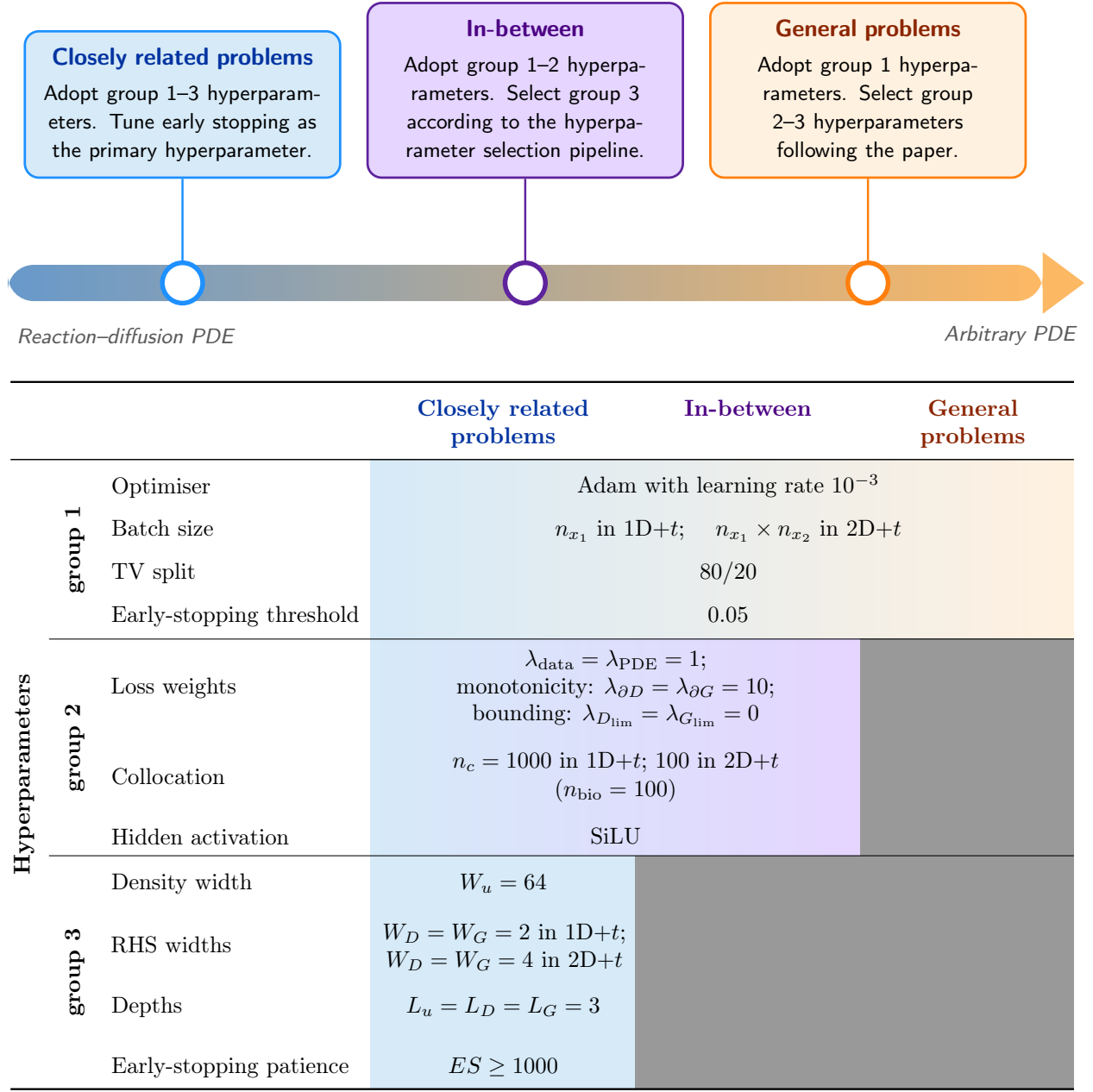
\begin{figure}[H]
\centering
\resizebox{\textwidth}{!}{%
\begin{tikzpicture}[font=\sffamily]
\def\xL{0.3}
\def\xR{15.7}
\def\xa{2.9}
\def\xb{8.0}
\def\xc{13.1}
\def\ycard{1.55}
\begin{scope}
  \clip[rounded corners=11pt] (\xL,-0.27) rectangle (\xR,0.27);
  \shade[left color=blue3, middle color=purple4, right color=orange3]
        (\xL,-0.27) rectangle (\xR,0.27);
\end{scope}
\fill[orange3] (\xR,-0.46) -- (\xR+0.62,0) -- (\xR,0.46) -- cycle;
\node[anchor=west, text=gray!70!black, font=\sffamily\footnotesize\itshape]
      at (\xL,-0.78) {\shortstack[l]{Reaction--diffusion PDE}};
\node[anchor=east, text=gray!70!black,
      font=\sffamily\footnotesize\itshape]
      at (\xR+0.62,-0.78)
      {\shortstack[r]{Arbitrary PDE}};
\draw[blue2, line width=0.9pt] (\xa,0.27) -- (\xa,\ycard);
\filldraw[fill=white, draw=blue2, line width=2pt] (\xa,0) circle (0.30);
\node[draw=blue2, fill=blue5, rounded corners=6pt, line width=1pt,
      text width=4.2cm, align=center, inner sep=7pt, anchor=south,
      font=\sffamily\small] at (\xa,\ycard)
  {\textbf{\textcolor{blue1}{Closely related problems}}\\[3pt]
   \footnotesize Adopt group 1--3 hyperparameters. Tune early stopping as the primary hyperparameter.};
\draw[purple3, line width=0.9pt] (\xb,0.27) -- (\xb,\ycard);
\filldraw[fill=white, draw=purple3, line width=2pt] (\xb,0) circle (0.30);
\node[draw=purple3, fill=purple5, rounded corners=6pt, line width=1pt,
      text width=4.2cm, align=center, inner sep=7pt, anchor=south,
      font=\sffamily\small] at (\xb,\ycard)
  {\textbf{\textcolor{purple1}{In-between}}\\[3pt]
   \footnotesize Adopt group 1--2 hyperparameters. Select group 3 according to the hyperparameter selection pipeline.};
\draw[orange2, line width=0.9pt] (\xc,0.27) -- (\xc,\ycard);
\filldraw[fill=white, draw=orange2, line width=2pt] (\xc,0) circle (0.30);
\node[draw=orange2, fill=orange5, rounded corners=6pt, line width=1pt,
      text width=4.2cm, align=center, inner sep=7pt, anchor=south,
      font=\sffamily\small] at (\xc,\ycard)
  {\textbf{\textcolor{orange1}{General problems}}\\[3pt]
   \footnotesize Adopt group 1 hyperparameters. Select group 2--3 hyperparameters following the paper.};
\end{tikzpicture}%
}

\vspace{1em}

\centering
\small
\setlength{\tabcolsep}{5pt}
\renewcommand{\arraystretch}{1.35}
\begin{NiceTabular}{@{}c c l >{\centering\arraybackslash}p{3.7cm} >{\centering\arraybackslash}p{3.1cm} >{\centering\arraybackslash}p{3.1cm}@{}}
\CodeBefore
  \begin{tikzpicture}
    \shade[left color=blue5, middle color=purple5, right color=orange5] (2-|4) rectangle (6-|7);
    \shade[left color=blue5, right color=purple5] (6-|4) rectangle (9-|6);
    \fill[gray!80] (6-|6) rectangle (9-|7);
    \shade[left color=blue5, right color=blue5] (9-|4) rectangle (13-|5);
    \fill[gray!80] (9-|5) rectangle (13-|7);
  \end{tikzpicture}
\Body
\toprule
 & & & \textcolor{blue1}{\textbf{Closely related problems}}
     & \textcolor{purple1}{\textbf{In-between}}
     & \textcolor{orange1}{\textbf{General problems}} \\
\midrule
\Block{11-1}{\rotate \textbf{Hyperparameters}}
 & \Block{4-1}{\rotate \textbf{group 1}}
 & Optimiser        & \Block{1-3}{Adam with learning rate $10^{-3}$} \\[0.25em] 
 & & Batch size     & \Block{1-3}{$n_{x_1}$ in $1\text{D}{+}t$; $\quad n_{x_1} \times n_{x_2}$ in $2\text{D}{+}t$} \\[0.25em]
 & & TV split       & \Block{1-3}{$80/20$} \\[0.25em]
 & & Early-stopping threshold & \Block{1-3}{$0.05$} \\[0.25em]
 & \Block{3-1}{\rotate \textbf{group 2}}
 & \Block[l]{1-1}{\rule{0pt}{1.2em}Loss weights}     & \Block{1-2}{\shortstack{\rule{0pt}{1.2em}$\lambda_{\text{data}}=\lambda_{\text{PDE}}=1$;\\
                                 monotonicity: $\lambda_{\partial D}=\lambda_{\partial G}=10$;\\
                                 bounding: $\lambda_{D_{\lim}}=\lambda_{G_{\lim}}=0$}} &
                    & \Block{3-1}{} \\[2.25em]
 & & Collocation    & \Block{1-2}{\shortstack{$n_c=1000$ in $1\text{D}{+}t$; $100$ in $2\text{D}{+}t$\\ ($n_{\text{bio}}=100$)}} \\[1em]
 & & Hidden activation & \Block{1-2}{SiLU} \\
 & \Block{4-1}{\rotate \textbf{group 3}}
 & \rule{0pt}{1.5em}Density width    & $W_u=64$ & \Block{4-2}{} \\[1em]
 & & \Block[l]{1-1}{\rule{0pt}{0em}RHS widths}
   & \Block{1-1}{\shortstack{$W_D=W_G=2$ in $1\text{D}{+}t$;\\ $W_D=W_G=4$ in $2\text{D}{+}t$}} \\[1em]
 & & Depths         & $L_u=L_D=L_G=3$ \\[1em]
 & & Early-stopping patience    & $ES\geq1000$ \\
\bottomrule
\CodeAfter
  \begin{tikzpicture}
    \draw[black, line width=0.4pt] (6-|2) -- (6-|7);
    \draw[black, line width=0.4pt] (9-|2) -- (9-|7);
  \end{tikzpicture}
\end{NiceTabular}
\caption{\textbf{A user guide to hyperparameter selection.}
 \textit{Top}. Which hyperparameters a researcher is recommended to reuse from the table below, and how the remaining ones are obtained, depend on how closely their problem matches the reaction--diffusion setting.
\textit{Bottom}. Hyperparameters categorised by groups and problem; grey cells mark hyperparameters that must be re-derived on the user's own data.}
\label{fig:combined_researcher_groups}
\end{figure}

We anticipate that a range of researchers looking to apply BINNs will find this work useful. 
Recall from Section~\ref{sec:methods_intro} that hyperparameters are grouped by how they are set: 
group~1 consists of generic neural-network settings typically set at standard defaults, 
group~2 consists of BINN-specific settings typically set from domain knowledge and practical considerations, 
and group~3 consists of the problem-specific hyperparameters (network capacities and early-stopping patience) that the workflow in Fig.~\ref{fig:pipeline_flow} tunes.
How the hyperparameters from this work can be reused depends on how closely the considered problem resembles the reaction--diffusion setting studied here: (i) researchers with a closely related problem (reaction--diffusion dynamics on $1\text{D}{+}t$ or $2\text{D}{+}t$ data) can adopt the values reported here directly; 
(ii) those with an in-between problem (a PDE whose RHS terms are of comparable functional complexity to those studied here) re-derive the group~3 hyperparameters on their own data while reusing those from group~2; and (iii) researchers tackling a new problem (a PDE that differs substantially from the setting studied here) re-derive both group~2 and group~3 hyperparameters via an analogous study. 
Figure~\ref{fig:combined_researcher_groups} summarises these recommendations and lists the hyperparameter values.

Our workflow is deliberately geared towards practical use on real problems, and this pragmatic focus comes with limitations. 
Because a full sweep over all hyperparameters would be computationally infeasible, the workflow adopts a semi-sequential approach: it selects the state MLP width, then the RHS MLP widths, then the early-stopping patience, following the network's data flow and focusing on widths rather than examining the MLP depths in comparable detail. 
Although earlier selections are re-checked as the downstream hyperparameters are chosen, this semi-sequential approach may miss well-performing hyperparameter configurations that a full joint optimisation would uncover. 
A further practical aspect is that some practitioner judgement is inevitable in evaluating workflow outputs.
No fixed quantitative thresholds would transfer across application problems, and the diagnostics are therefore read qualitatively, by identifying where a cost-performance curve saturates or where the learned functions stop changing.

A limitation of the study itself is that we validate the workflow against synthetic ground truth rather than experimental data, and at only modest noise levels (up to $5\%$).
This lets us assess the recovered functions against a known reference, but it leaves open how the procedure behaves under the additional confounders of real measurements, including higher noise levels and model misspecification.
However, to demonstrate applicability, we apply the workflow to real $2\text{D}{+}t$ microscopy data in a concurrent study \cite{lavery_physics-informed_2026}. Future work could investigate the workflow's robustness to extensions of the BINN setting, including PDEs with additional or different RHS terms (for example, advection), and RHS terms that depend not only on density but also on covariates such as space, time, or background environmental factors.

The work presented in this study lowers the barrier to BINN-based equation learning by making hyperparameter selection for BINNs systematic, documented, and ground-truth-free, with its choices verified against synthetic ground truth. 
It thereby supports wider and more reproducible use of BINNs across biological modelling problems.

\clearpage
\newpage

\section*{Code}
Full code is available on GitHub:
\begin{center}
\url{https://github.com/williamlavery/binn-hyperparam-selection}
\end{center}
\section*{Acknowledgements}
JAC and SH were supported by the Kjell och M{\"a}rta Beijer Foundation. SH was supported by the Swedish Research Council (project 2024-05621).

\clearpage
\newpage 
\printbibliography

\setcounter{equation}{0}
\renewcommand{\theequation}{A.\arabic{equation}}
\renewcommand{\theHequation}{A.\arabic{equation}}
\setcounter{figure}{0}
\renewcommand{\thefigure}{A.\arabic{figure}}
\renewcommand{\theHfigure}{A.\arabic{figure}}
\clearpage
\newpage
\appendix
\section{Synthetic data scenarios and BINN architecture overview}
\label{app:data_gen}
\input{figures/Appendix_fig_1}
\newpage
\section{Ground-truth verification of the learned functions}
Comparisons between the predicted and true data-generating functions for case studies II--IV are provided below.
\subsection{Case~II: non-constant diffusion (\texorpdfstring{$\boldsymbol{1\mathrm{D}{+}t}$}{1D+t})}
\label{app:case2-mse}
\begin{figure}[h!]
\centering
\begin{minipage}{0.32\linewidth}
  \centering
  \cornerlabelimgshift[width=\linewidth]
    {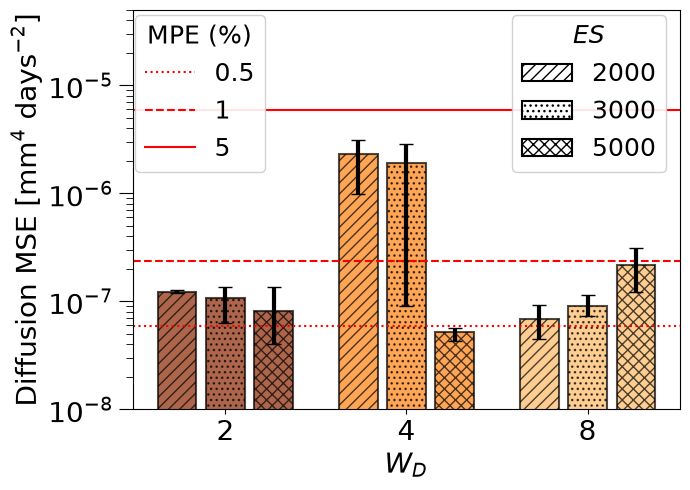}
    {(a)}{north}{xshift=5pt,yshift=-5pt}
\end{minipage}
\hfill
\begin{minipage}{0.32\linewidth}
  \centering
  \cornerlabelimgshift[width=\linewidth]
    {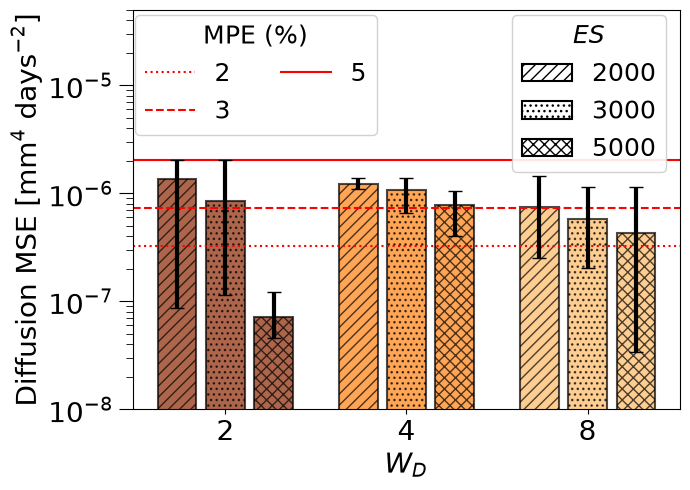}
    {(b)}{north}{xshift=15pt,yshift=-5pt}
\end{minipage}
\hfill
\begin{minipage}{0.32\linewidth}
  \centering
  \cornerlabelimgshift[width=\linewidth]
    {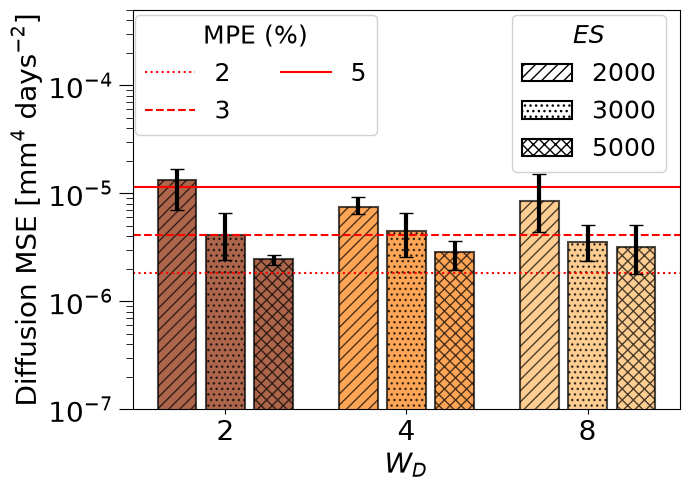}
    {(c)}{north}{xshift=5pt,yshift=-5pt}
\end{minipage}
\vspace{-0.2cm}
\caption{\textbf{Ground-truth verification for Case~II.} Mean-squared error (MSE) of the learned diffusion function $\hat D(u)$ with respect to the known forms, across the diffusion-network width ($W_D$) and the early-stopping patience ($ES$), one panel per form: \textbf{(a)} linear ($D_2$), \textbf{(b)} quadratic ($D_3$), \textbf{(c)} exponential ($D_4$).}
\label{fig:A2}
\end{figure}
\subsection{Case~III: coupled diffusion and growth
  (\texorpdfstring{$\boldsymbol{1\mathrm{D}{+}t}$}{1D+t})}
\label{app:case3-mse}
\begin{figure}[H]
\centering
\setlength{\twocolwidthDiag}{0.32\textwidth}
\begin{minipage}[c]{\twocolwidthDiag}
  \centering
  \cornerlabelimgshift[width=\linewidth]{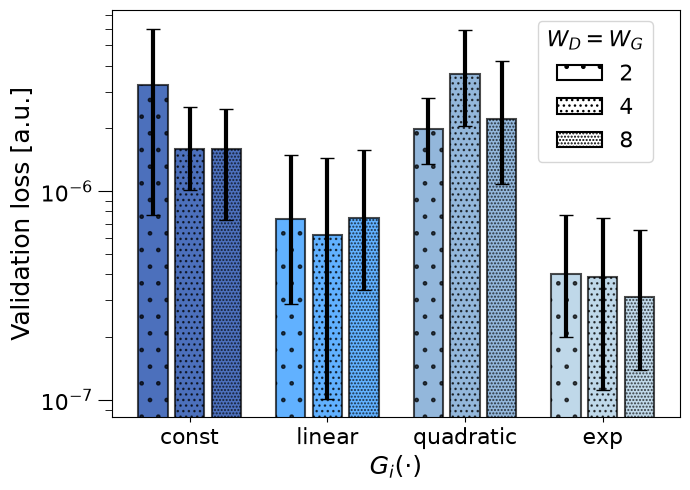}{(a)}{north}{xshift=10pt,yshift=10pt}
\end{minipage}
\hspace{0.4cm}
\begin{minipage}[c]{\twocolwidthDiag}
  \centering
  \cornerlabelimgshift[width=\linewidth]
    {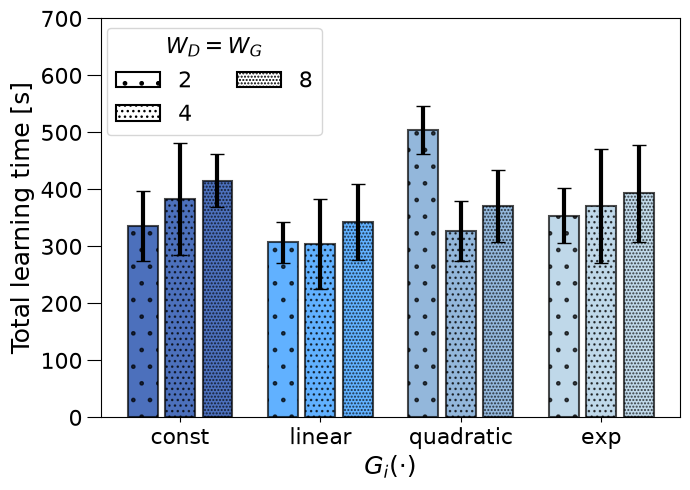}
    {(b)}{north}{xshift=10pt,yshift=10pt}
\end{minipage}
\vspace{1em}
\begin{minipage}[c]{\twocolwidthDiag}
  \centering
  \cornerlabelimgshift[width=\linewidth]
    {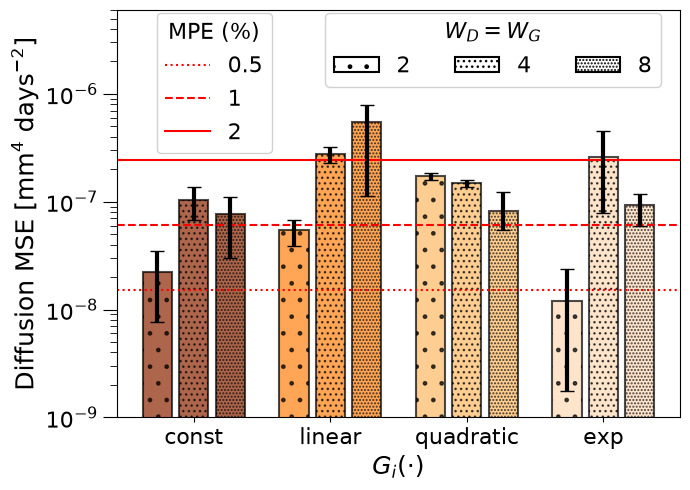}
    {(c)}{north}{xshift=10pt,yshift=10pt}
\end{minipage}
\hspace{0.4cm}
\begin{minipage}[c]{\twocolwidthDiag}
  \centering
  \cornerlabelimgshift[width=\linewidth]
    {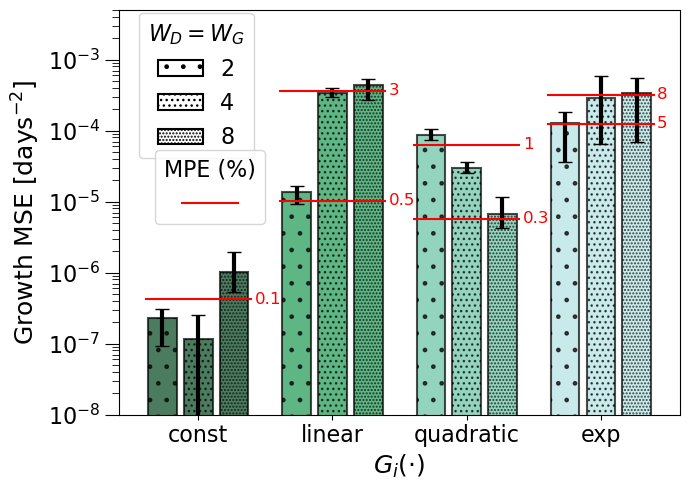}
    {(d)}{north}{xshift=10pt,yshift=10pt}
\end{minipage}
\vspace{-0.5cm}
\caption{
\textbf{Cost-performance and function-error diagnostics for Case~III across right-hand side (RHS) network widths.}
Constant diffusion ($D_1$) paired with four growth forms---constant ($G_1$), linear ($G_2$), quadratic ($G_3$), and exponential ($G_4$). Diffusion and growth network widths are always equal ($W_D = W_G$). Early-stopping patience is fixed at $3000$ epochs. 
Panel \textbf{(a)} shows the validation loss across the four ground-truth growth functions and the RHS network widths, \textbf{(b)} the corresponding runtimes, \textbf{(c)} the diffusion-function error, and \textbf{(d)} the growth-function error.}
\label{fig:A3}
\end{figure}

\begin{figure}[H]
\centering 
\setlength{\twocolwidthDiag}{0.32\textwidth}
\begin{minipage}[c]{\twocolwidthDiag}
  \centering
  \cornerlabelimgshift[width=\linewidth]{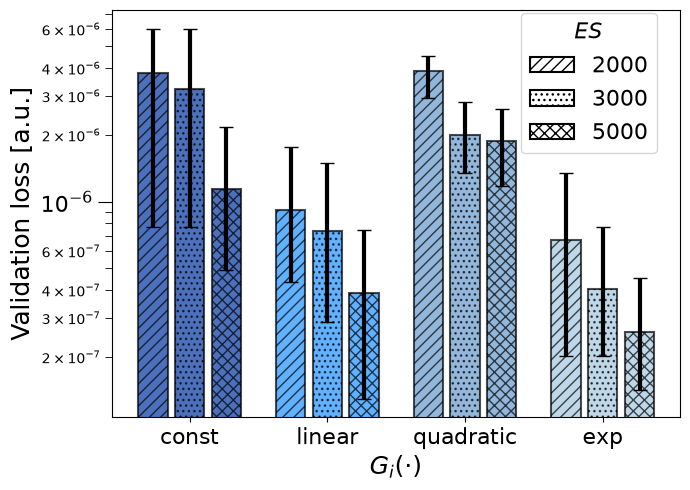}{(a)}{north}{xshift=10pt,yshift=10pt}
\end{minipage}
\hspace{0.4cm}
\begin{minipage}[c]{\twocolwidthDiag}
  \centering
  \cornerlabelimgshift[width=\linewidth]
    {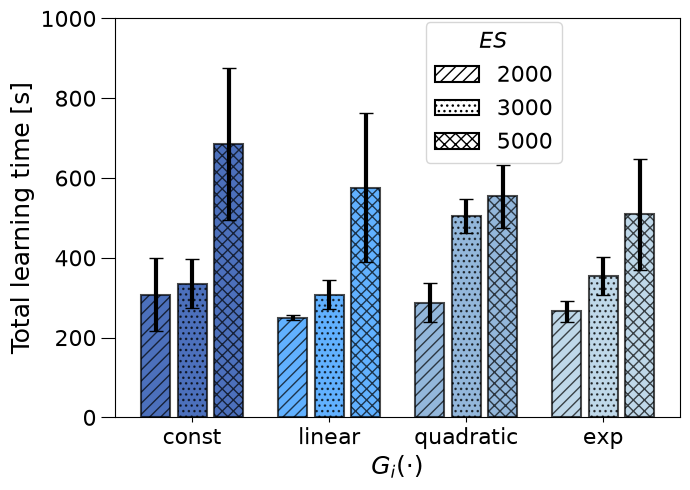}
    {(b)}{north}{xshift=10pt,yshift=10pt}
\end{minipage}
\vspace{1em}
\begin{minipage}[c]{\twocolwidthDiag}
  \centering
  \cornerlabelimgshift[width=\linewidth]
    {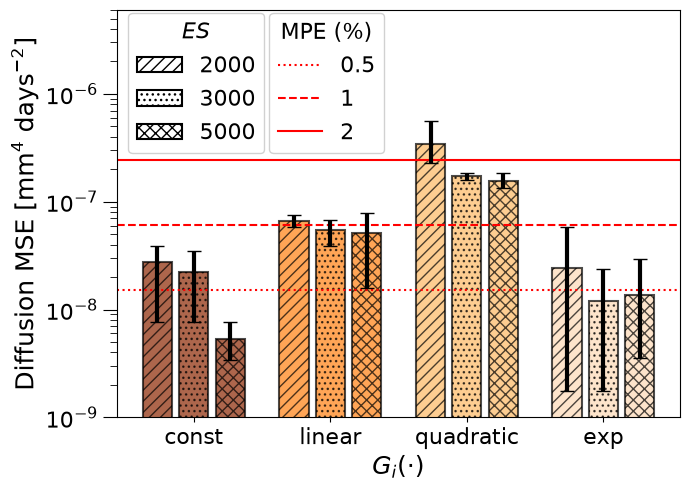}
    {(c)}{north}{xshift=10pt,yshift=10pt}
\end{minipage}
\hspace{0.4cm}
\begin{minipage}[c]{\twocolwidthDiag}
  \centering
  \cornerlabelimgshift[width=\linewidth]
    {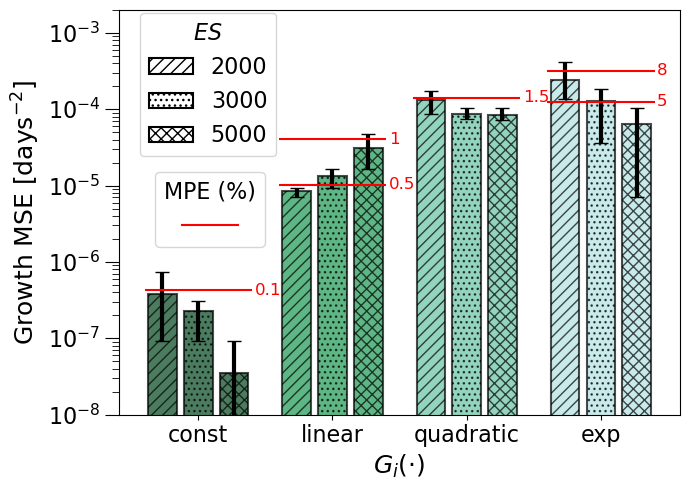}
    {(d)}{north}{xshift=10pt,yshift=10pt}
\end{minipage}
\vspace{-0.5cm}
\caption{\textbf{Cost-performance and function-error diagnostics for Case~III across early-stopping patiences.} Constant diffusion ($D_1$) paired with four growth forms---constant ($G_1$), linear ($G_2$), quadratic ($G_3$), and exponential ($G_4$). Early-stopping patience across $ES \in \{2000,3000,5000\}$, and right-hand side network widths $W_D = W_G = 2$.
\textbf{(a)} shows the validation loss across the four ground-truth growth functions and early-stopping patience, \textbf{(b)} the corresponding runtimes, \textbf{(c)} the diffusion-function error, and \textbf{(d)} the growth-function error.}
\label{fig:A4}
\end{figure}

\subsection{Case~IV: coupled diffusion and growth with noisy data (\texorpdfstring{$\boldsymbol{2\mathrm{D}{+}t}$}{2D+t})}
\label{app:case4-mse}
\begin{figure}[h]
\centering

\setlength{\panelwidthtwoD}{0.31\textwidth}
\setlength{\boxheaderawidthtwoD}{0.31\textwidth}
\setlength{\boxheaderbwidthtwoD}{0.64\textwidth}

\newcommand{\figheaderfonttwoD}{\scriptsize}

\newcommand{\plainpaneltwoD}[1]{%
\begin{minipage}[t]{\panelwidthtwoD}
\centering
\cornerlabelimgshift[width=\linewidth]
{#1}{}{north}{xshift=5pt,yshift=-5pt}
\end{minipage}%
}

\begin{tikzpicture}

\matrix (m) [
matrix of nodes,
column sep=8pt,
row sep=0pt,
nodes={inner sep=0pt, outer sep=0pt, anchor=center}
]{
\plainpaneltwoD{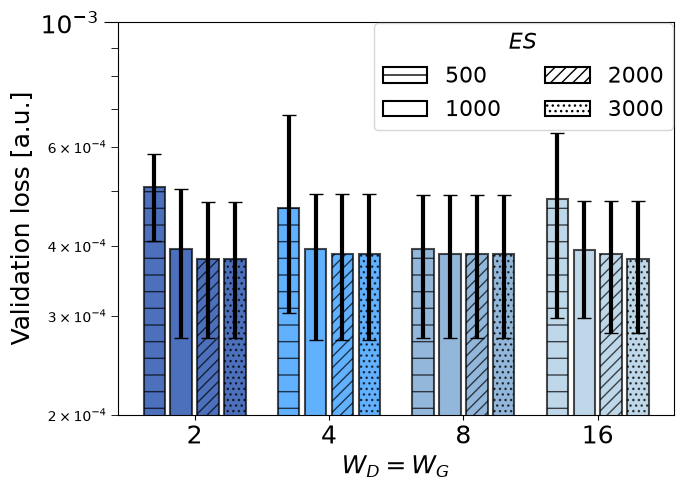}
&
\plainpaneltwoD{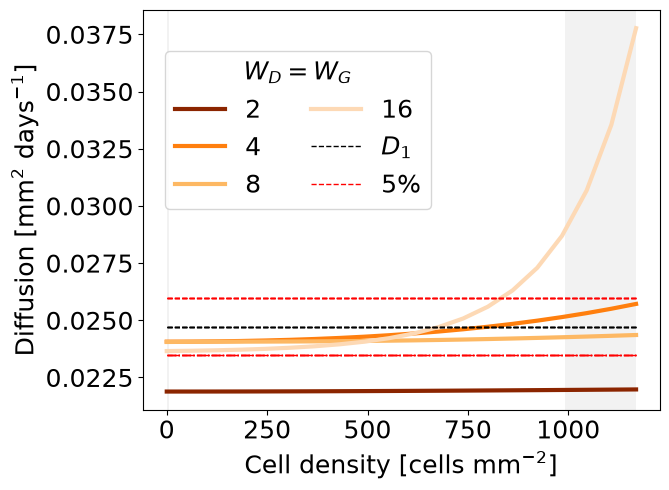}
&
\plainpaneltwoD{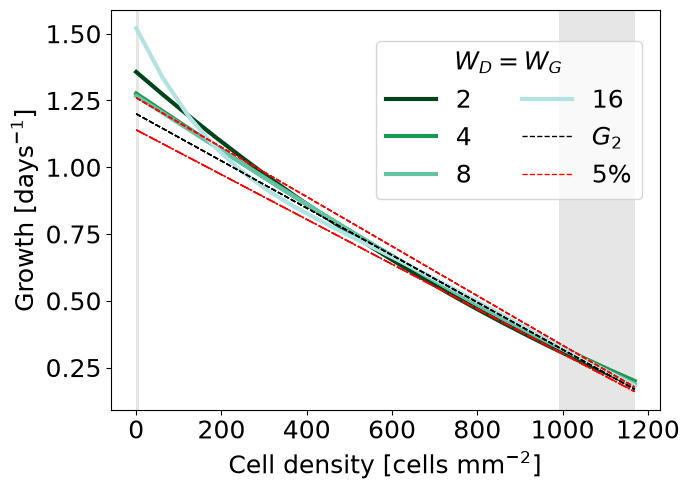}
\\
};

\coordinate (boxNW) at ([xshift=-4pt,yshift=18pt]m.north west);
\coordinate (boxSE) at ([xshift= 4pt,yshift= 2pt]m.south east);

\draw[
lightgray,
line width=1pt,
rounded corners=5pt
]
(boxNW) rectangle (boxSE);

\coordinate (vDivAB) at
($(m-1-1.east)!0.5!(m-1-2.west)$);

\draw[
lightgray,
line width=1pt
]
(vDivAB |- boxNW) -- (vDivAB |- boxSE);

\coordinate (headerAC) at
(m-1-1.north);

\coordinate (headerBC) at
($([xshift=-2pt]m-1-2.north west)!0.5!
  ([xshift= 2pt]m-1-3.north east)$);

\node[
fill=white,
text=black,
font=\figheaderfonttwoD,
anchor=north,
inner xsep=1pt,
inner ysep=0.5pt,
align=center,
text width=\boxheaderawidthtwoD
] at ([yshift=15pt]headerAC)
{\small \textbf{(a)} Performance};

\node[
fill=white,
text=black,
font=\figheaderfonttwoD,
anchor=north,
inner xsep=1pt,
inner ysep=0.5pt,
align=center,
text width=\boxheaderbwidthtwoD
] at ([yshift=15pt]headerBC)
{\small \textbf{(b)} Learned diffusion and growth across $W_D=W_G$ ($ES=3000$)};

\end{tikzpicture}

\caption{
\textbf{Performance for Case~IV across right-hand side (RHS) widths.}
Constant diffusion ($D_1$) paired with linear growth ($G_2$).
Panel \textbf{(a)} shows the validation loss across four joint RHS MLP widths $W_D=W_G \in \{2,4,8,16 \}$ and four early-stopping patience values $ES\in\{500,1000,2000,3000\}$. Panel \textbf{(b)} shows the corresponding diffusion and growth predictions for $ES=3000$.}
\label{fig:A5}

\end{figure}
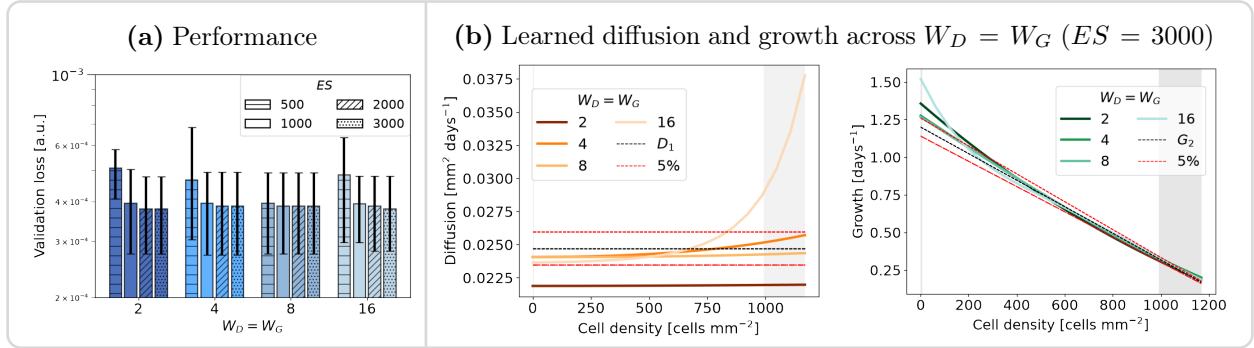
\begin{figure}[H]
\centering

\begin{minipage}{0.38\linewidth}
  \centering
  \cornerlabelimgshift[width=\linewidth]
    {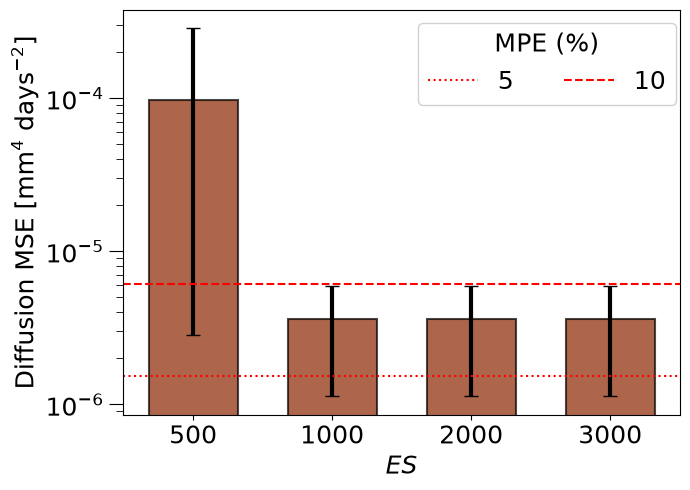}
    {(a)}{north}{xshift=0pt,yshift=-10pt}
\end{minipage}
\hspace{1cm}
\begin{minipage}{0.38\linewidth}
  \centering
  \cornerlabelimgshift[width=\linewidth]
    {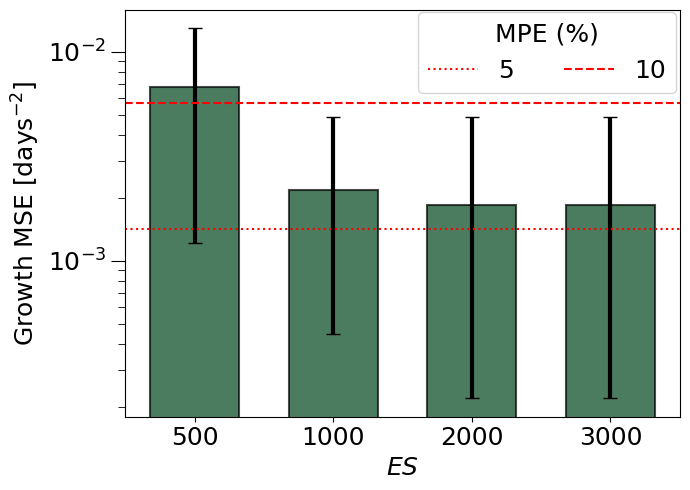}
    {(b)}{north}{xshift=5pt,yshift=-10pt}
\end{minipage}

\caption{\textbf{Ground-truth verification for Case~IV with $W_D=W_G=4$.} Mean-squared error (MSE) of learned functions against the known $D_1(u)$ and $G_2(u)$, across the early-stopping patiences ($ES$), for the noisy $2\text{D}{+}t$ data. \textbf{(a)} Diffusion-function error. \textbf{(b)} Growth-function error.}
\label{fig:A6}
\end{figure}

\clearpage
\newpage
\setcounter{equation}{0}
\renewcommand{\theequation}{S.\arabic{equation}}
\renewcommand{\theHequation}{S.\arabic{equation}}
\setcounter{section}{0}
\renewcommand{\thesection}{SM\arabic{section}}
\renewcommand{\theHsection}{SM\arabic{section}}
\setcounter{figure}{0}
\renewcommand{\thefigure}{S.\arabic{figure}}
\renewcommand{\theHfigure}{S.\arabic{figure}}

\captionsetup[figure]{font=small}

\setcounter{page}{1}
\section*{Supplementary Material}

In Section~\ref{sec:SM1} we motivate our choices for the group 2 hyperparameters, and in Section~\ref{sec:SM2} we motivate our choice of the data-loss objective. All selections can be made using the ground-truth-free hyperparameter-selection strategy. However, to provide insight into why specific hyperparameters exert their observed effects, we present errors with respect to the ground-truth diffusion and growth functions side by side with the metrics available in experimental settings. 
\\[1em]
To ensure transparency and reproducibility of the BINN architecture used, we provide a detailed description of the noise implementation, the ground-truth error metrics, data preprocessing, loss function definitions, MLP design, and early-stopping implementation in Section~\ref{sec:SM3}.

\vspace{1em}
\begingroup
\makeatletter
\renewcommand*\l@section{\@dottedtocline{1}{1.5em}{2.75em}}%
\renewcommand*\l@subsection{\@dottedtocline{2}{4.7em}{3.6em}}%
\makeatother
\setcounter{tocdepth}{2}%
\startcontents[supmat]%
\printcontents[supmat]{}{1}{\noindent\textbf{\large Contents}\par\vspace{0.5em}}%
\endgroup
\vspace{1em}

\newpage
\section{Group 2 hyperparameters} 
\label{sec:SM1}
\subsection{PDE and data loss weights} \label{sec:SM:pde_data_loss}

\begin{figure}[H]
\centering
\setlength{\groupboxwidth}{0.9\textwidth}
\begin{tabular}{@{}c c@{}}
\begin{subfigure}[c]{0.3\linewidth}
    \cornerlabelimgshift[width=\linewidth]
        {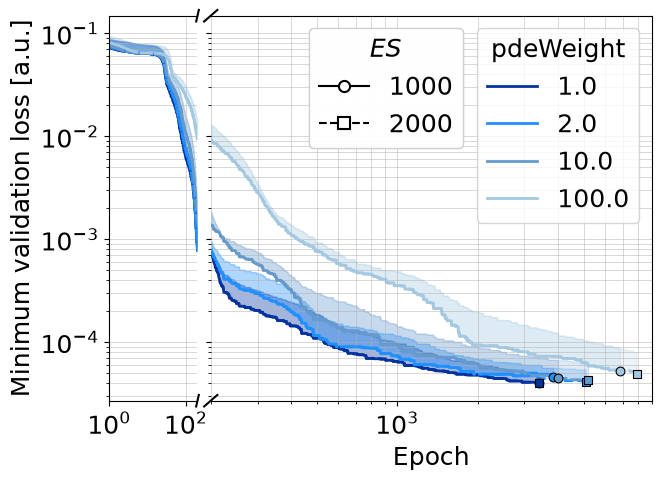}
        {(a)}{north}{xshift=5pt,yshift=10pt}
\end{subfigure} &
\begin{subfigure}[c]{0.3\linewidth}
    \cornerlabelimgshift[width=\linewidth]
        {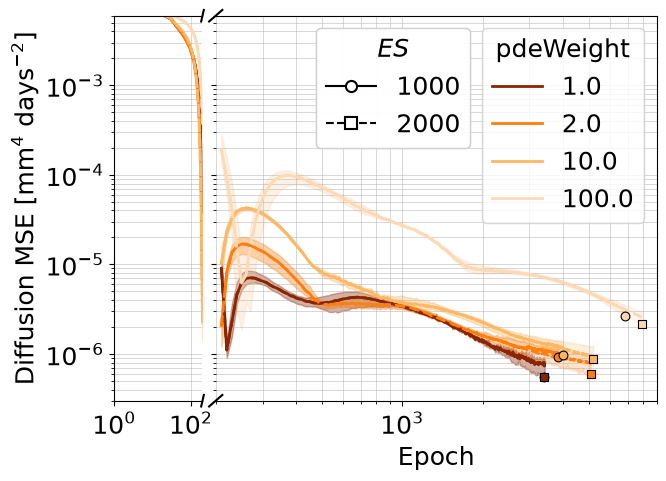}
        {(b)}{north}{xshift=5pt,yshift=10pt}
\end{subfigure}
\end{tabular}

\vspace{0.05cm}

\begin{tabular}{@{}c c c@{}}
\begin{subfigure}[c]{0.3\linewidth}
    \cornerlabelimgshift[width=\linewidth]
        {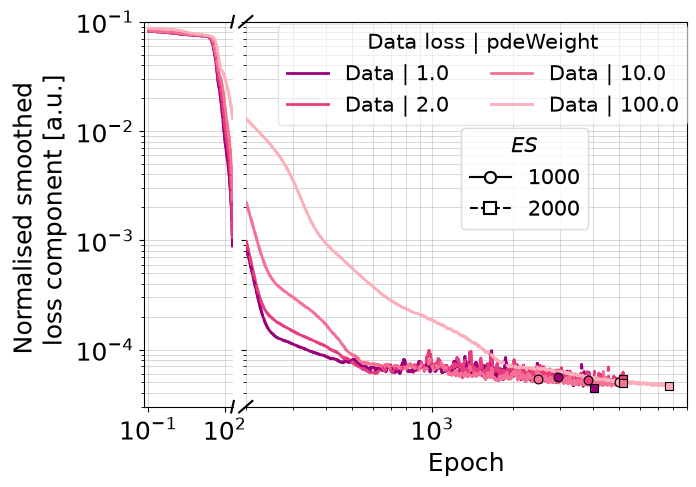}
        {(c)}{north}{xshift=5pt,yshift=10pt}
\end{subfigure} &
\begin{subfigure}[c]{0.3\linewidth}
    \cornerlabelimgshift[width=\linewidth]
        {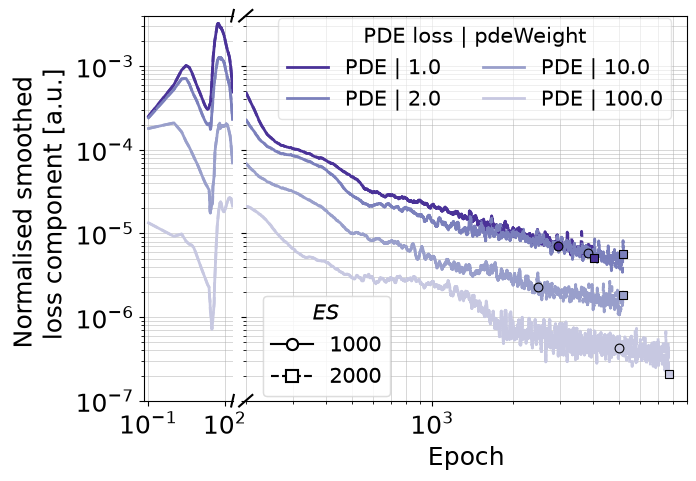}
        {(d)}{north}{xshift=5pt,yshift=10pt}
\end{subfigure} &
\begin{subfigure}[c]{0.3\linewidth}
    \cornerlabelimgshift[width=\linewidth]
        {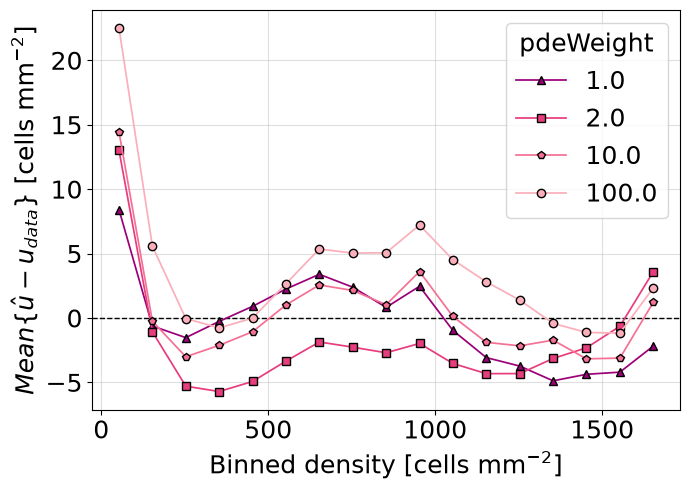}
        {(e)}{north}{xshift=5pt,yshift=10pt}
\end{subfigure}
\end{tabular}

\vspace{0.08cm}

\begin{tikzpicture}
\node[
draw=lightgray,
line width=1pt,
rounded corners=5pt,
inner sep=0pt,
text width=0.49\textwidth,
align=center
] (box1) {
\begin{minipage}{\linewidth}
\centering
\vspace*{0.1cm}
\begin{tabular}{@{}c c@{}}
    \begin{subfigure}[c]{0.47\linewidth}
        \cornerlabelimgshift[width=\linewidth]
            {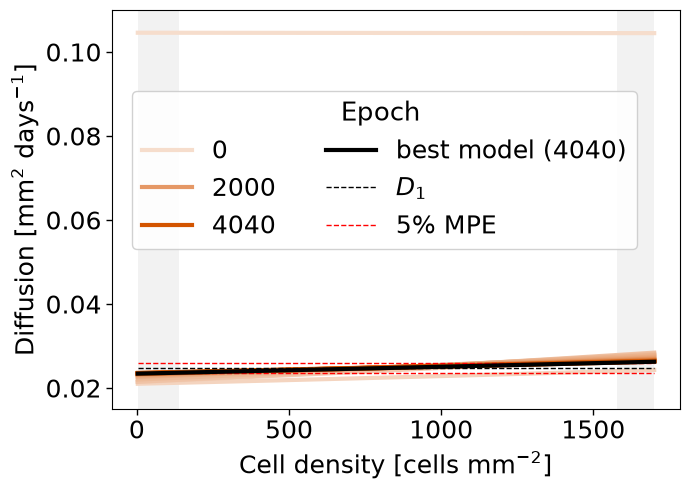}
            {}{north}{xshift=5pt,yshift=-5pt}
    \end{subfigure} &
    \begin{subfigure}[c]{0.47\linewidth}
        \cornerlabelimgshift[width=\linewidth]
            {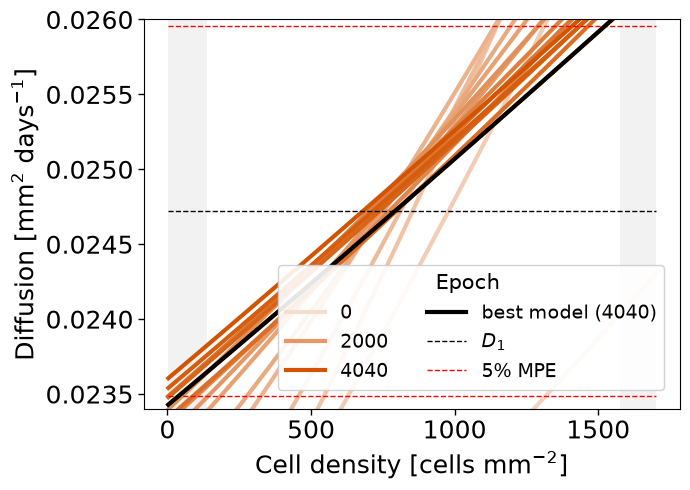}
            {}{north}{xshift=5pt,yshift=-5pt}
    \end{subfigure}
\end{tabular}
\end{minipage}
};
\node[fill=white, text=black] at (box1.north)
    {\textbf{(f)} $\lambda_{\text{PDE}} = 1$};
\end{tikzpicture}
\hfill
\begin{tikzpicture}
\node[
draw=lightgray,
line width=1pt,
rounded corners=5pt,
inner sep=0pt,
text width=0.49\textwidth,
align=center
] (box2) {
\begin{minipage}{\linewidth}
\centering
\vspace*{0.1cm}
\begin{tabular}{@{}c c@{}}
    \begin{subfigure}[c]{0.47\linewidth}
        \cornerlabelimgshift[width=\linewidth]
            {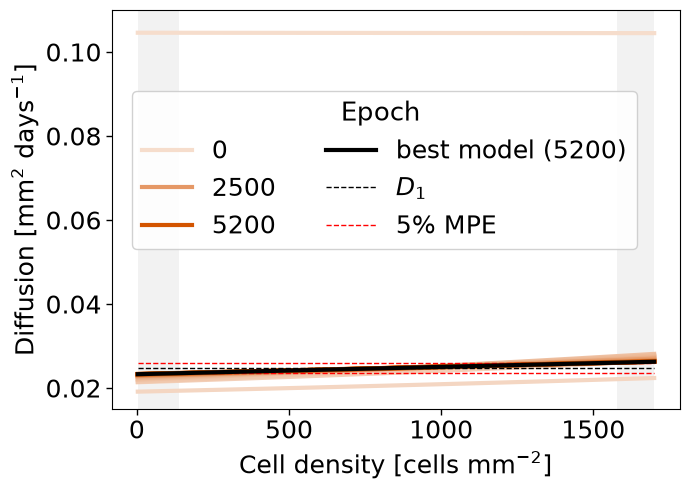}
            {}{north}{xshift=5pt,yshift=-5pt}
    \end{subfigure} &
    \begin{subfigure}[c]{0.47\linewidth}
        \cornerlabelimgshift[width=\linewidth]
            {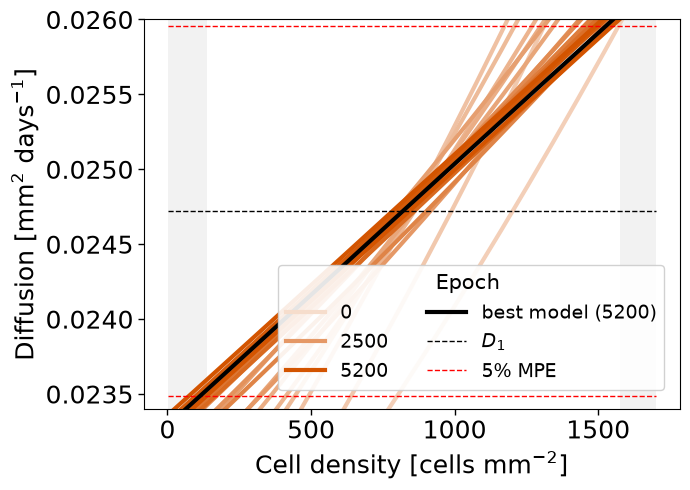}
            {}{north}{xshift=5pt,yshift=-5pt}
    \end{subfigure}
\end{tabular}
\end{minipage}
};
\node[fill=white, text=black] at (box2.north)
    {\textbf{(g)} $\lambda_{\text{PDE}} = 2$};
\end{tikzpicture}

\vspace{0.0cm}

\begin{tikzpicture}
\node[
draw=lightgray,
line width=1pt,
rounded corners=5pt,
inner sep=0pt,
text width=0.49\textwidth,
align=center
] (box3) {
\begin{minipage}{\linewidth}
\centering
\vspace*{0.1cm}
\begin{tabular}{@{}c c@{}}
    \begin{subfigure}[c]{0.47\linewidth}
        \cornerlabelimgshift[width=\linewidth]
            {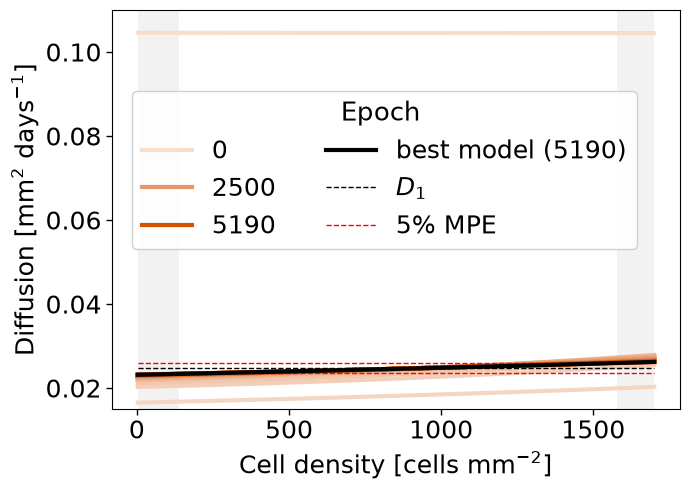}
            {}{north}{xshift=5pt,yshift=-5pt}
    \end{subfigure} &
    \begin{subfigure}[c]{0.47\linewidth}
        \cornerlabelimgshift[width=\linewidth]
            {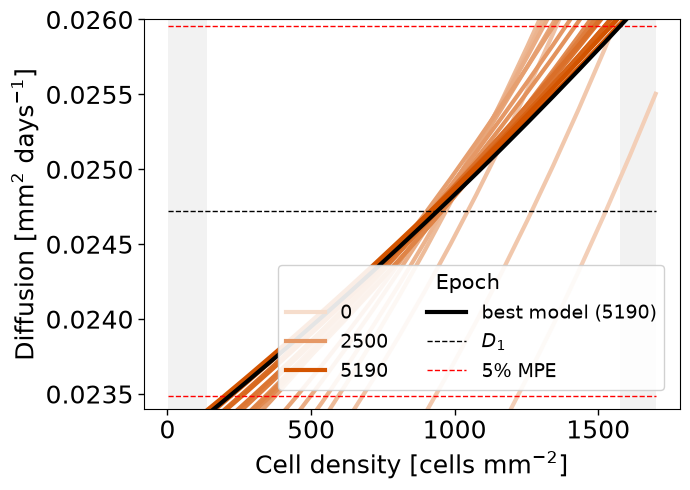}
            {}{north}{xshift=5pt,yshift=-5pt}
    \end{subfigure}
\end{tabular}
\end{minipage}
};
\node[fill=white, text=black] at (box3.north)
    {\textbf{(h)} $\lambda_{\text{PDE}} = 10$};
\end{tikzpicture}
\hfill
\begin{tikzpicture}
\node[
draw=lightgray,
line width=1pt,
rounded corners=5pt,
inner sep=0pt,
text width=0.49\textwidth,
align=center
] (box4) {
\begin{minipage}{\linewidth}
\centering
\vspace*{0.1cm}
\begin{tabular}{@{}c c@{}}
    \begin{subfigure}[c]{0.47\linewidth}
        \cornerlabelimgshift[width=\linewidth]
            {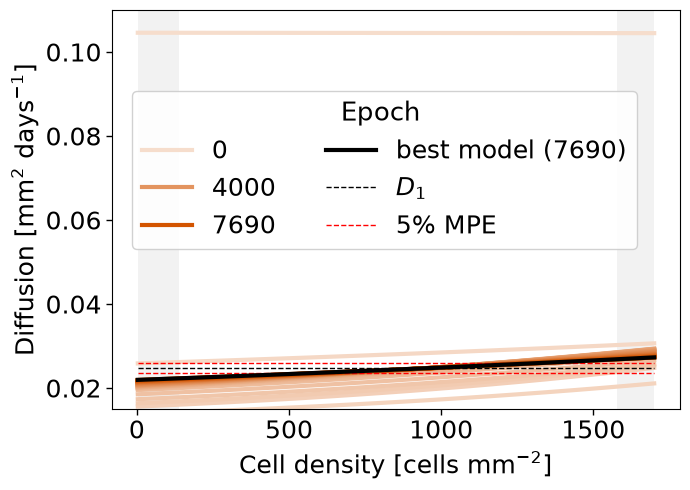}
            {}{north}{xshift=5pt,yshift=-5pt}
    \end{subfigure} &
    \begin{subfigure}[c]{0.47\linewidth}
        \cornerlabelimgshift[width=\linewidth]
            {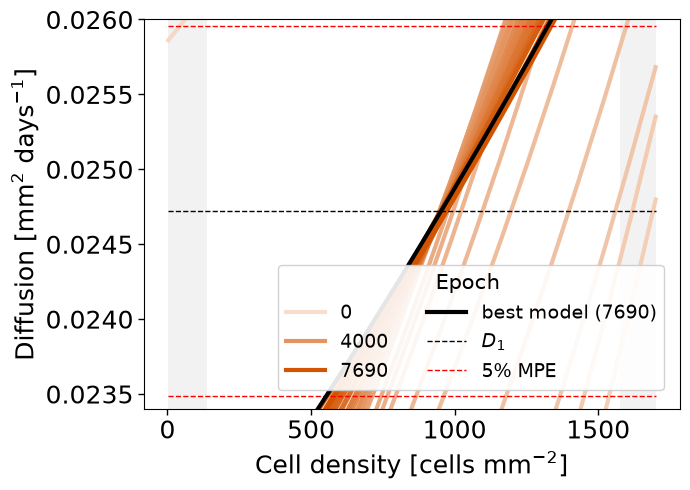}
            {}{north}{xshift=5pt,yshift=-5pt}
    \end{subfigure}
\end{tabular}
\end{minipage}
};
\node[fill=white, text=black] at (box4.north)
    {\textbf{(i)} $\lambda_{\text{PDE}} = 100$};
\end{tikzpicture}

\caption{
\textbf{Performance across PDE-loss weights $\lambda_{\text{PDE}}$ for synthetic Case~I with 1\% mean percentage error noise.}
Ground truth: $D_1(u)$. Hyperparameters: $\lambda_{\text{data}}=1$, $W_u=64$, $W_D=2$, early-stopping patiences $ES \in \{1000, 2000\} $, and $\lambda_{\text{PDE}} \in \{1,2,10,100\}$.
\textbf{(a)}, \textbf{(b)} consist of $3$ training--validation (TV) splits: markers show the minimum for each $ES$, and shading shows the corresponding min--max range. \textbf{(c)}--\textbf{(i)} consist of the first TV split at $ES=2000$.
\textbf{(a)} running minimum validation loss. 
\textbf{(b)} running minimum diffusion-function mean-squared error (MSE).
\textbf{(c)} running normalised mean validation data loss smoothed over the preceding 25 epochs.
\textbf{(d)} running normalised mean validation PDE loss smoothed over the preceding 25 epochs.
\textbf{(e)} mean difference between the density network and the data as a function of density (bin size of $100\text{ cells mm}^{-2}$).
Blocks \textbf{(f)}--\textbf{(i)} show diffusion predictions sampled every 100 epochs with light-to-dark shading corresponding to increasing epochs, with the final best model indicated by a solid black line. The $\pm5\%$ mean percentage error (MPE) bands are shown for reference.
}
\label{fig:sm1}
\end{figure}

Here we motivate why we set $\lambda_{\text{PDE}}=\lambda_{\text{data}} = 1$. To do this we use synthetic Case I with the
addition of noise ($\gamma = 1$). The noise level is set to 1\% MPE. We use 
$W_{u} = 64$, $W_D = 2$, and early-stopping patiences $ES \in \{1000,2000\}$. 
We investigate $\lambda_{\text{PDE}} \in \{1, 2, 10, 100\}$,
with $\lambda_{\text{data}} = 1$ fixed in Eq.~\eqref{eq:total_loss}. The motivation for increasing $\lambda_{\text{PDE}}$ 
is to examine whether bringing the smaller PDE loss closer to the data loss 
in magnitude at the start of training improves performance. We set \(\lambda_{\text{bio}} = 1\), with \(\lambda_{D_{\lim}} = 0\) and 
\(\lambda_{\partial D} = 10\); see Section~\ref{sec:SM:bio_loss} for the rationale behind 
these values.
\\[1em]
Figure~\ref{fig:sm1}a--b shows that increasing $\lambda_{\text{PDE}}$ does not qualitatively alter the running-minimum validation-loss or diffusion-function MSE profiles, although both are systematically shifted upward in value. The increase in validation loss is expected, since a larger $\lambda_{\text{PDE}}$ directly increases the weighted contribution of the PDE residual to the total loss. By contrast, the associated slowdown in diffusion-function recovery is more subtle and is difficult to diagnose from the total validation loss alone. To remove the dependence on the loss weights and aid comparison of the training dynamics, we therefore plot the data and PDE loss components normalised by $\lambda_{\text{data}}$ and $\lambda_{\text{PDE}}$, respectively (Fig.~\ref{fig:sm1}c,d). We also examine the difference between the data and density predictions for a given TV split as a function of density to provide further information (Fig.~\ref{fig:sm1}e).

Figure~\ref{fig:sm1}c shows that during the early stages of training, larger values of $\lambda_{\text{PDE}}$ impede recovery of the density, as evidenced by the substantially higher data loss. This delayed density recovery is accompanied by slower diffusion-function recovery. Beyond approximately 500 epochs, however, the data loss curves for $\lambda_{\text{PDE}} \leq 10$ become very similar. Although the normalised PDE loss is substantially lower for $\lambda_{\text{PDE}}=10$ than for $\lambda_{\text{PDE}}=1,2$, the corresponding diffusion predictions (Fig.~\ref{fig:sm1}f--i) show that this reduction does not materially improve recovery by the end of training. This suggests that accurate density matching is the main bottleneck limiting diffusion-function recovery.

For $\lambda_{\text{PDE}}=100$, more than 1000 epochs are required before the data loss becomes comparable to that obtained with smaller PDE weights, and a similar delay is consequently observed in the diffusion-function recovery. Moreover, Fig.~\ref{fig:sm1}i shows that approximately 4000 additional epochs are required before the diffusion predictions begin to stabilise, relative to the runs with small $\lambda_{\text{PDE}}$. Despite the data and PDE losses eventually reaching values comparable to those obtained for smaller $\lambda_{\text{PDE}}$, Fig.~\ref{fig:sm1}f--i show that $\lambda_{\text{PDE}}=100$ nevertheless produces the poorest final recovery of the diffusion function. To more carefully diagnose this discrepancy, we examine the difference between the observed and learned density profiles. This reveals a substantially larger error in the low-density regime for $\lambda_{\text{PDE}}=100$ (Fig.~\ref{fig:sm1}e). Since the initial condition is a cosine profile (see Table~\ref{tab:datagen}), these low densities correspond to the troughs of the early-time profiles, where the density is overestimated. This systematic density overestimation is associated with underestimation of the inferred diffusion function in the same density regime. 
In practice, pronounced and persistent discrepancies concentrated around particular density values may therefore provide a useful diagnostic warning sign, especially when those discrepancies can be linked to a specific feature of the underlying dynamics.
\\[1em]
Overall, increasing the constant $\lambda_{\text{PDE}}$ does not systematically improve diffusion-function recovery; instead, it can both slow optimisation and impair accurate recovery. Intuitively, strong enforcement of the PDE constraint is most beneficial only after the data have been fitted sufficiently well. Placing too much emphasis on the PDE term early in training can therefore hinder the data fitting needed for accurate subsequent recovery of the diffusion function.
\\[1em]
The data and PDE terms compete along a Pareto front whose shape is skewed by how 
the system parameters scale the 
residual~\cite{rohrhofer_data_2023}.
Among fixed weights, Fig.~\ref{fig:sm1} suggests 
that the naive choice of equal weights $\lambda_{\text{data}} = \lambda_{\text{PDE}} = 1$ 
is a reasonable default.
Dynamic weighting strategies instead balance the terms 
adaptively, which can be more effective, but can incur additional computational cost and typically require tuning. 
We retain fixed weights here, leaving more effective balancing, for instance via constrained-optimisation schemes such as PINNverse~\cite{almanstotter_pinnverse_2025}, to future work.

\subsection{Weights in biological loss} \label{sec:SM:bio_loss}
\begin{figure}[H]
\centering

\setlength{\groupboxwidth}{0.6\textwidth}

\begin{tikzpicture}
    \node[
        draw=lightgray,
        line width=1pt,
        rounded corners=5pt,
        inner sep=0pt,
        align=center
    ] (group1box) {
        \begin{minipage}{1\textwidth}
            \centering
            
            \vspace*{0.3cm} 

            \begin{tabular}{@{}c c c@{}}
                \begin{subfigure}{0.3\textwidth}
                    \cornerlabelimgshift[width=\linewidth]
                    {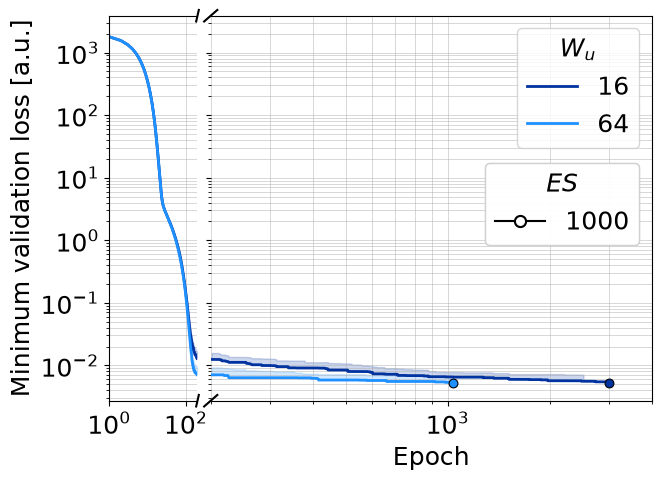}
                    {(a)}{north}{xshift=5pt,yshift=-12pt}
                \end{subfigure} &
                \begin{subfigure}{0.3\textwidth}
                    \cornerlabelimgshift[width=\linewidth]
                    {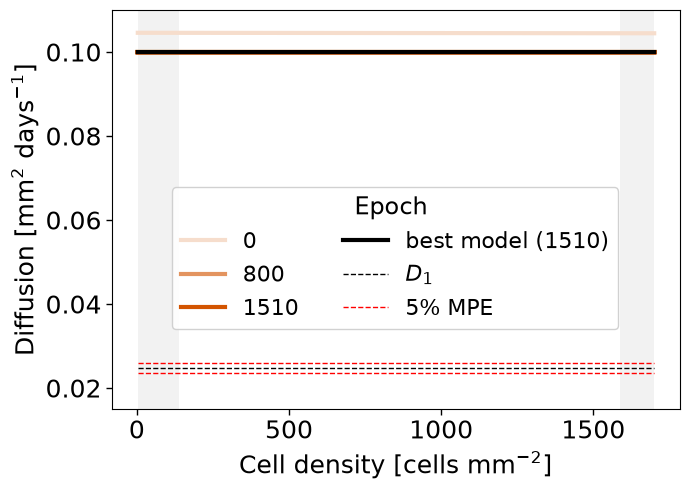}
                    {(b) $W_u = 16$}{north}{xshift=5pt,yshift=-12pt}
                \end{subfigure} &
                \begin{subfigure}{0.3\textwidth}
                    \cornerlabelimgshift[width=\linewidth]
                    {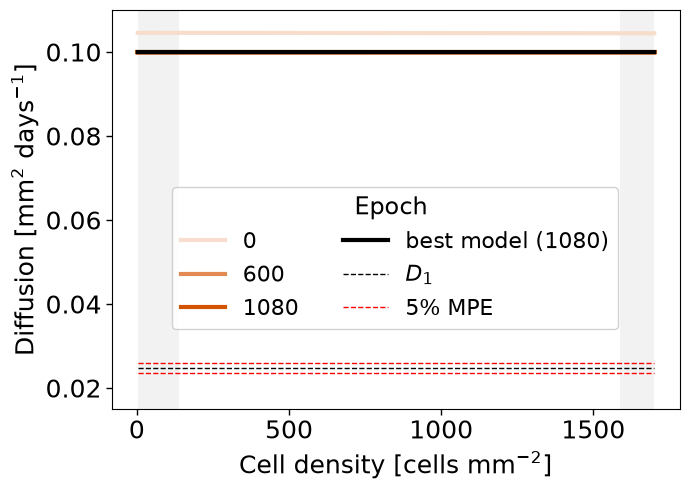}
                    {(c) $W_u = 64$}{north}{xshift=5pt,yshift=-12pt}
                \end{subfigure}
            \end{tabular}

        \end{minipage}
    };

    \node[fill=white, text=black] at (group1box.north)
    { $\lambda_{D_{\lim}} = \lambda_{\partial D} = 10^4$};
\end{tikzpicture}

\begin{tikzpicture}
    \node[
        draw=lightgray,
        line width=1pt,
        rounded corners=5pt,
        inner sep=0pt,
        align=center
    ] (group2box) {
        \begin{minipage}{1\textwidth}
            \centering
            
            \vspace*{0.3cm}

            \begin{tabular}{@{}c c c@{}}
                \begin{subfigure}{0.3\textwidth}
                    \cornerlabelimgshift[width=\linewidth]
                    {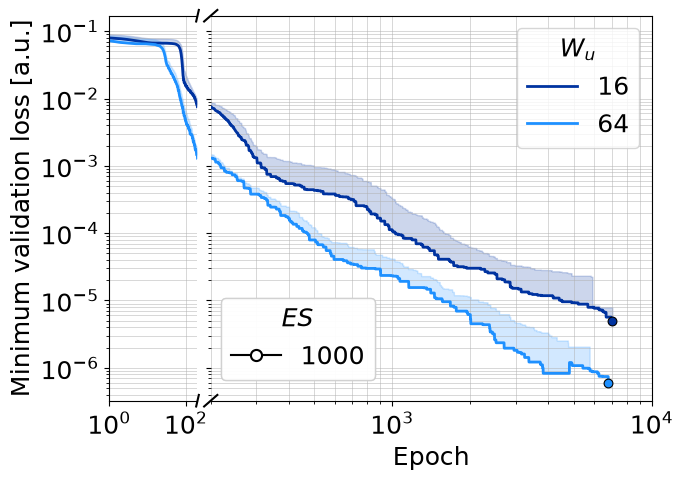}
                    {(d)}{north}{xshift=5pt,yshift=-12pt}
                \end{subfigure} &
                \begin{subfigure}{0.3\textwidth}
                    \cornerlabelimgshift[width=\linewidth]
                    {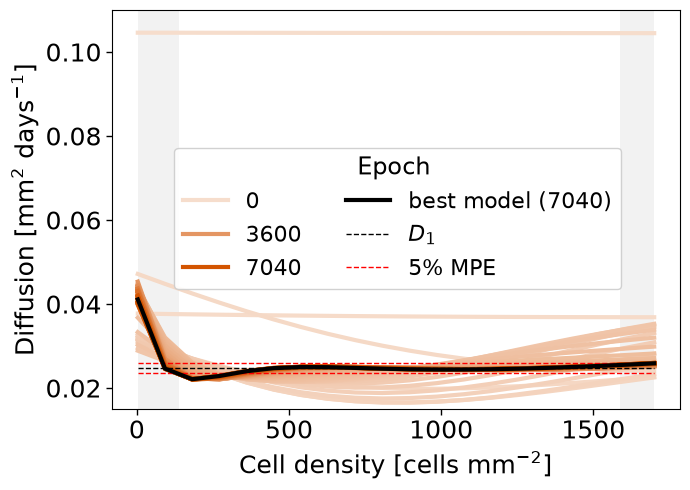}
                    {(e) $W_u = 16$}{north}{xshift=5pt,yshift=-12pt}
                \end{subfigure} &
                \begin{subfigure}{0.3\textwidth}
                    \cornerlabelimgshift[width=\linewidth]
                    {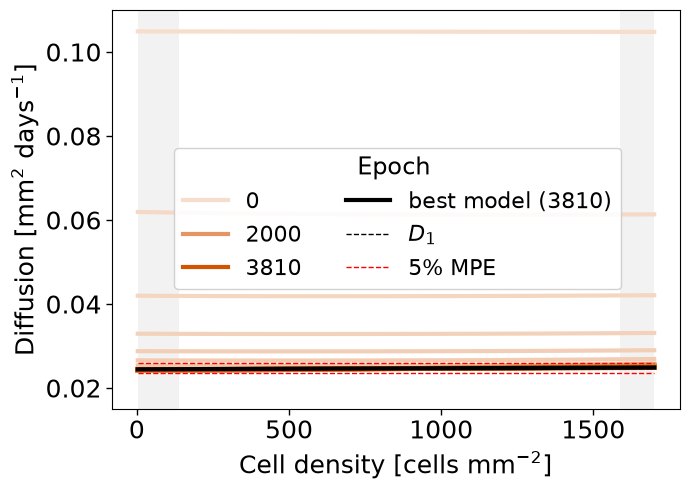}
                    {(f) $W_u = 64$}{north}{xshift=5pt,yshift=-12pt}
                \end{subfigure}
            \end{tabular}

        \end{minipage}
    };

    \node[fill=white, text=black] at (group2box.north)
    {$\lambda_{D_{\lim}} = \lambda_{\partial D} = 0$};
\end{tikzpicture}
\caption{\textbf{Performance when using large weights in the biological loss for synthetic Case~I.} Ground truth: $D_1(u)$. Hyperparameters: $W_u\in\{16,64\}$, $W_D=2$, and early-stopping patience $ES=1000$.  \textbf{(a)} Running minimum validation loss across $3$ training--validation (TV) splits: markers show the minimum for $ES=1000$, and shading shows the corresponding min--max range.
\textbf{(b)} Diffusion predictions for $W_u=16$ sampled every $100$ epochs for the first TV split.
 Light-to-dark shading corresponds to increasing epochs, with the best model indicated by a solid black line. The $\pm5\%$ mean percentage error (MPE) bands are shown for reference. \textbf{(c)} Same as \textbf{(b)} for $W_u=64$. \textbf{(d)}--\textbf{(f)} as in \textbf{(a)}--\textbf{(c)} for $\lambda_{D_{\lim}} = \lambda_{\partial D} = 0$.}
\label{fig:sm1-2}

\end{figure}

Here we motivate the choice of the biological weights in Eq.~\eqref{eq:bio_loss}: $\lambda_{D_{\lim}} = \lambda_{G_{\lim}} = 0$ and $\lambda_{\partial D} = \lambda_{\partial G} = 10$. To do this we use synthetic Case I. We use $W_u \in \{16, 64\}$, $W_D = 2$, and early-stopping patience $ES = 1000$.
Since we enforce constraints on the normalised diffusion and growth predictions equally, we couple the corresponding weights such that $\lambda_{D_{\lim}} = \lambda_{G_{\lim}}$ and $\lambda_{\partial D} = \lambda_{\partial G}$ throughout.
\\[1em]
To enforce constraints strictly, we initially set $\lambda_{D_{\lim}}=\lambda_{\partial D}=10^4$. However, this produces a persistent plateau in minimum validation loss (Fig.~\ref{fig:sm1-2}a), indicating that the density network fails to learn the data while diffusion predictions remain fixed near $d_{\max}=0.10\,\mathrm{mm}^2\,\mathrm{days}^{-1}$ throughout training (Fig.~\ref{fig:sm1-2}b,c). This suggests excessively large penalty weights overly restrict optimisation, preventing the density network from accurately representing the data and consequently inhibiting diffusion learning.

Subsequently setting $\lambda_{D_{\lim}} = \lambda_{\partial D} = 0$ allows validation loss profiles to decrease rapidly across several orders of magnitude without pronounced plateaus (Fig.~\ref{fig:sm1-2}d). This improvement is reflected in rescaled diffusion predictions, which quickly drop from initialised values near $0.10\,\mathrm{mm}^2\,\mathrm{days}^{-1}$ to approximately $0.04\,\mathrm{mm}^2\,\mathrm{days}^{-1}$ (Fig.~\ref{fig:sm1-2}e,f) before converging. Visualising predictions across epochs suggests the bounding constraint is contextually redundant: because network outputs are initialised near one while the true normalised diffusion coefficient is substantially smaller, the network naturally moves below the upper bound without explicit enforcement. However, unlike the bounding constraint, the monotonicity constraint cannot simply be removed without undesirable consequences (evidenced in Fig.~\ref{fig:sm1-2}e for $W_u = 16$).

\begin{figure}[H]
\centering
\setlength{\groupboxwidth}{1\textwidth}

\begin{tikzpicture}
\node[
    draw=lightgray, 
    line width=1pt, 
    rounded corners=5pt, 
    inner sep=0pt, 
    text width=\groupboxwidth, 
    align=center
] (box1) {
    \begin{minipage}{\linewidth}
    \centering
    \vspace*{0.35cm}
    \begin{tabular}{@{}c c c@{}}
        \begin{subfigure}[c]{0.3\linewidth}
            \cornerlabelimgshift[width=\linewidth]
                {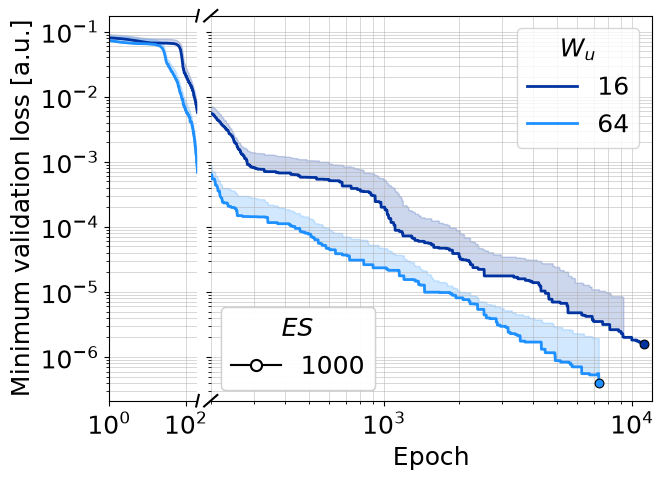}
                {}{north}{xshift=5pt,yshift=-5pt}
        \end{subfigure} & 
        \begin{subfigure}[c]{0.3\linewidth}
            \cornerlabelimgshift[width=\linewidth]
                {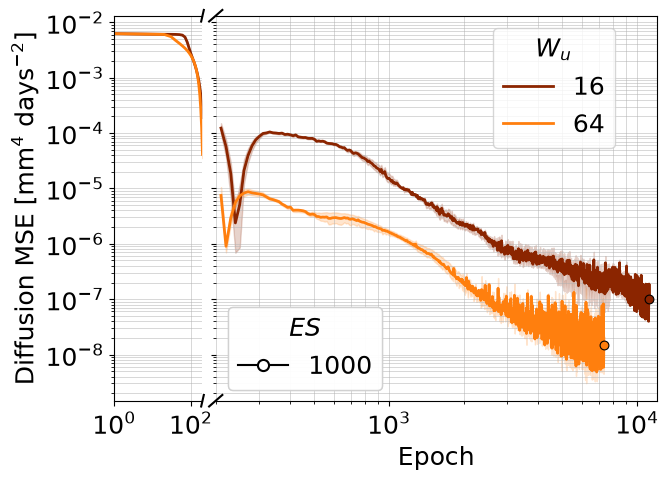}
                {}{north}{xshift=5pt,yshift=-5pt}
        \end{subfigure} & 
        \begin{subfigure}[c]{0.3\linewidth}
            \cornerlabelimgshift[width=\linewidth]
                {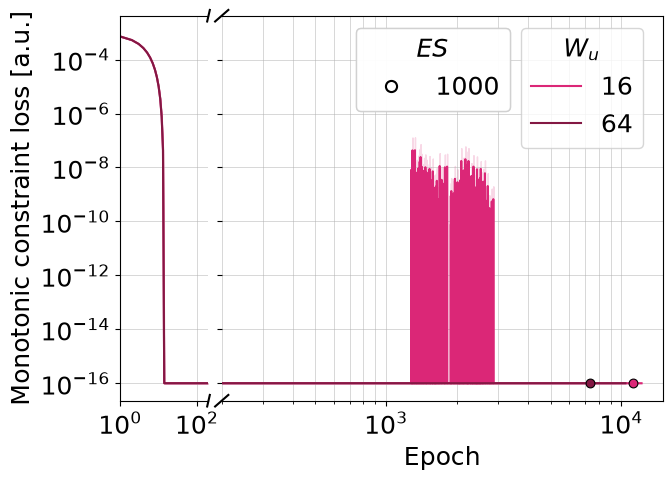}
                {}{north}{xshift=5pt,yshift=-5pt}
        \end{subfigure}
    \end{tabular}
    \end{minipage}
};
\node[fill=white, text=black] at (box1.north) {\textbf{(a)} \textbf{$\lambda_{\partial D} = 10$}};
\end{tikzpicture}

\begin{tikzpicture}
\node[
    draw=lightgray, 
    line width=1pt, 
    rounded corners=5pt, 
    inner sep=0pt, 
    text width=\groupboxwidth, 
    align=center
] (box2) {
    \begin{minipage}{\linewidth}
    \centering
    \vspace*{0.35cm}
    \begin{tabular}{@{}c c c@{}}
        \begin{subfigure}[c]{0.3\linewidth}
            \cornerlabelimgshift[width=\linewidth]
                {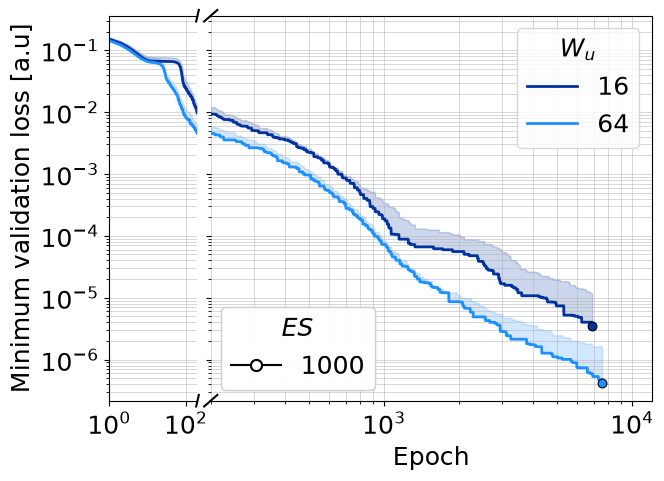}
                {}{north}{xshift=5pt,yshift=-5pt}
        \end{subfigure} & 
        \begin{subfigure}[c]{0.3\linewidth}
            \cornerlabelimgshift[width=\linewidth]
                {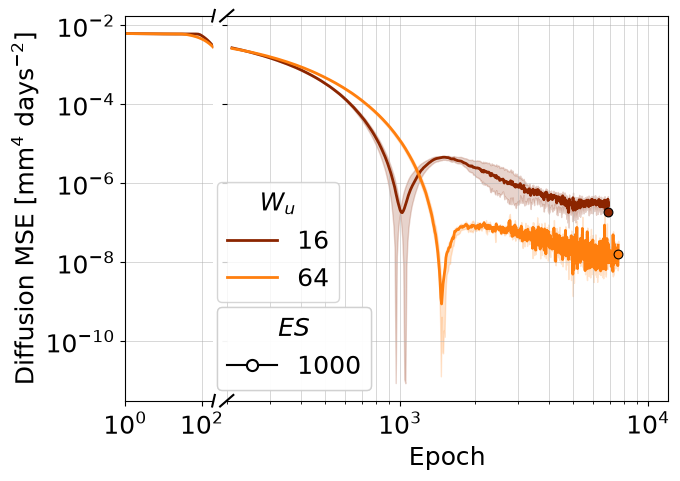}
                {}{north}{xshift=5pt,yshift=-5pt}
        \end{subfigure} & 
        \begin{subfigure}[c]{0.3\linewidth}
            \cornerlabelimgshift[width=\linewidth]
                {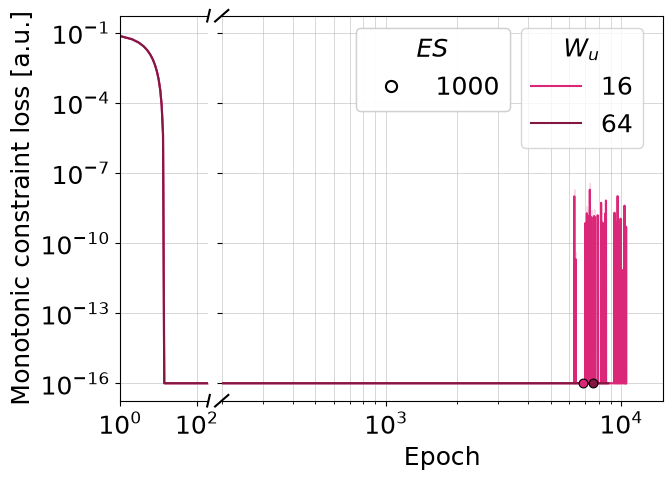}
                {}{north}{xshift=5pt,yshift=-5pt}
        \end{subfigure}
    \end{tabular}
    \end{minipage}
};
\node[fill=white, text=black] at (box2.north) {\textbf{(b)} \textbf{$\lambda_{\partial D} = 10^3$}};
\end{tikzpicture}

\begin{tikzpicture}
\node[
    draw=lightgray, 
    line width=1pt, 
    rounded corners=5pt, 
    inner sep=0pt, 
    text width=\groupboxwidth, 
    align=center
] (box3) {
    \begin{minipage}{\linewidth}
    \centering
    \vspace*{0.35cm}
    \begin{tabular}{@{}c c c@{}}
        \begin{subfigure}[c]{0.3\linewidth}
            \cornerlabelimgshift[width=\linewidth]
                {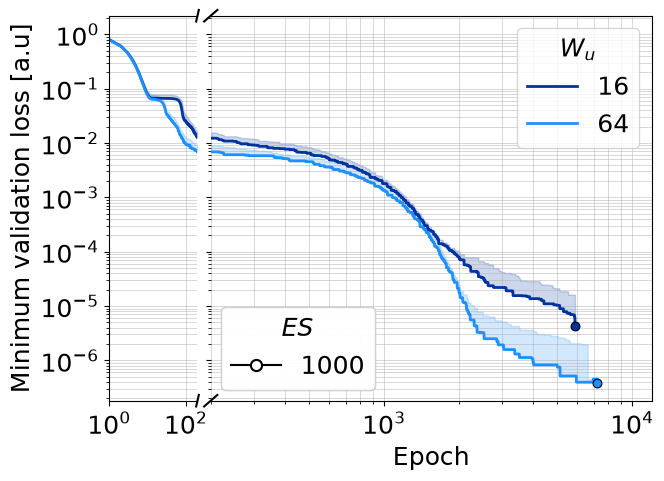}
                {}{north}{xshift=5pt,yshift=-5pt}
        \end{subfigure} & 
        \begin{subfigure}[c]{0.3\linewidth}
            \cornerlabelimgshift[width=\linewidth]
                {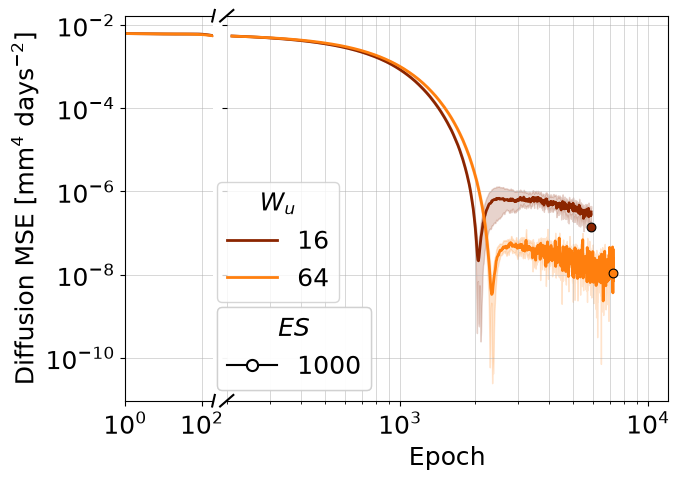}
                {}{north}{xshift=5pt,yshift=-5pt}
        \end{subfigure} & 
        \begin{subfigure}[c]{0.3\linewidth}
            \cornerlabelimgshift[width=\linewidth]
                {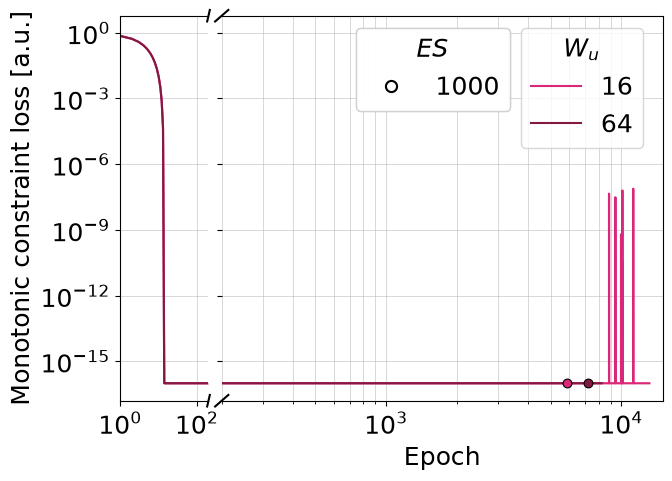}
                {}{north}{xshift=5pt,yshift=-5pt}
        \end{subfigure}
    \end{tabular}
    \end{minipage}
};
\node[fill=white, text=black] at (box3.north) {\textbf{(c)} \textbf{$\lambda_{\partial D} = 10^4$}};
\end{tikzpicture}

\begin{tikzpicture}
\node[
    draw=lightgray, 
    line width=1pt, 
    rounded corners=5pt, 
    inner sep=0pt, 
    text width=\groupboxwidth, 
    align=center
] (box4) {
    \begin{minipage}{\linewidth}
    \centering
    \vspace*{0.35cm}
    \begin{tabular}{@{}c c c@{}}
        \begin{subfigure}[c]{0.3\linewidth}
            \cornerlabelimgshift[width=\linewidth]
                {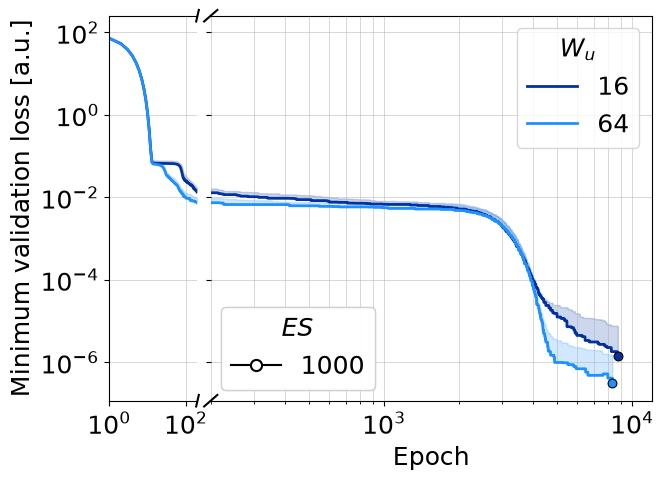}
                {}{north}{xshift=5pt,yshift=-5pt}
        \end{subfigure} & 
        \begin{subfigure}[c]{0.3\linewidth}
            \cornerlabelimgshift[width=\linewidth]
                {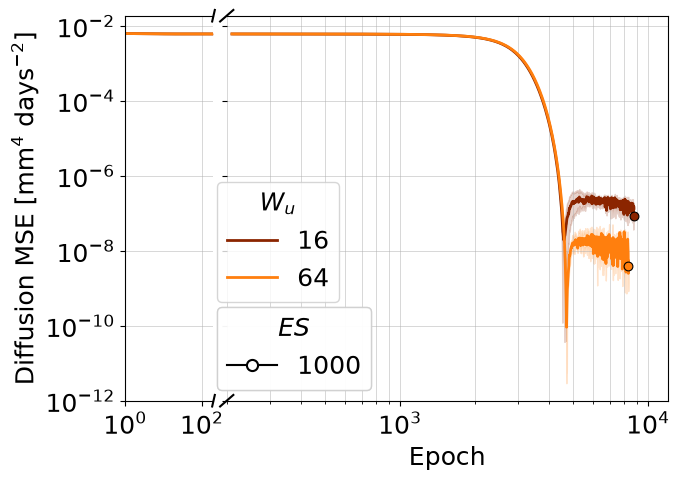}
                {}{north}{xshift=5pt,yshift=-5pt}
        \end{subfigure} & 
        \begin{subfigure}[c]{0.3\linewidth}
            \cornerlabelimgshift[width=\linewidth]
                {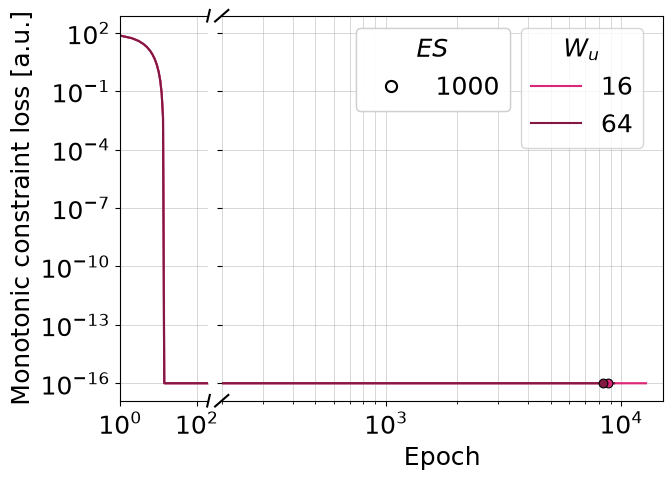}
                {}{north}{xshift=5pt,yshift=-5pt}
        \end{subfigure}
    \end{tabular}
    \end{minipage}
};
\node[fill=white, text=black] at (box4.north) {\textbf{(d)} \textbf{$\lambda_{\partial D} = 10^6$}};
\end{tikzpicture}

 \caption{\textbf{Performance across monotonicity weights $\lambda_{\partial D}$ for synthetic Case~I.} Ground truth: $D_1(u)$. Hyperparameters: $W_u \in \{16,64\}$, $W_D=2$, and early-stopping patience $ES=1000$. All plots consist of $3$ training--validation splits: markers show the minimum for $ES=1000$, and shading shows the corresponding min--max range. Within each block \textbf{(a)}--\textbf{(d)}, from left to right: running minimum validation loss, running minimum diffusion-function mean-squared error (MSE), and running mean monotonicity loss.}
\label{fig:sm1-3}
\end{figure}

\noindent Since the bounding constraint is essentially redundant here, we set $\lambda_{D_{\lim}} = 0$ and 
look only at the effect of $\lambda_{\partial D} \in \{10, 10^3, 10^4, 10^6\}$. In Fig.~\ref{fig:sm1-3}a--b, 
smaller weights ($\lambda_{\partial D} = 10$ and $10^3$) keep the healthy 
validation-loss profiles seen when $\mathcal{L}_{\text{bio}} = 0$. In contrast, larger weights introduce a validation-loss plateau (Fig.~\ref{fig:sm1-3}c--d), leading to slower convergence in the diffusion predictions (Fig.~\ref{fig:sm1-4}). Even so, all $\lambda_{\partial D}$
give similar final diffusion predictions, especially for $W_u = 64$. 
\noindent Across all six cases (three TV splits across $W_u \in \{16, 64\}$), the network satisfies the monotonicity constraint to machine precision at the best model---the saved checkpoint at epoch $e_{\mathrm{sav}}$ (Algorithm~\ref{alg:early_stopping}), which is the model carried forward (Fig.~\ref{fig:sm1-3}). However, for $W_u = 16$, the diffusion function is
non-monotonic at numerous points during training and for $\lambda_{\partial D} = 10^3, 10^4$ this can occur near the end of training (Fig.~\ref{fig:sm1-3}). Increasing 
the weight to $\lambda_{\partial D} = 10^6$ fixes this issue. The weight is 
now large enough that every network quickly reaches and keeps monotonicity. 
Furthermore, this setting yields the best learned diffusion functions for both widths, but reintroduces the validation-loss plateau observed in Fig.~\ref{fig:sm1-2}a.
\\[1em]
The user therefore has three options: (i) strongly enforce the constraint and offset optimiser stiffness with a higher $ES$, at the cost of longer runtimes; (ii) use a smaller weight that promotes constraint satisfaction without hindering optimisation, while verifying that the constraint is met; or (iii) schedule the weight to combine both approaches, at the cost of an additional scenario-dependent hyperparameter. We adopt (ii), setting $\lambda_{D_{\lim}}=\lambda_{G_{\lim}}=0$ and $\lambda_{\partial D}=\lambda_{\partial G}=10$.
\input{SM/figures/SM1/SM12_fig3}
\clearpage
\newpage
\subsection{Number of collocation points}\label{sec:SM:collocation_pts}
Here we examine the effect of the number of PDE collocation points $n_c$ per minibatch. To do this, we use the ($D_1,G_2$) pairing from synthetic Case III with $W_{u} = 64$, $W_D = W_G = 4$, and early-stopping patiences $ES \in \{1000,3000\}$. The collocation points set the quality of the PDE loss gradient at the cost of additional
runtime; we investigate $n_c \in \{10, 100, 1000\}$.
\\[1em]
The PDE loss $\mathcal{L}_{\text{PDE}}$ in Eq.~\eqref{eq:pde_loss} is a Monte Carlo estimate of the squared PDE residual integrated over the space--time domain (evaluated on $n_c$ points drawn each minibatch). At low $n_c$, this estimate becomes sparse and noisy, providing the optimiser with a weak, unrepresentative signal of the global residual. While increasing $n_c$ stabilises optimisation, it increases per-epoch runtime---leading to saturating returns once $n_c$ is large enough to cover the domain sufficiently densely. Figure~\ref{fig:sm1-5}a shows this saturation effect: the $n_c = 10$ runs reach the main loss drop
markedly later than $n_c = 100$ and $n_c = 1000$, while the latter two are nearly coincident, and reaching that drop earlier costs little. Runtimes for $n_c = 10$ and $n_c = 100$ are
similar (Fig.~\ref{fig:sm1-5}b), and only rise sharply at $n_c = 1000$. The consequence of using too few collocation points is seen in the component losses (Fig.~\ref{fig:sm1-5}c): the larger validation loss at
$n_c = 10$ is driven by the data term, while its PDE component is, in fact, lower than at $n_c = 1000$. This is an under-sampling artefact. With only $10$ points the BINN can
drive the empirical residual small at those locations without necessarily satisfying the PDE elsewhere.
\\[1em]
Function recovery shows the same saturation as the validation loss (Fig.~\ref{fig:sm1-5}d,e): 
the diffusion- and growth-function MSE profiles are nearly indistinguishable for $n_c = 100$ 
and $n_c = 1000$. Turning to the diffusion and growth predictions at $ES = 3000$, the 
$n_c = 10$ best model provides the most accurate recovery of the ground-truth diffusion and 
growth functions (Fig.~\ref{fig:sm1-5}f), even though its data loss remains higher 
(Fig.~\ref{fig:sm1-5}c), whereas the $n_c = 100$ (Fig.~\ref{fig:sm1-5}g) and $n_c = 1000$ 
(Fig.~\ref{fig:sm1-5}h) predictions are highly consistent across epochs. The $n_c = 10$ result 
is an identifiability issue: the BINN can recover diffusion and growth functions close to the 
ground truth while still carrying a larger systematic error in the density fit. We therefore do not select $n_c = 10$, since its accurate recovery is incidental rather than systematic and could not be identified as such without the ground truth.
\\[1em]
The dominant effect of increasing $n_c$ appears to be the reliability of the PDE training
signal and not necessarily a consistent improvement in the recovered functions. Moving from $n_c = 10$ to
$100$ materially changes the optimisation trajectory, whereas $100$ to $1000$ yields
diminishing returns. In this regime ($n_c \geq 100$), prediction quality appears to be limited primarily by optimisation dynamics and model--data properties rather than collocation error. Therefore, we use $n_c=100$ for the more computationally demanding $2\text{D}{+}t$ case, while the lower computational cost of $1\text{D}{+}t$ allows us to conservatively use $n_c=1000$.
\input{SM/figures/SM1/SM13_fig1}

\clearpage
\newpage
\subsection{Activation functions} \label{sec:SM:activation_funcs}
Here we examine how the choice of hidden-layer activation function (AF) affects
performance, an effect that can be substantial for PINNs \cite{Abbasi2024PAF}. To do this, we use synthetic Case~I. We use $W_u \in \{8,16,32,64,128\}$, $W_D = 2$, and early-stopping patiences $ES \in \{1000,2000,3000\}$. 

\begin{figure}[H]
\centering
\setlength{\groupboxwidth}{0.32\textwidth}

\begin{tabular}{@{}c c c@{}}

\begin{tikzpicture}
\node[
    draw=lightgray, 
    line width=1pt, 
    rounded corners=5pt, 
    inner sep=0pt, 
    text width=\groupboxwidth, 
    align=center
] (box1) {
    \begin{minipage}{\linewidth}
    \centering
    \vspace*{0.35cm}
    \begin{subfigure}[c]{0.9\linewidth}
        \cornerlabelimgshift[width=\linewidth]
            {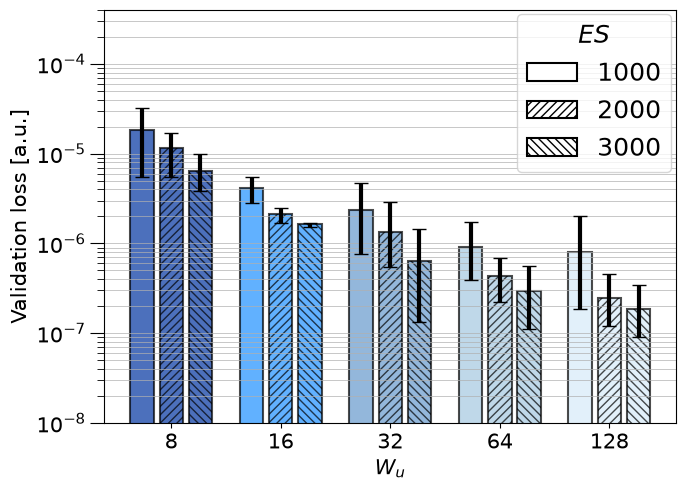}
            {}{north}{xshift=5pt,yshift=-5pt}
    \end{subfigure} \\[0.2cm]
    \begin{subfigure}[c]{0.9\linewidth}
        \cornerlabelimgshift[width=\linewidth]
            {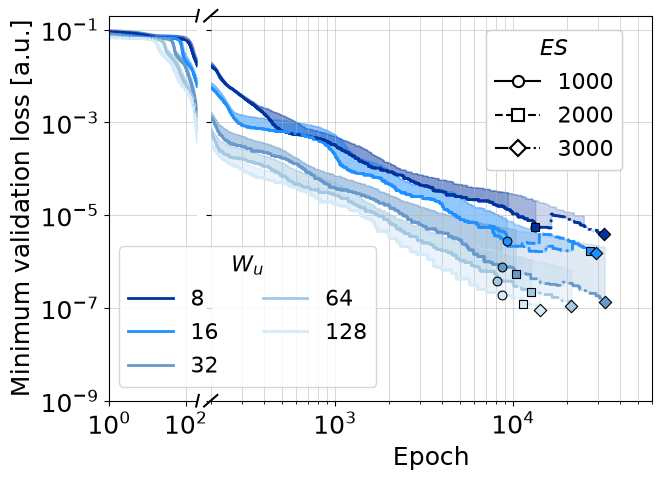}
            {}{north}{xshift=5pt,yshift=-5pt}
    \end{subfigure} \\[0.2cm]
    \begin{subfigure}[c]{0.9\linewidth}
        \cornerlabelimgshift[width=\linewidth]
            {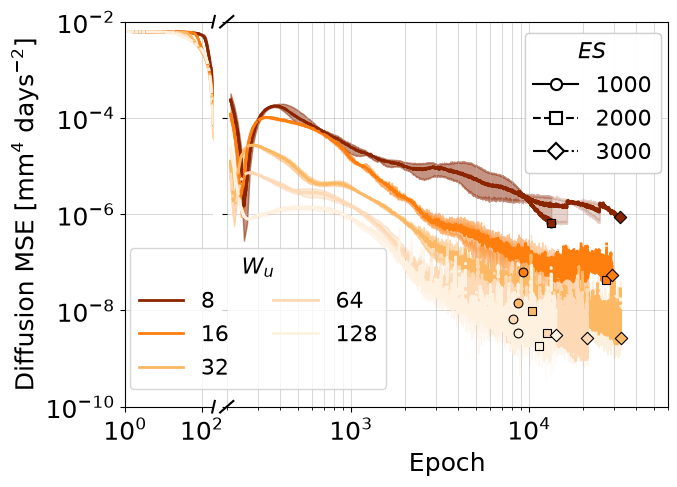}
            {}{north}{xshift=5pt,yshift=-5pt}
    \end{subfigure} \\[0.15cm]
    {\textbf{\small $W_u = 64$, $ES = 2000$}} \\[0.15cm]
    \begin{subfigure}[c]{0.9\linewidth}
        \cornerlabelimgshift[width=\linewidth]
            {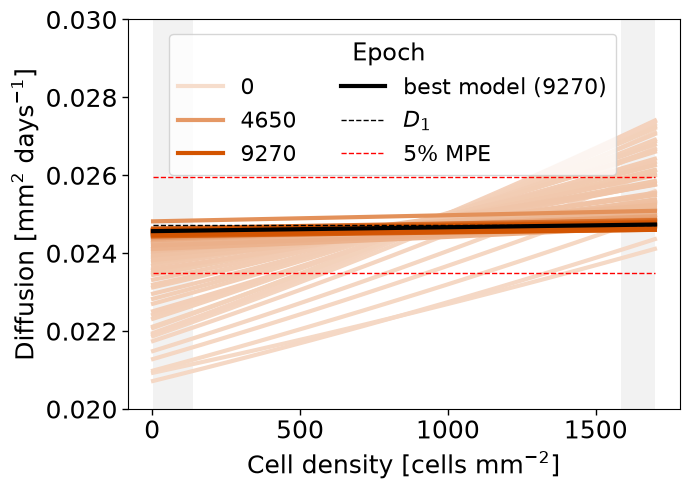}
            {}{north}{xshift=5pt,yshift=-5pt}
    \end{subfigure}
    \vspace*{0.15cm}
    \end{minipage}
};
\node[fill=white, text=black] at (box1.north) {\textbf{(a)} AF = \texttt{SiLU}};
\end{tikzpicture}

&

\begin{tikzpicture}
\node[
    draw=lightgray, 
    line width=1pt, 
    rounded corners=5pt, 
    inner sep=0pt, 
    text width=\groupboxwidth, 
    align=center
] (box2) {
    \begin{minipage}{\linewidth}
    \centering
    \vspace*{0.35cm}
    \begin{subfigure}[c]{0.9\linewidth}
        \cornerlabelimgshift[width=\linewidth]
            {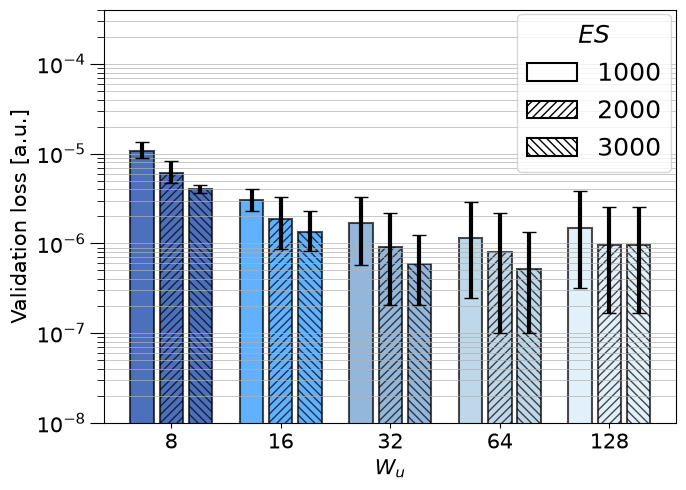}
            {}{north}{xshift=5pt,yshift=-5pt}
    \end{subfigure} \\[0.2cm]
    \begin{subfigure}[c]{0.9\linewidth}
        \cornerlabelimgshift[width=\linewidth]
            {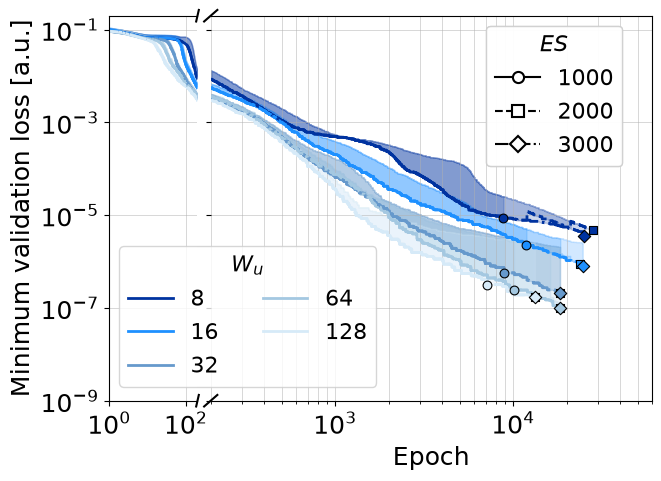}
            {}{north}{xshift=5pt,yshift=-5pt}
    \end{subfigure} \\[0.2cm]
    \begin{subfigure}[c]{0.9\linewidth}
        \cornerlabelimgshift[width=\linewidth]
            {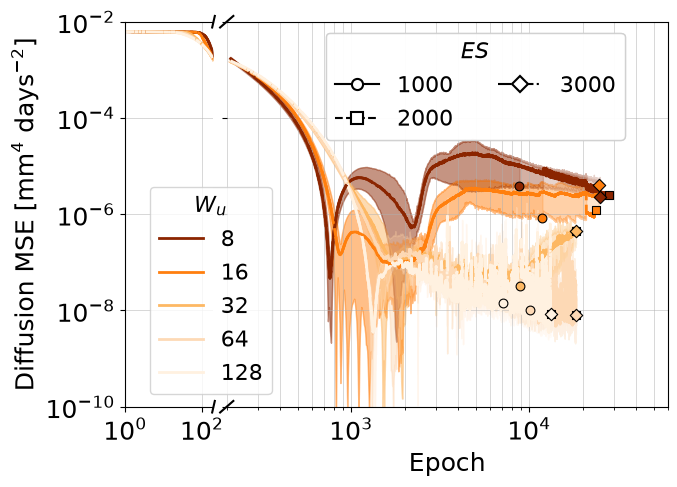}
            {}{north}{xshift=5pt,yshift=-5pt}
    \end{subfigure} \\[0.15cm]
    {\textbf{\small $W_u = 64$, $ES = 2000$}} \\[0.15cm]
    \begin{subfigure}[c]{0.9\linewidth}
        \cornerlabelimgshift[width=\linewidth]
            {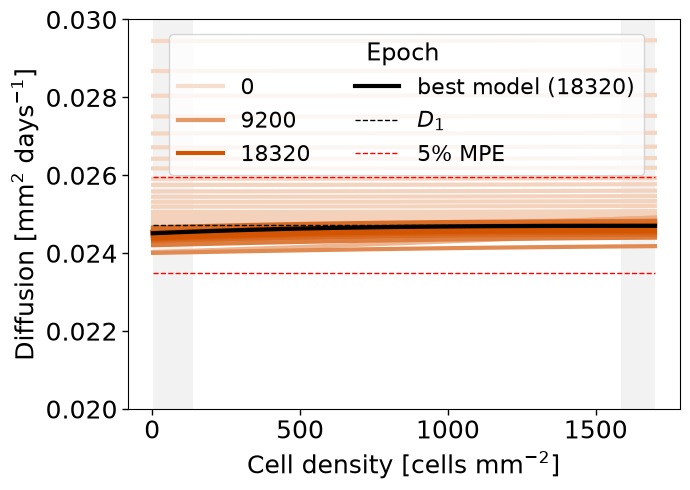}
            {}{north}{xshift=5pt,yshift=-5pt}
    \end{subfigure}
    \vspace*{0.15cm}
    \end{minipage}
};
\node[fill=white, text=black] at (box2.north) {\textbf{(b)} AF = \texttt{tanh}};
\end{tikzpicture}

&

\begin{tikzpicture}
\node[
    draw=lightgray, 
    line width=1pt, 
    rounded corners=5pt, 
    inner sep=0pt, 
    text width=\groupboxwidth, 
    align=center
] (box3) {
    \begin{minipage}{\linewidth}
    \centering
    \vspace*{0.35cm}
    \begin{subfigure}[c]{0.9\linewidth}
        \cornerlabelimgshift[width=\linewidth]
            {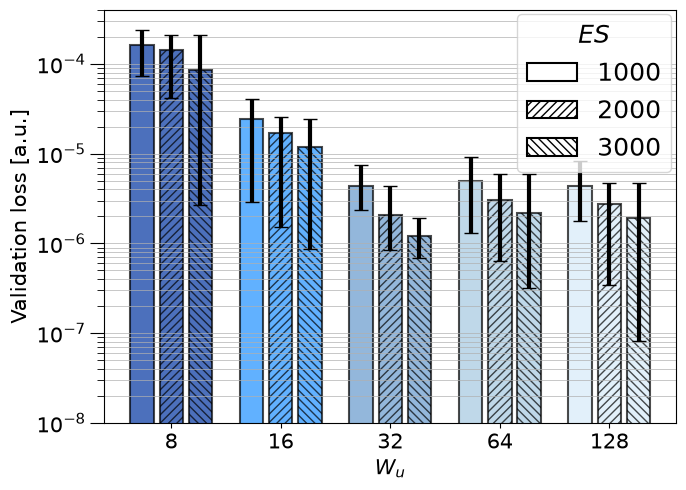}
            {}{north}{xshift=5pt,yshift=-5pt}
    \end{subfigure} \\[0.2cm]
    \begin{subfigure}[c]{0.9\linewidth}
        \cornerlabelimgshift[width=\linewidth]
            {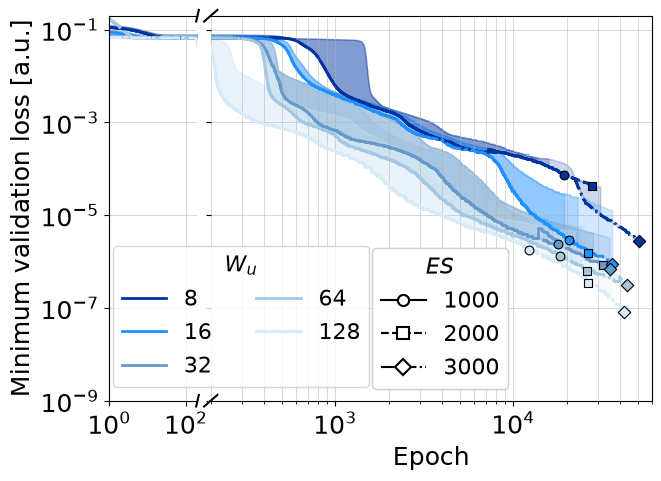}
            {}{north}{xshift=5pt,yshift=-5pt}
    \end{subfigure} \\[0.2cm]
    \begin{subfigure}[c]{0.9\linewidth}
        \cornerlabelimgshift[width=\linewidth]
            {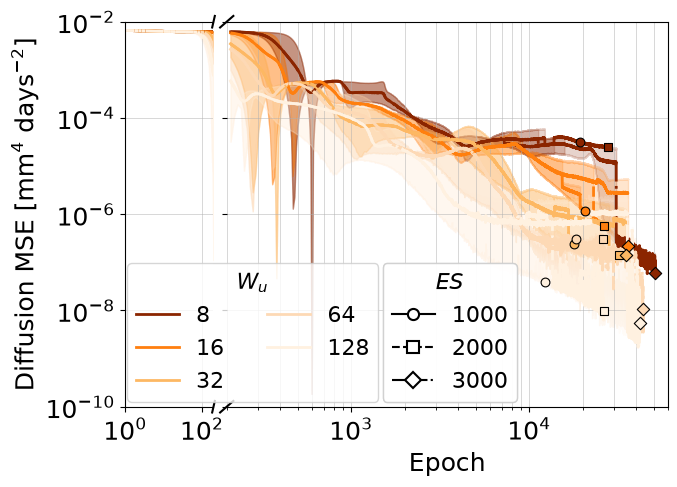}
            {}{north}{xshift=5pt,yshift=-5pt}
    \end{subfigure} \\[0.15cm]
    {\textbf{\small $W_u = 64$, $ES = 2000$}} \\[0.15cm]
    \begin{subfigure}[c]{0.9\linewidth}
        \cornerlabelimgshift[width=\linewidth]
            {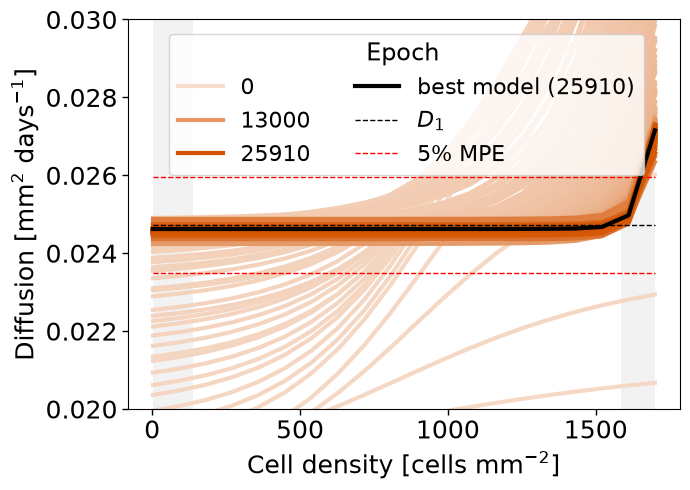}
            {}{north}{xshift=5pt,yshift=-5pt}
    \end{subfigure}
    \vspace*{0.15cm}
    \end{minipage}
};
\node[fill=white, text=black] at (box3.north) {\textbf{(c)} AF = \texttt{sigmoid}};
\end{tikzpicture}

\end{tabular}

\caption{\textbf{Performance across activation functions \texttt{SiLU}, \texttt{tanh} and \texttt{sigmoid} for synthetic Case~I.} Hyperparameters: $W_u \in \{8,16,32,64,128\}$, $W_D=2$, early-stopping patiences $ES\in \{1000,2000,3000\}$ and activation functions $\{\texttt{SiLU},\texttt{tanh},\texttt{sigmoid}\}$. For all running-minimum plots across training--validation (TV) splits: markers show the minimum for each $ES$, and shading shows the corresponding min--max range. For each of \textbf{(a)}--\textbf{(c)} \textit{top to bottom:} validation loss, running minimum validation loss, running minimum diffusion mean-squared error (MSE), and diffusion predictions across epochs for the first TV split of $W_u = 64, W_D=2$. Light-to-dark shading corresponds to increasing epochs, with the final best model indicated by a solid black line. The $\pm 5\%$ mean percentage error (MPE) bands are shown for reference. }
\label{fig:sm1-6}
\end{figure}

We evaluate three smooth activation
functions commonly used in PINN applications \cite{wang_auto-pinn_2023}:
(i) \texttt{sigmoid}: $\sigma(x) = \frac{1}{1+e^{-x}}$;
(ii) hyperbolic tangent (\texttt{tanh}):
$\texttt{tanh}(x) = \frac{e^{x}-e^{-x}}{e^{x}+e^{-x}}$;
(iii) \texttt{SiLU} (Sigmoid Linear Unit), also known as Swish \cite{Elfwing2018SiLU, ramachandran_searching_2017}:
$\texttt{SiLU}(x) = x\,\sigma(x) = \frac{x}{1+e^{-x}}$. Note that \texttt{tanh} and \texttt{sigmoid} are related by $\texttt{tanh}(x) = 2\sigma(2x)-1$, so they
span similar function classes; \texttt{SiLU} differs qualitatively in being unbounded
above and non-monotone, decreasing for $x < x^\ast$ to a global minimum $\texttt{SiLU}(x^\ast) \approx -0.278$ at $x^\ast \approx -1.278$, which can aid
gradient flow during training \cite{ramachandran_searching_2017}.
Since the PDE residual involves second spatial derivatives of the density network, and both the residual and the monotonicity penalties involve first derivatives of the diffusion and growth networks, the AFs must be at least twice continuously differentiable in $\mathrm{NN}_u$ and continuously differentiable in $\mathrm{NN}_D$ and $\mathrm{NN}_G$. Thus, piecewise-linear functions such as ReLU are excluded. In our setting, the primary bottleneck in the downstream recovery of the diffusion and growth terms is the quality of the density network's approximation of the density field: a poorly resolved density network propagates errors into the diffusion and growth networks regardless of their capacity. Because the density network is the limiting factor, we select the AF on it and share the choice across all three as a single hyperparameter. 
\\[1em]
As shown in Fig.~\ref{fig:sm1-6}a,b, \texttt{SiLU} and \texttt{tanh} achieve similar final validation losses and exhibit comparable loss trajectories for widths up to $W_u=32$. Beyond this point, the mean validation loss for \texttt{tanh} plateaus and the min--max spread across TV splits increases, whereas the validation loss for \texttt{SiLU} continues to decrease. In contrast, \texttt{sigmoid} performs substantially worse (Fig.~\ref{fig:sm1-6}c), with validation losses significantly larger and markedly greater variability across $W_u$. Furthermore, its performance does not improve for $W_u>32$. The diffusion predictions across epochs for $W_u = 64$, $ES = 2000$ are generally consistent with the validation losses. \texttt{SiLU} reaches its best model fastest and recovers a near-constant diffusion in close agreement with what \texttt{tanh} attains, though \texttt{tanh} requires nearly twice as many epochs (Fig.~\ref{fig:sm1-6}a,b). \texttt{sigmoid} is both the slowest and the least well-behaved: although close to constant through the central density range, its prediction turns sharply upward at high densities (Fig.~\ref{fig:sm1-6}c). This discrepancy in the diffusion predictions aligns with the poorer validation loss achieved compared to the other AFs.
\\[1em]
Turning to the diffusion MSE profiles, we see that \texttt{SiLU} exhibits a steady reduction in error with both epoch and $W_u$ (Fig.~\ref{fig:sm1-6}a). \texttt{tanh} attains comparably accurate diffusion estimates only at the largest widths, while its MSE trajectory is markedly non-monotonic: following an initial sharp decrease, the error rises again for narrower networks (Fig.~\ref{fig:sm1-6}b). This behaviour is not apparent from the validation-loss profiles alone, but can be deduced by examining the consistency of the diffusion predictions across epochs and relative to the final \texttt{SiLU} predictions (Fig.~\ref{fig:sm1-6}a,b). By comparison, \texttt{sigmoid} converges substantially more slowly, with the diffusion MSE remaining near $\mathcal{O}(10^{-4})$ for many epochs and training generally terminating only after considerably more epochs (Fig.~\ref{fig:sm1-6}c). This poorer convergence is consistent with the higher validation losses obtained with \texttt{sigmoid} relative to \texttt{tanh} and \texttt{SiLU}.
\\[1em]
Taken together, \texttt{SiLU} yields competitive final validation losses, consistent loss profiles across epochs, relatively fast convergence, and superior diffusion-function recovery for $W_u \ge 64$. While \texttt{tanh} achieves competitive final validation losses, its diffusion-function recovery improves only at larger widths and comes at the expense of heightened sensitivity to $ES$, whereas \texttt{sigmoid} is consistently the weakest AF across both validation loss and function recovery metrics. 
Consequently, we adopt \texttt{SiLU} as the default AF 
across all three networks.
\newpage
\section{Least-squares versus generalised least-squares data losses} \label{sec:SM2}

\begin{figure}[H]
    \centering
    \setlength{\groupboxwidth}{0.75\textwidth}

    \begin{tabular}{c c c}
        \begin{subfigure}{0.32\textwidth}
            \cornerlabelimgshift[width=\linewidth]
                {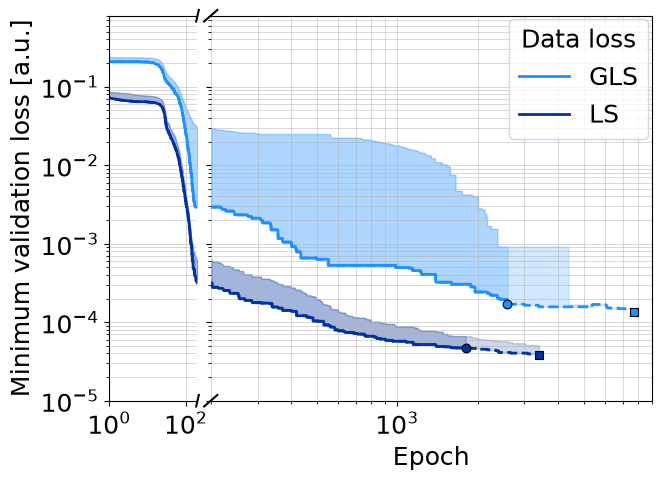}
                {(a)}{north}{xshift=5pt,yshift=10pt}
        \end{subfigure} &
        \begin{subfigure}{0.32\textwidth}
            \cornerlabelimgshift[width=\linewidth]
                {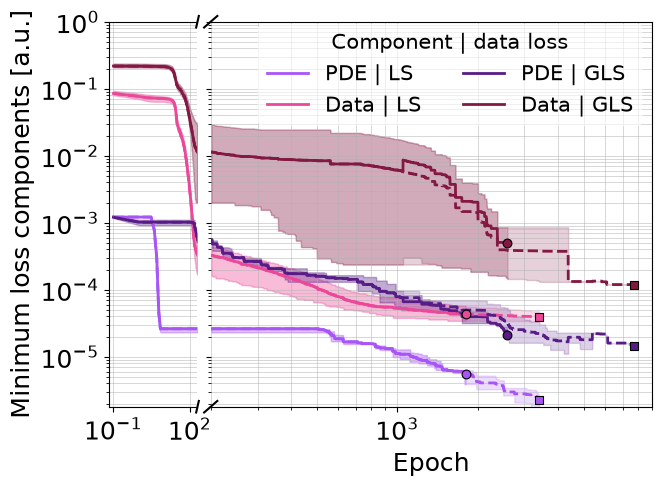}
                {(b)}{north}{xshift=5pt,yshift=10pt}
        \end{subfigure} &
        \begin{subfigure}{0.32\textwidth}
            \cornerlabelimgshift[width=\linewidth]
                {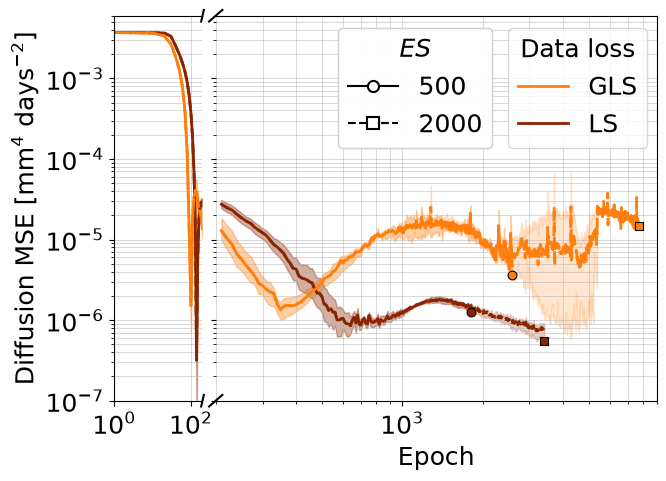}
                {(c)}{north}{xshift=5pt,yshift=10pt}
        \end{subfigure}
        \\

        \begin{subfigure}{0.32\textwidth}
            \cornerlabelimgshift[width=\linewidth]
                {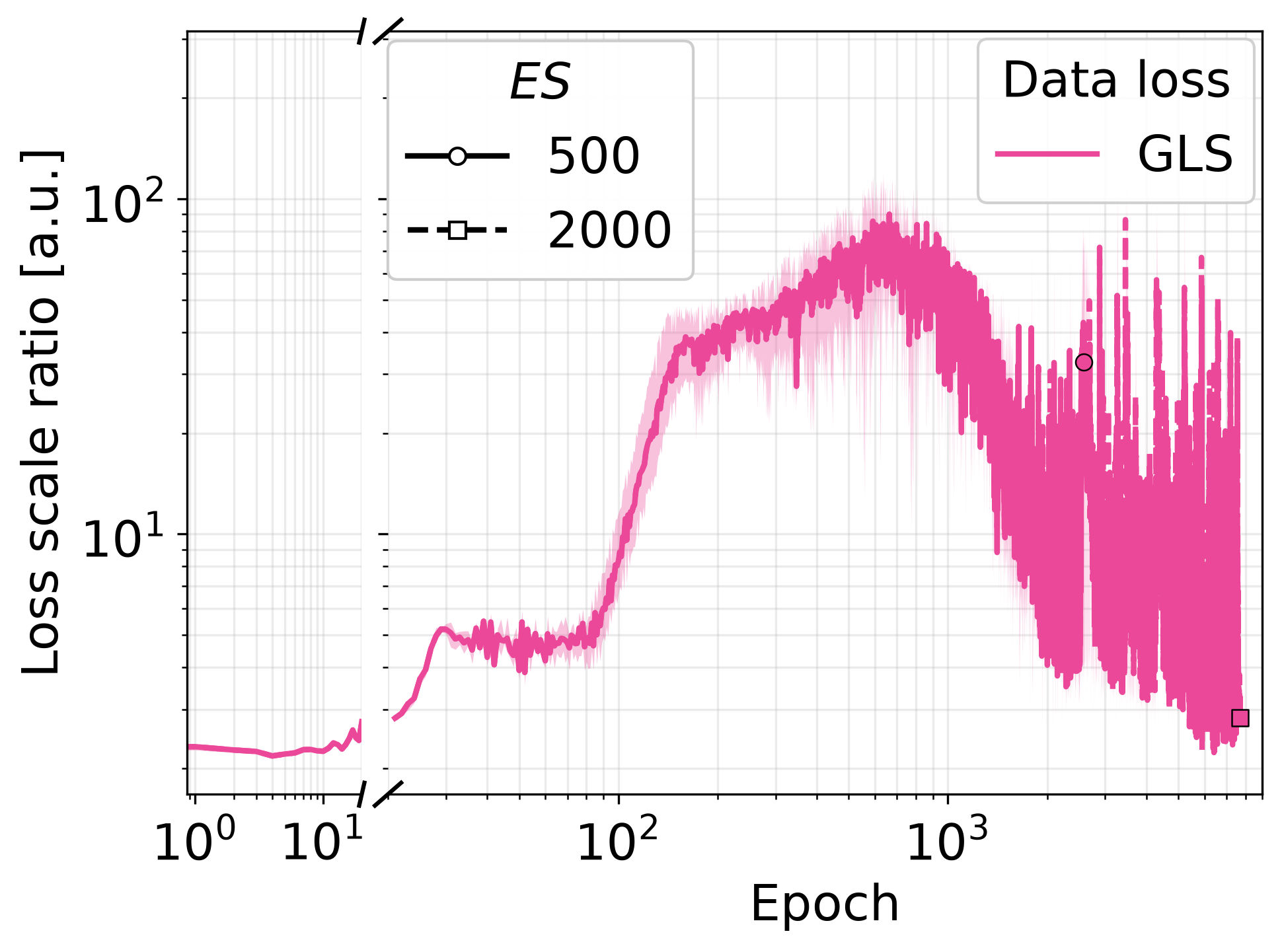}
                {(d)}{north}{xshift=5pt,yshift=10pt}
        \end{subfigure} &
        \begin{subfigure}{0.32\textwidth}
            \cornerlabelimgshift[width=\linewidth]
                {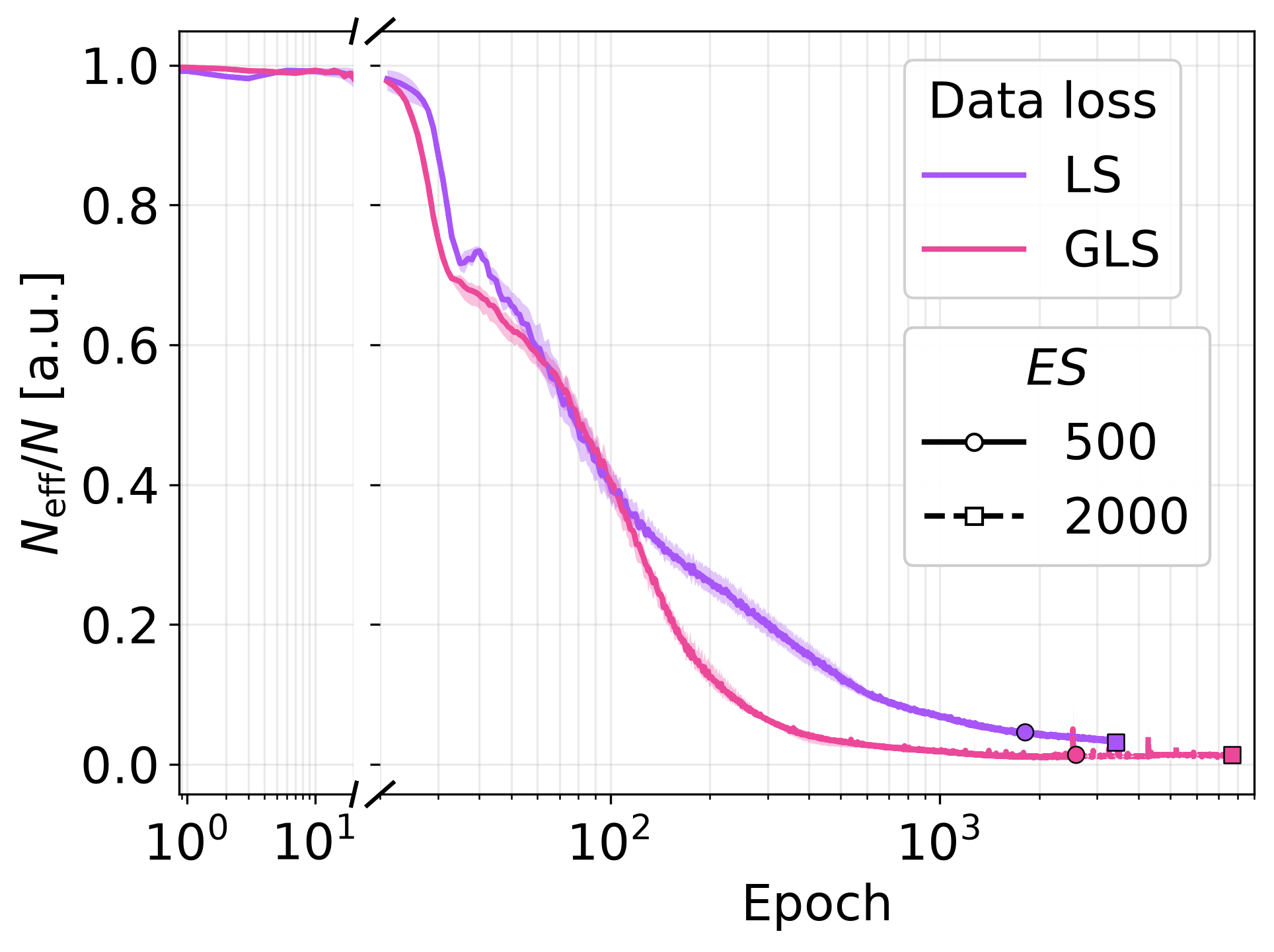}
                {(e)}{north}{xshift=5pt,yshift=10pt}
        \end{subfigure} &
        \begin{subfigure}{0.32\textwidth}
            \cornerlabelimgshift[width=\linewidth]{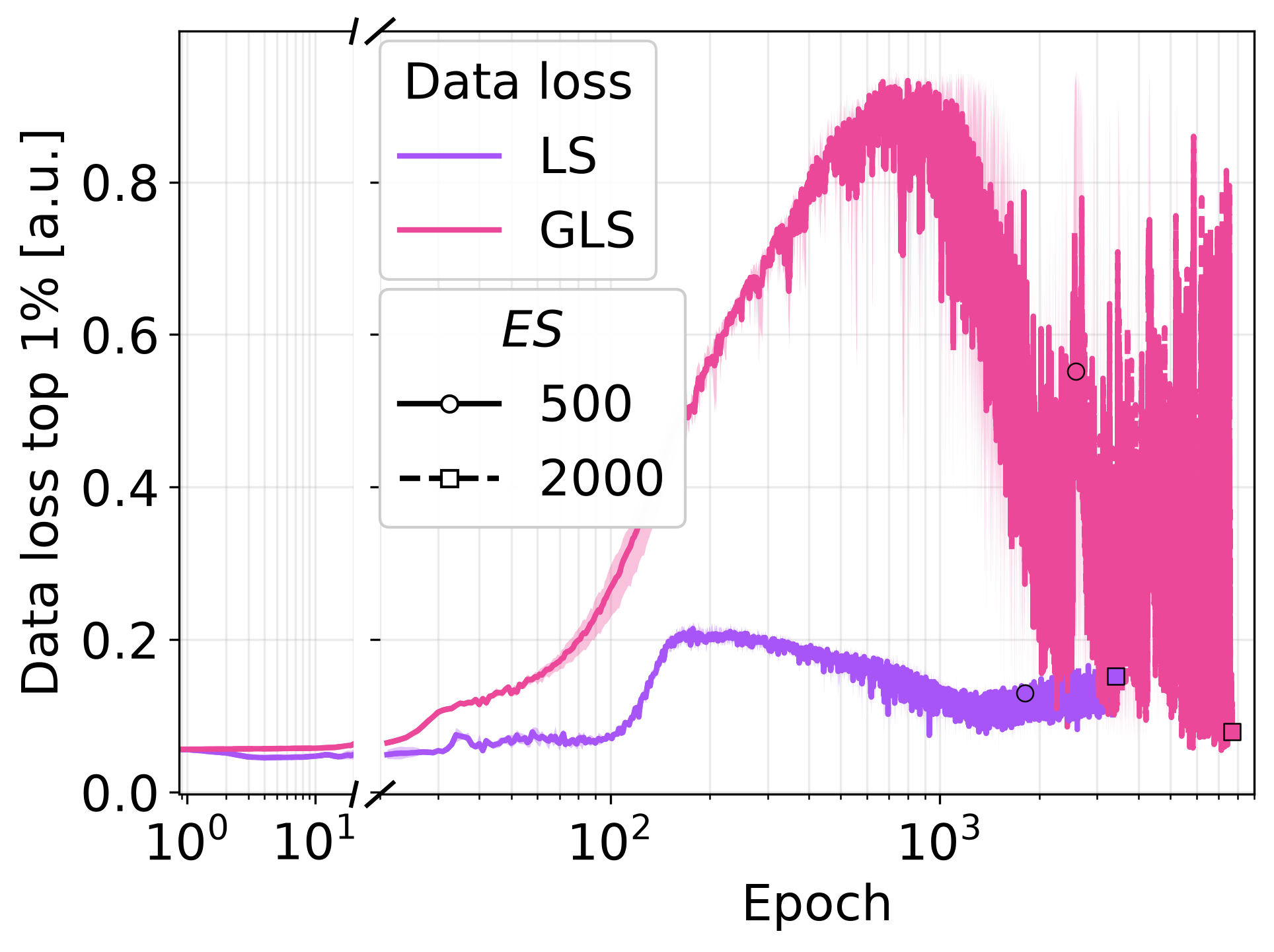}
                {(f)}{north}{xshift=5pt,yshift=10pt}
        \end{subfigure}
    \end{tabular}

    \begin{tikzpicture}
    \node[
        draw=lightgray,
        line width=1pt,
        rounded corners=5pt,
        inner xsep=0pt,
        inner ysep=2pt,
        text width=\groupboxwidth,
        align=center
    ] (box3) {
        \begin{minipage}{\linewidth}
        \centering
        \vspace{-0.1cm}
        \begin{tabular}{@{}c @{\hspace{1cm}} c@{}}
            \begin{subfigure}[c]{0.4\linewidth}
                \cornerlabelimgshift[width=\linewidth]
                    {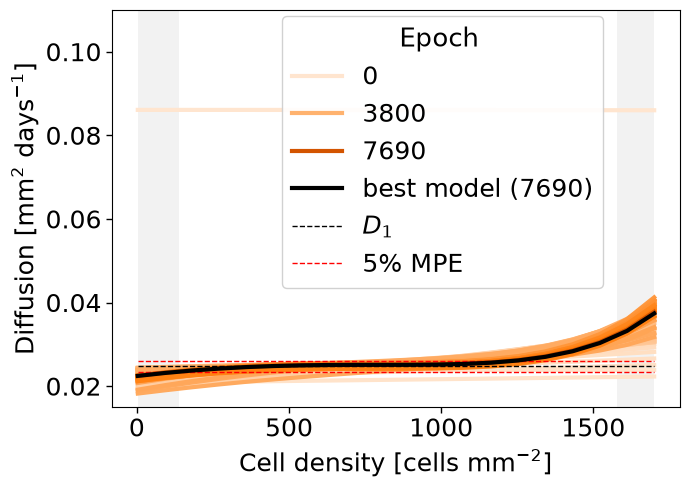}
                    {}{north}{xshift=5pt,yshift=10pt}
            \end{subfigure} &
            \begin{subfigure}[c]{0.4\linewidth}
                \cornerlabelimgshift[width=\linewidth]
                    {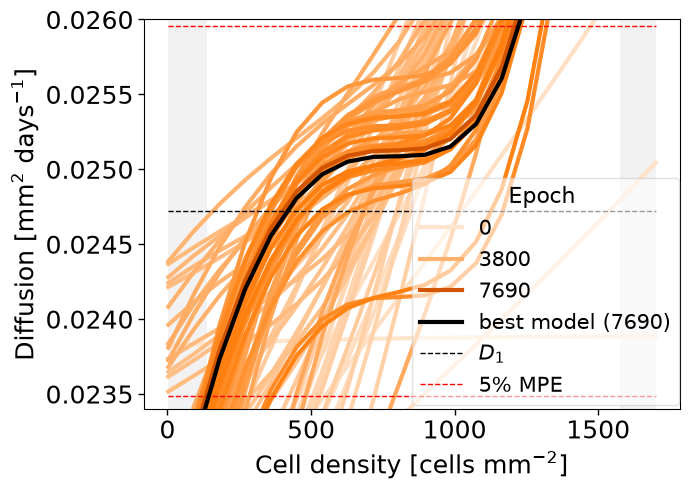}
                    {}{north}{xshift=5pt,yshift=10pt}
            \end{subfigure}
        \end{tabular}
        \end{minipage}
    };
    \node[fill=white, text=black] at (box3.north) {\textbf{(g) GLS}};
    \end{tikzpicture}

    \vspace{0.2cm}

    \begin{tikzpicture}
    \node[
        draw=lightgray,
        line width=1pt,
        rounded corners=5pt,
        inner xsep=0pt,
        inner ysep=2pt,
        text width=\groupboxwidth,
        align=center
    ] (box4) {
        \begin{minipage}{\linewidth}
        \centering
         \vspace{-0.05cm}
        \begin{tabular}{@{}c @{\hspace{1cm}} c@{}}
            \begin{subfigure}[c]{0.4\linewidth}
                \cornerlabelimgshift[width=\linewidth]
                    {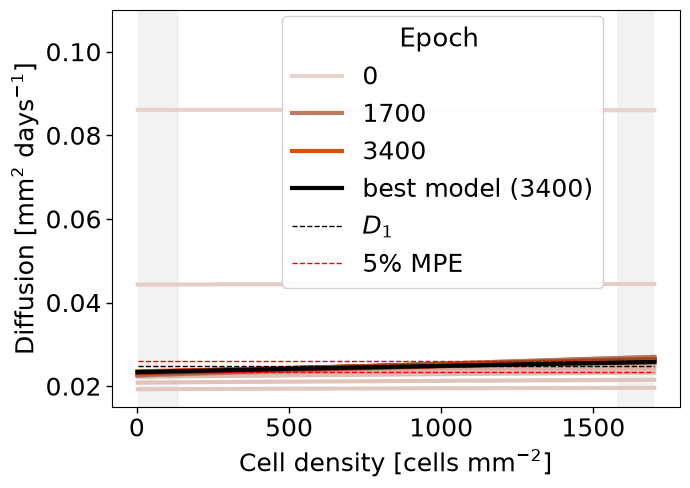}
                    {}{north}{xshift=5pt,yshift=10pt}
            \end{subfigure} &
            \begin{subfigure}[c]{0.4\linewidth}
                \cornerlabelimgshift[width=\linewidth]
                    {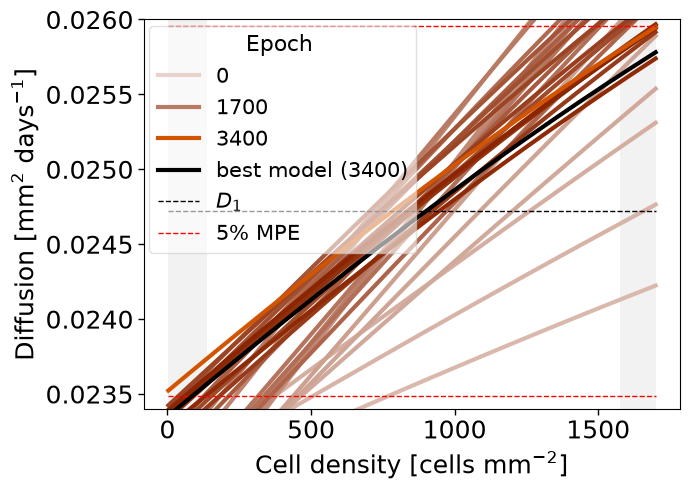}
                    {}{north}{xshift=5pt,yshift=10pt}
            \end{subfigure}
        \end{tabular}
        \end{minipage}
    };
    \node[fill=white, text=black] at (box4.north) {\textbf{(h) LS}};
    \end{tikzpicture}

\caption{
\textbf{Performance of generalised least squares (GLS) loss versus least squares (LS) loss for synthetic Case~I with 1\% mean percentage error noise.} 
Ground truth: $D_1(u)$. Hyperparameters: $W_u = 64$, $W_D = 2$, early-stopping patiences $ES \in \{500,2000\}$. LS and GLS correspond to $\mathcal{L}_{\text{data}}$ and $\mathcal{L}_{\text{GLS}}$ objectives, respectively. \textbf{(a)}--\textbf{(f)} consist of $3$ training--validation (TV) splits: markers show the minimum for each $ES$, and shading shows the corresponding min--max range. 
\textbf{(a)} running minimum validation loss.
\textbf{(b)} running minimum validation data and PDE loss components.
\textbf{(c)} running minimum diffusion-function MSE.
\textbf{(d)} ratio $\mathcal{L}_{\text{GLS}}/\mathcal{L}_{\text{data}}$ across epochs.
\textbf{(e)} effective sample size $N_{\text{eff}}$ as a fraction of total sample size $N$, computed from the weights $w_k^{\text{LS}} = (\hat{u}_k^{\text{LS}})^{-2}$ and $w_k^{\text{GLS}} = (\hat{u}_k^{\text{GLS}})^{-2}$ (data-generating $\gamma=1$) for the LS- and GLS-trained models, respectively. \textbf{(f)} fraction of the total data objective formed by the largest 1\% of pointwise contributions.
Blocks \textbf{(g)}--\textbf{(h)} show, for the first TV split, diffusion predictions sampled every 100 epochs with light-to-dark shading corresponding to increasing epochs, with the final best model indicated by a solid black line. The $\pm5\%$ mean percentage error (MPE) bands are shown for reference.
}
\label{fig:sm2}
\end{figure}

Here we compare the generalised least-squares (GLS) objective of \textcite{Lagergren2020} to the standard least-squares (LS) loss. To do this we use synthetic Case I with the addition of noise ($\gamma = 1$). The noise level is set to 1\% MPE. We use 
$W_{u} = 64$, $W_D = 2$, and early-stopping patiences $ES \in \{500, 2000\}$. 
\\[1em]
From a statistical perspective, the GLS loss is the
natural choice under the noise model in
Eq.~\eqref{eq:err_model}, assuming that the variance-scaling coefficients \(\omega_{i,j,s}\) are
uniform; see Section~\ref{sec:supmat_data_loss} for details.
Nevertheless, in practice we find that the standard LS loss often yields better performance. To understand why this is the case, we compare how each objective weights individual data points and how these weights can affect optimisation. 
\\[1em]
The density data lie on the grid (written for $d=2$ as in the main text; Case~I here has $d=1$)
\[
(x_{1,i},x_{2,j},t_s) \in \mathcal{G},
\qquad
(i,j,s) \in
\{1,\ldots,n_{x_1}\}
\times
\{1,\ldots,n_{x_2}\}
\times
\{1,\ldots,n_t\},
\]
with $N = n_{x_1} n_{x_2} n_t$ data points in total. As defined in Eq.~\eqref{eq:data_loss}, the LS data loss for $2\text{D}{+}t$ is
\begin{align*}
\mathcal{L}_{\text{data}}(\hat{u})
=
\frac{1}{N}
\sum_{i=1}^{n_{x_1}}
\sum_{j=1}^{n_{x_2}}
\sum_{s=1}^{n_t}
\left(
u_{\text{data}}^{i,j,s}
-
\hat{u}(x_{1,i},x_{2,j},t_s)
\right)^2,
\end{align*}
and the GLS loss is
\begin{align*}
\mathcal{L}_{\text{GLS}}(\hat{u})
=
\frac{1}{N}
\sum_{i=1}^{n_{x_1}}
\sum_{j=1}^{n_{x_2}}
\sum_{s=1}^{n_t}
\left(
\frac{
u_{\text{data}}^{i,j,s} -
\hat{u}(x_{1,i},x_{2,j},t_s)
}
{\hat{u}(x_{1,i},x_{2,j},t_s)^\gamma}
\right)^2.
\end{align*}
In practice, clipping is required to ensure that
\(\hat{u}^\gamma \neq 0\) at all sampled points, and \(\gamma\) must typically
be estimated from data. To isolate the effect of the loss itself, we instead
consider the idealised setting in which \(\gamma\) is known \textit{a priori}, and compare
training under \(\mathcal{L}_{\text{data}}\) and
\(\mathcal{L}_{\text{GLS}}\). To interpret the difference between these objectives, we flatten the grid coordinates $(x_{1,i},x_{2,j},t_s) \in \mathcal{G}$ into a single row-major index $k = (i - 1)n_{x_2}n_t + (j - 1)n_t + s$, defining $\hat{u}_k = \hat{u}(x_{1,i},x_{2,j},t_s)$ and $u_{\text{data}}^k = u_{\text{data}}^{i,j,s}$. The GLS loss for a single point $k$ can then be written as
\[
\left( \frac{u_{\text{data}}^k - \hat{u}_k}{\hat{u}_k^\gamma} \right)^2 = w_k \left( u_{\text{data}}^k - \hat{u}_k \right)^2, \qquad w_k = \hat{u}_k^{-2\gamma}.
\]
Thus, GLS induces prediction-dependent weights: each pointwise error's contribution to the total objective depends on the current prediction $\hat{u}_k$, with smaller predictions receiving greater weight.
\\[1em]
We use three diagnostics based on these weights. First, the ratio $\mathcal{L}_{\text{GLS}}/\mathcal{L}_{\text{data}}$ measures the relative scale of the weighted and unweighted data losses. Second, the effective sample fraction $N_{\mathrm{eff}}/N$ quantifies how broadly a set of weights is distributed across the dataset. To compare the LS- and GLS-trained models on an equal footing, we evaluate it for both models using the data-generating exponent $\gamma = 1$, i.e.\ with weights
\[
w_k^{\text{LS}} = \bigl(\hat{u}_k^{\text{LS}}\bigr)^{-2},
\qquad
w_k^{\text{GLS}} = \bigl(\hat{u}_k^{\text{GLS}}\bigr)^{-2},
\]
where $\hat{u}_k^{\text{LS}}$ and $\hat{u}_k^{\text{GLS}}$ are the density predictions of the models trained with $\mathcal{L}_{\text{data}}$ and $\mathcal{L}_{\text{GLS}}$, respectively. For either set of weights $\{w_k\}_{k=1}^{N}$, the effective sample size is defined by
\[
N_{\mathrm{eff}}
=
\frac{\left(\sum_{k=1}^{N} w_k\right)^2}{\sum_{k=1}^{N} w_k^2}
=
\frac{1}{\sum_{k=1}^{N} \bar{w}_k^2},
\qquad
\bar{w}_k = \frac{w_k}{\sum_{k'=1}^{N} w_{k'}}.
\]
Accordingly, \(N_{\mathrm{eff}}=N\) when all points are weighted equally,
whereas \(N_{\mathrm{eff}}\approx 1\) when nearly all weight is concentrated on
a single point, so that \(N_{\mathrm{eff}}/N\) measures how much of the dataset is
effectively contributing. Third, we compute the fraction of each data loss contributed by its largest \(1\%\) of pointwise contributions.
\\[1em]
Figure~\ref{fig:sm2}a shows that the choice of data loss substantially 
affects the absolute scale of the total validation loss, and Fig.~\ref{fig:sm2}b shows that this difference is 
driven by the data loss component. Because we train on normalised densities for Case I, the density predictions predominantly satisfy $\hat{u}_k \leq 1$; consequently, the GLS weights satisfy $w_k^{\text{GLS}} \geq 1$ and inflate the weighted data term relative to the PDE component. Furthermore, the min--max shading across TV splits in Fig.~\ref{fig:sm2}a reveals a clear difference between the loss functions: under GLS, one model maintains a markedly higher validation loss and terminates prematurely, indicating greater sensitivity to the specific split.
\\[1em]
The running-minimum diffusion-function MSE in Fig.~\ref{fig:sm2}c mirrors the overall behaviour of the validation loss. For both early-stopping patience values, $\mathcal{L}_{\text{data}}$ yields better diffusion-function predictions than $\mathcal{L}_{\text{GLS}}$. In both the GLS and LS cases, the learned diffusion initially passes through the correct scale of the true constant diffusion $D_1$, producing a transiently small MSE. Subsequently, however, the diffusion predictions become meaningful only as the density fit stabilises. The divergence in the diffusion-function MSE during these later stages of training highlights the impact of the chosen data loss on the learned diffusion.
\\[1em]
Figure~\ref{fig:sm2}d shows that the ratio $\mathcal{L}_{\text{GLS}}/\mathcal{L}_{\text{data}}$ is initially of order one but grows to values in the tens around 200 epochs, indicating that the GLS loss becomes substantially larger than the unweighted LS term. In Fig.~\ref{fig:sm2}e, $N_{\mathrm{eff}}/N$ decreases for both models as the density fit resolves the low-density regions, but beyond approximately 100 epochs it falls further for the GLS-trained model, coinciding with the jump in the loss ratio. Hence, the GLS-trained model produces predictions whose weights are more unequally distributed, with a small number of low-density points carrying most of the weight. Figure~\ref{fig:sm2}f confirms this concentration of the loss on a few data points, with the largest $1\%$ of pointwise GLS contributions accounting for a large proportion of the total loss. Figure~\ref{fig:sm2}g,h illustrates the effect on the diffusion predictions: under $\mathcal{L}_{\text{GLS}}$, the predictions change more rapidly across epochs and settle further from the ground truth, whereas $\mathcal{L}_{\text{data}}$ yields more accurate final diffusion-function predictions.
\\[1em]
Taken together, these results suggest that, relative to the LS loss, GLS changes the optimisation
geometry in a significant way. Rather than simply accounting for heteroscedastic
noise, the prediction-weighted GLS increasingly concentrates the data-fitting
signal on low-density regions where \(\hat{u}\) is small and
\(w_k\) is therefore large. In the coupled BINN setting, this
concentration appears to be detrimental: the density network must provide a
globally good approximation of \(u(\mathbf{x},t)\) to support both the PDE residual and
accurate learning of the diffusion (and growth) terms, whereas an overly concentrated data
loss can distort that global fit. By contrast, LS distributes the data-fitting
pressure more uniformly across the density surface, and this more even
conditioning appears to yield a more favourable optimisation in practice.

\clearpage
\newpage
\section{Implementation details}\label{sec:SM3}

\subsection{Noise implementation}
All synthetic noise is implemented according to the following procedure, stated here for the $2\text{D}{+}t$ case; the $1\text{D}{+}t$ case follows analogously.
The noisy
observations are generated according to
\[
u_{\text{data}}^{i,j,s} =
\max\!\left( u(x_{1,i}, x_{2,j}, t_s) + \omega\, u^{\gamma}(x_{1,i}, x_{2,j}, t_s)\, \varepsilon_{i,j,s},\; 0 \right),
\]
following the error model of Eq.~\eqref{eq:err_model}, with the truncation at zero
enforcing nonnegativity of the density and the variance-scaling coefficient taken to be
constant across all observations, i.e., $\omega_{i,j,s} \equiv \omega$.

\paragraph{Noise mean percentage error.}
To avoid division by (near-)zero densities at the domain boundary, the relative error is
evaluated only over the index set
\[
\mathcal{M} = \bigl\{ (i,j,s) : u(x_{1,i}, x_{2,j}, t_s) \ge u_{\text{floor}} \bigr\},
\qquad u_{\text{floor}} = 10^{-3},
\]
with $M = |\mathcal{M}|$ the number of retained observations.
For a prescribed noise level $\tau$ in $2\text{D}{+}t$, expressed as an (absolute) mean percentage error (MPE), the scalar $\omega$ is obtained numerically, for each realisation of the noise, such that the achieved MPE lies within a tolerance $\delta$ of the target, $\lvert \mathrm{MPE} - \tau \rvert \leq \delta,$ where
\begin{align*}
\mathrm{MPE}
&= \frac{100}{M}
\sum_{(i,j,s)\in\mathcal{M}}
\left|
\frac{
u(x_{1,i},x_{2,j},t_s)
-
u_{\mathrm{data}}^{i,j,s}
}{
u(x_{1,i},x_{2,j},t_s)
}
\right|
\\
&=
\frac{100}{M}
\sum_{(i,j,s)\in\mathcal{M}}
\left|
\frac{
u(x_{1,i},x_{2,j},t_s)
-
\max\left(
u(x_{1,i},x_{2,j},t_s)
+
\omega\,u^{\gamma}(x_{1,i},x_{2,j},t_s)\,
\varepsilon_{i,j,s},
\,0
\right)
}{
u(x_{1,i},x_{2,j},t_s)
}
\right|,
\end{align*}
and $\delta = 10^{-4}$.

\subsection{Ground-truth error metrics} \label{sec:SM:metrics}

Because the synthetic data are generated from known coefficient functions, the learned
functions can be compared directly against them. The metrics defined here quantify that
comparison. They require the ground truth, are therefore unavailable in an experimental
setting, and play no part in the selection workflow of Section~\ref{sec:pipeline}; they are
reported alongside the ground-truth-free diagnostics only to verify the selections that
workflow makes.

Let $f$ denote a ground-truth coefficient function ($D$ or $G$) and $\hat{f}$ its learned surrogate, which corresponds to the network output. Both
are evaluated on a fixed, equally spaced grid of $n_{\text{MSE}}$ scaled density values
\[
\{\tilde{u}_q\}_{q=1}^{n_{\text{MSE}}}
\subseteq [\tilde{u}_{\min},\tilde{u}_{\max}
],
\]
where $n_{\text{MSE}}=20$, and $\tilde{u}_{\min}$ and
$\tilde{u}_{\max}$ are the minimum and maximum of the normalised densities, so that the comparison is restricted to the density range the data support. Note that the growth terms $G_1$ and $G_3$ drive the density
to $\tilde{u}_{\max} \approx 2.9$ and $2.5$, respectively, whereas in all other cases $\tilde{u}_{\max} \leq 1$. The mean-squared error is
\begin{equation}
\mathrm{MSE}(\hat{f})
=
\frac{1}{n_{\text{MSE}}}\sum_{q=1}^{n_{\text{MSE}}}\bigl(f_{\text{scale}} \hat{f}(\tilde{u}_q) - f(\tilde{u}_q)\bigr)^2,
\label{eq:fn_mse}
\end{equation}
where $f_{\text{scale}}$ is the respective scaling factor ($d_{\mathrm{scale}}$ or $g_{\mathrm{scale}}$; see Section~\ref{sec:data preprocessing}). Therefore, the MSE carries the squared units of $f_{\text{scale}}$ ($\mathrm{mm}^4\,\mathrm{days}^{-2}$ for diffusion,
$\mathrm{days}^{-2}$ for growth). No weighting or restriction by density occupancy is applied: every grid point contributes equally, and the
$5$th and $95$th density percentiles marked in the profile figures are indicative only. Since the units of Eq.~\eqref{eq:fn_mse} differ between the two functions and the ground-truth magnitudes vary across the forms in
Table~\ref{tab:expressions}, MSE values are not directly comparable.
For context, we use the uniformly rescaled function
$f_p=(1+p/100)f$, whose mean percentage error is
\begin{align*}
\mathrm{MPE}(f_p)
= \frac{100}{n_{\mathcal{Q}}}\sum_{q\in\mathcal{Q}}
\left|\frac{f_p(\tilde{u}_q)-f(\tilde{u}_q)}{f(\tilde{u}_q)}\right|
= \frac{100}{n_{\mathcal{Q}}}\left(\frac{\lvert p \rvert}{100}n_{\mathcal{Q}}\right)
=  p,
\end{align*}
where $\mathcal{Q}=\{q:f(\tilde{u}_q)\neq 0\}$, of size $n_{\mathcal{Q}}$, since terms with $f(\tilde{u}_q)=0$ are dropped from the summation. The mean-squared error is given by
\begin{equation}
\mathrm{MSE}(f_p)
= \frac{1}{n_{\text{MSE}}}\sum_{q=1}^{n_{\text{MSE}}}\bigl(f_p(\tilde{u}_q) - f(\tilde{u}_q)\bigr)^2 = \left(\frac{p}{100}\right)^2\frac{1}{n_{\text{MSE}}} \sum_{q=1}^{n_{\text{MSE}}}f^2(\tilde{u}_q).
\label{eq:fp_mse}
\end{equation}
In MSE plots, a line labelled $p\%$ marks $\mathrm{MSE}(f_p)$ from Eq.~\eqref{eq:fp_mse}.
In plots of the learned function against $u$, the $\pm p\%$ band
is given by $(1\pm p/100)f(u)$.

\subsection{Data preprocessing} \label{sec:data preprocessing}
One should non-dimensionalise three components relevant to BINN training:
(i) the inputs, (ii) the loss functions, and (iii) the outputs.
The input scaling fixes the characteristic constants used throughout the non-dimensionalisation, and the loss functions must then be formulated consistently with this scaling. The outputs require no separate treatment: they are non-dimensional by construction once (i) and (ii) are applied, and are rescaled to physical units after training via the characteristic scales (e.g.\ $d_{\text{scale}}\hat{D}$ and $g_{\text{scale}}\hat{G}$). Accordingly, we address item (i) below and item (ii) in Section~\ref{sec:supmat:losses}.

Let $\mathrm{L}_m$ denote characteristic length scales in each spatial direction $x_m$, $m=1,\dots,d$. Let $T$ be a representative time scale, $U$ a characteristic magnitude of the cell density $u(\mathbf{x},t)$, and let $d_{\text{scale}}$ and $g_{\text{scale}}$ denote characteristic values of the diffusion and growth functions $D(\cdot)$ and $G(\cdot)$, respectively. In general, $U$ is chosen as a characteristic density of the observed data; for the scratch-assay setting considered here the natural choice is the biological density scale $U = K = 1700\text{ cells mm}^{-2}$~\cite{Jin2016} used to define the ground-truth functions in Table~\ref{tab:expressions}. Under these assumptions, the physical variables satisfy
\begin{align*}
x_m &= \mathcal{O}(\mathrm{L}_m), \quad m=1,\dots,d, &
t &= \mathcal{O}(T), \nonumber\\
u_{\text{data}} &= \mathcal{O}(U), &
D(\cdot) &= \mathcal{O}(d_{\text{scale}}), &
G(\cdot) &= \mathcal{O}(g_{\text{scale}}).
\end{align*}
\noindent
We introduce dimensionless variables (denoted by tildes) via
\begin{equation}
\tilde{x}_m = \frac{x_m}{\mathrm{L}_m}, \quad
\tilde{t} = \frac{t}{T}, \quad
\tilde{u}_{\text{data}} = \frac{u_{\text{data}}}{U}, \quad
\tilde{u}(\tilde{\mathbf{x}},\tilde{t}) = \frac{u(\mathbf{x},t)}{U},
\label{eq:nd_vars}
\end{equation}
 where
$\mathbf{x}=(x_1,\dots,x_d)$, $\mathbf{L}^{-1}=(\mathrm{L}_1^{-1},\dots,\mathrm{L}_d^{-1})$, or, in vector form, $\tilde{\mathbf{x}} = \mathbf{L}^{-1}\circ\mathbf{x}$, for the Hadamard (componentwise) product $\circ$. We assume the measurement noise is sufficiently small that the true density $u(\cdot)$ is of the same order as the observed data $u_{\text{data}}$.

\noindent We further define the rescaled diffusion and growth functions
\begin{equation*}
\tilde{D}(\tilde{u}) = \frac{D(u)}{d_{\text{scale}}}, \qquad
\tilde{G}(\tilde{u}) = \frac{G(u)}{g_{\text{scale}}},
\end{equation*}
which ensures all quantities are $\mathcal{O}(1)$. The derivative operators transform according to
\begin{align}
\frac{\partial}{\partial t}
&= \frac{1}{T} \frac{\partial}{\partial \tilde{t}},
\nonumber\\
\nabla
&= \left(\frac{\partial}{\partial x_1},\dots,\frac{\partial}{\partial x_d}\right)
= \mathbf{L}^{-1} \circ \tilde{\nabla},
\qquad
\tilde{\nabla}
= \left(\frac{\partial}{\partial \tilde{x}_1},\dots,\frac{\partial}{\partial \tilde{x}_d}\right).
\label{eq:space_deriv}
\end{align}
Substituting Eq.~\eqref{eq:nd_vars}--Eq.~\eqref{eq:space_deriv} into the dimensional Eq.~\eqref{eq:reaction_diff_eq_2d} yields the non-dimensional form
\begin{align*}
\frac{\partial \tilde{u}}{\partial \tilde{t}}
=
d_{\text{scale}}\, T \, \left( \mathbf{L}^{-2} \circ
\tilde{\nabla} \right) \cdot
\bigl(
\tilde{D}(\tilde{u}) \, \tilde{\nabla} \tilde{u}
\bigr)
+
T g_{\text{scale}} \,
\tilde{G}(\tilde{u}) \, \tilde{u}.
\end{align*}
If all spatial lengths are equal $\mathrm{L}_1,\dots, \mathrm{L}_d = \mathrm{L}$ then we can instead use

\begin{align}
\frac{\partial \tilde{u}}{\partial \tilde{t}}
=
d_{\text{scale}}\, \frac{T}{\mathrm{L}^2} \ 
\tilde{\nabla} \cdot
\bigl(
\tilde{D}(\tilde{u}) \, \tilde{\nabla} \tilde{u}
\bigr)
+
T g_{\text{scale}} \,
\tilde{G}(\tilde{u}) \, \tilde{u}.
\label{eq:nd_pde_upd}
\end{align}
We approximate $\tilde{u}(\tilde{\mathbf{x}},\tilde{t})$, $\tilde{D}(\tilde{u})$, and $\tilde{G}(\tilde{u})$ by the MLP surrogates
$\hat{u}(\tilde{\mathbf{x}},\tilde{t})$, $\hat{D}(\tilde{u})$, and $\hat{G}(\tilde{u})$.

\subsubsection{Partial preprocessing}
If $\mathrm{L}_1, \dots, \mathrm{L}_d,T = \mathcal{O}(1)$ we can incorporate these scales back into the derivative operators and proceed with the original $x_1, x_2,\dots, x_d$, and $t$ in Eq.~\eqref{eq:nd_pde_upd} without introducing training inefficiency. This leads to

\begin{equation*}
\frac{\partial \tilde{u}}{\partial t}
=
d_{\text{scale}} 
\nabla \cdot
\bigl(
\tilde{D}(\tilde{u}) \, \nabla \tilde{u}
\bigr)
+
g_{\text{scale}} \,
\tilde{G}(\tilde{u}) \, \tilde{u},
\end{equation*}
where we now approximate $\tilde{u}(\mathbf{x},t)$, $\tilde{D}(\tilde{u})$, and $\tilde{G}(\tilde{u})$ by the MLP surrogates
$\hat{u}(\mathbf{x},t)$, $\hat{D}(\tilde{u})$, and $\hat{G}(\tilde{u})$, respectively. Moreover, in our simulated experiments we generate the data directly in the model's dimensionless density $\tilde{u} = u/K = \mathcal{O}(1)$, where $K$ is the biological density scale introduced above (Table~\ref{tab:expressions}). The data passed to the pipeline are therefore already $\mathcal{O}(1)$, so we take the pipeline's characteristic density to be $U=1$ and apply no further rescaling (i.e., $\tilde{u}_{\text{data}}=u_{\text{data}}$). The training variable then coincides with the model's dimensionless density. Equivalently, the $1/K$ normalisation is absorbed at data-generation time, and $U=1$ is simply the special case of the general choice $U=K$ that would be made for un-normalised (e.g.\ experimental) data. Since both diffusion and growth depend only on $\tilde{u}=u/U$, the factor $U$ cancels from the governing equation and does not affect the recovered diffusion and growth functions; we may therefore rescale to physical units via $u = \tilde{u}\cdot K$ purely to dimensionalise the density axis post hoc.

\subsection{Loss definitions} \label{sec:supmat:losses}
To make the loss functions directly comparable with the main text, we set $d=2$. However, extending this to $d=1$ or $d=3$ is straightforward. In this subsection, we write the general forms in the non-dimensionalised coordinates of Section~\ref{sec:data preprocessing}, but in the implementation, all
losses are evaluated in the original quantities $x_1, x_2$, $t$, and $u$, which is possible without rescaling because $\mathrm{L}$, $T$, and $U$ are
$\mathcal{O}(1)$. 
\subsubsection{Data loss} 
\label{sec:supmat_data_loss}
\paragraph{Data locations.}
As outlined in Section~\ref{sec:bg}, observed density measurements are available at discrete spatiotemporal locations
\begin{align*}
\mathcal{G}
=
\left\{
(x_{1,i},x_{2,j},t_s)
\right\}_{%
i=1,\dots,n_{x_1};\,
j=1,\dots,n_{x_2};\,
s=1,\dots,n_t
}
\end{align*}
with corresponding observations $u_{\text{data}}^{i,j,s}$.
These points are fixed by the experimental design and form the points where the data loss is enforced. 

\paragraph{Loss function.}
Under the noise model in Eq.~\eqref{eq:err_model}, the negative log-likelihood
(NLL) leads to a GLS objective. For a 2D spatial domain, define the residual
\begin{align*}
r_{i,j,s}
=
u_{\text{data}}^{i,j,s}
-
u(x_{1,i}, x_{2,j}, t_s).
\end{align*}
By Eq.~\eqref{eq:err_model},
$r_{i,j,s} = \omega_{i,j,s} \, u^{\gamma}(x_{1,i}, x_{2,j}, t_s)\,
\varepsilon_{i,j,s}$ with
$\varepsilon_{i,j,s} \overset{\text{iid}}{\sim} \mathcal{N}(0,1)$, so the
residuals are independent, zero-mean Gaussian variables with variance
\begin{align*}
\mathrm{Var}(r_{i,j,s})
=
\left(
\omega_{i,j,s}
\, u^{\gamma}(x_{1,i}, x_{2,j}, t_s)
\right)^2.
\end{align*}
The NLL of the observations is therefore
\begin{align*}
\frac{1}{2}
\sum_{i=1}^{n_{x_1}}
\sum_{j=1}^{n_{x_2}}
\sum_{s=1}^{n_t}
\left[
\frac{r_{i,j,s}^2}{\mathrm{Var}(r_{i,j,s})}
+
\log \mathrm{Var}(r_{i,j,s})
\right]
+
\frac{n_{x_1} n_{x_2} n_t}{2} \log (2\pi).
\end{align*}
Following \textcite{Lagergren2020}, we retain only the
inverse-variance-weighted residual term and drop the log-variance and constant contributions, 
giving the weighted least-squares objective
\begin{align*}
\sum_{i=1}^{n_{x_1}}
\sum_{j=1}^{n_{x_2}}
\sum_{s=1}^{n_t}
\frac{r_{i,j,s}^2}{\mathrm{Var}(r_{i,j,s})}.
\end{align*}
Here, we approximate $u$ with $\hat{u}$. Furthermore, the coefficients $\omega_{i,j,s}$ are often assumed to be
constant (or slowly varying), and we adopt this constant-$\omega$ assumption
here, so that they reduce to a uniform multiplicative factor, which we absorb into $\lambda_{\text{data}}$. Therefore, we approximate the noise scaling
using only the density-dependent factor $\hat{u}^{\gamma}$ and normalise by the number of data points. This yields the
GLS-inspired loss of \textcite{Lagergren2020},
\begin{align*}
\mathcal{L}_{\text{GLS}}(\hat{u})
=
\frac{1}{n_{x_1} n_{x_2} n_t}
\sum_{i=1}^{n_{x_1}}
\sum_{j=1}^{n_{x_2}}
\sum_{s=1}^{n_t}
\left(
\frac{
\tilde{u}_{\text{data}}^{i,j,s}
-
\hat{u}(\tilde{x}_{1,i}, \tilde{x}_{2,j}, \tilde{t}_s)
}{
\hat{u}^{\gamma}(\tilde{x}_{1,i}, \tilde{x}_{2,j}, \tilde{t}_s)
}
\right)^2,
\end{align*}
where the density measurements are normalised as
$
\tilde{u}_{\text{data}}^{i,j,s}
=
U^{-1} \, u_{\text{data}}^{i,j,s}
$. We nevertheless adopt a standard LS loss, i.e.\ $\mathcal{L}_{\text{GLS}}$ with
$\gamma = 0$, since each ingredient of the GLS weighting is problematic here.
Regions of very low density are common in biological data, so the weights
$\hat{u}^{-2\gamma}$ can become arbitrarily large in practice and require clipping; $\gamma$ is
 difficult to estimate; and since the weights depend on $\hat{u}$,
dropping the log-variance term allows the objective to be reduced by inflating $\hat{u}^{\gamma}$ instead of fitting the data. The non-dimensionalised LS (or MSE) data loss that we use is
\begin{equation*}
\mathcal{L}_{\text{data}}(\hat{u})
=
\frac{1}{n_{x_1} n_{x_2} n_t}
\sum_{i=1}^{n_{x_1}}
\sum_{j=1}^{n_{x_2}}
\sum_{s=1}^{n_t}
\left(
\tilde{u}_{\text{data}}^{i,j,s}
-
\hat{u}(\tilde{x}_{1,i}, \tilde{x}_{2,j}, \tilde{t}_s)
\right)^2.
\end{equation*}
See Section~\ref{sec:SM2} for an empirical comparison of
$\mathcal{L}_{\text{GLS}}$ and $\mathcal{L}_{\text{data}}$.
\subsubsection{PDE loss}

\paragraph{Collocation points.}
The PDE residual loss is enforced at $n_c$ collocation points obtained by
sampling uniformly over the spatiotemporal domain delimited by the coordinate-wise
extrema of the data. Carrying out this sampling in the dimensional coordinates yields
\begin{align*}
\mathcal{C}
=
\{(x_{1,k}^{(c)}, x_{2,k}^{(c)}, t_{k}^{(c)})\}_{k=1}^{n_c}.
\end{align*}
The same procedure applied to the non-dimensionalised domain
$[\tilde{x}_{1,\min},\tilde{x}_{1,\max}]\times[\tilde{x}_{2,\min},\tilde{x}_{2,\max}]\times[\tilde{t}_{\min},\tilde{t}_{\max}]$ yields
\begin{align*}
\tilde{\mathcal{C}}
=
\{(\tilde{x}_{1,k}^{(c)}, \tilde{x}_{2,k}^{(c)}, \tilde{t}_{k}^{(c)})\}_{k=1}^{n_c}.
\end{align*}

\paragraph{Loss function.}%
The PDE loss is defined as the mean-squared residual evaluated at
collocation points $\tilde{\mathcal{C}}$,
\begin{align*}
\mathcal{L}_{\text{PDE}}(\hat{u}, \hat{D}, \hat{G})
&=
\frac{1}{n_c}
\sum_{k=1}^{n_c}
\left[
\left.\frac{\partial \hat{u}}{\partial \tilde{t}}\right|_{k}
-
\left.\left(\boldsymbol{\alpha}_{\text{PDE},1}\circ
\tilde{\nabla}\right)\!\cdot\!\left(\hat{D}(\hat{u})\,\tilde{\nabla} \hat{u}\right)\right|_{k}
-
\alpha_{\text{PDE},2}\,\hat{G}_k\,\hat{u}_k
\right]^2,
\end{align*}
where $\hat{u}_k = \hat{u}(\tilde{x}_{1,k}^{(c)}, \tilde{x}_{2,k}^{(c)}, \tilde{t}_k^{(c)})$, $\hat{D}_k = \hat{D}(\hat{u}_k)$, and $\hat{G}_k = \hat{G}(\hat{u}_k)$. The notation $\left.f\right|_{k}$ denotes evaluation of a function $f$ and its derivatives at the $k$-th collocation point. 
The coefficients restore the physical magnitudes of the non-dimensional
diffusion and growth surrogates and are given by
\begin{align*}
\boldsymbol{\alpha}_{\text{PDE},1} = d_{\text{scale}}\,T\,\mathbf{L}^{-2},
\qquad
\alpha_{\text{PDE},2} = T\,g_{\text{scale}}.
\end{align*}
When the characteristic length scales are equal,
$\mathrm{L}_1,\dots,\mathrm{L}_d=\mathrm{L}$, this reduces to the scalar coefficient
$\alpha_{\text{PDE},1}=d_{\text{scale}}\,T/\mathrm{L}^{2}$. \\[1em]
In our setting, we take $\mathrm{L},T=\mathcal{O}(1)$, which absorbs these scales into 
the derivative operators and allows us to evaluate the residual in the original coordinates with 
$\alpha_{\text{PDE},1}=d_{\text{scale}}$ and $\alpha_{\text{PDE},2}=g_{\text{scale}}$. Explicitly, evaluating the residual at collocation points $\mathcal{C}$, we use
\begin{align*}
\mathcal{L}_{\text{PDE}}(\hat{u}, \hat{D}, \hat{G})
&=
\frac{1}{n_c}
\sum_{k=1}^{n_c}
\left[
\left.\frac{\partial \hat{u}}{\partial t}\right|_{k}
-
d_{\text{scale}}\,
\left.\nabla\!\cdot\!\left(\hat{D}(\hat{u})\,\nabla \hat{u}\right)\right|_{k}
-
g_{\text{scale}}\,\hat{G}_k\,\hat{u}_k
\right]^2,
\end{align*}
where
$\hat{u}_k = \hat{u}(x_{1,k}^{(c)}, x_{2,k}^{(c)}, t_k^{(c)})$.
In the implementation we have $d_{\text{scale}} \equiv d_{\max}$ and $g_{\text{scale}} \equiv g_{\max}$.

\subsubsection{Biological loss}
\paragraph{Biological constraint points.} 
The biologically motivated penalty is imposed on a fixed equally spaced
grid of non-dimensionalised density values
$
\tilde{u}_q \in [\tilde{u}_{\min}, \tilde{u}_{\max}],\; q=1,\ldots,n_{\text{bio}},
$
where $\tilde{u}_{\min} = \min_{i,j,s} \{u^{i,j,s}_{\text{data}}/U\}$ and $\tilde{u}_{\max} = \max_{i,j,s} \{u^{i,j,s}_{\text{data}}/U\}$ denote the minimum and maximum (non-dimensional) density values observed in the data.
\paragraph{Loss function.}
To penalise violations of the prescribed bounds, of monotone increase of
$\hat{D}$, and of monotone decrease of $\hat{G}$, the biological loss is given by
\begin{align*}
\mathcal{L}_{\text{bio}}(\hat{D},\hat{G})
&=
\sum_{q=1}^{n_{\text{bio}}}
\Bigg[
\lambda_{D_{\lim}}
\Big(
\mathbb{I}_{\{\hat{D}_q<\tilde{d}_{\min}\}}(\hat{D}_q-\tilde{d}_{\min})^2
+
\mathbb{I}_{\{\hat{D}_q>\tilde{d}_{\max}\}}(\hat{D}_q-\tilde{d}_{\max})^2
\Big)
 \\[3pt]
&\quad
+
\lambda_{G_{\lim}}
\Big(
\mathbb{I}_{\{\hat{G}_q<\tilde{g}_{\min}\}}(\hat{G}_q-\tilde{g}_{\min})^2
+
\mathbb{I}_{\{\hat{G}_q>\tilde{g}_{\max}\}}(\hat{G}_q-\tilde{g}_{\max})^2
\Big)
 \\[3pt]
&\quad
+
\lambda_{\partial D}
\mathbb{I}_{\left\{
\left.\frac{\partial \hat{D}}{\partial \tilde{u}}\right|_{\tilde{u}_q}<0
\right\}}
\left(
\left.\frac{\partial \hat{D}}{\partial \tilde{u}}\right|_{\tilde{u}_q}
\right)^2
+
\lambda_{\partial G}
\mathbb{I}_{\left\{
\left.\frac{\partial \hat{G}}{\partial \tilde{u}}\right|_{\tilde{u}_q}>0
\right\}}
\left(
\left.\frac{\partial \hat{G}}{\partial \tilde{u}}\right|_{\tilde{u}_q}
\right)^2
\Bigg],
\end{align*}
where
$\hat{D}_q = \hat{D}(\tilde{u}_q)$,
$\hat{G}_q = \hat{G}(\tilde{u}_q)$,
the indicator function $\mathbb{I}_{\{\text{condition}\}}$ equals $1$ if the condition
holds and $0$ otherwise, and the non-dimensionalised bounds are
$\tilde{d}_{\min}=d_{\min}/d_{\text{scale}}$,
$\tilde{d}_{\max}=d_{\max}/d_{\text{scale}}$,
$\tilde{g}_{\min}=g_{\min}/g_{\text{scale}}$, and
$\tilde{g}_{\max}=g_{\max}/g_{\text{scale}}$.
The coefficients $\lambda_{D_{\lim}}$ and $\lambda_{G_{\lim}}$ weight the penalties
enforcing the prescribed bounds on $\hat{D}$ and $\hat{G}$, respectively, while
$\lambda_{\partial D}$ and $\lambda_{\partial G}$ weight the corresponding
derivative-constraint penalties.

\subsection{Network design}
To describe the multilayer perceptrons (MLPs) used here in an accessible manner, we first present a generic $\ell$-layer architecture consisting of $\ell-1$ hidden layers followed by a single output layer. 
\subsubsection{Generic MLP}
Assume the network processes a minibatch of inputs $\mathbf{X}\in\mathbb{R}^{n_b\times n^{(0)}}$, where \(n_b\) is the batch
size and \(n^{(0)}\) the input dimension. Take the
 widths of the hidden layers to be \(n^{(1)},\dots,n^{(\ell-1)}\) and the
output width as \(n^{(\ell)}\). 
Each layer \(l=1,\dots,\ell\) is characterised by a weight matrix
\(\boldsymbol{\Phi}^{(l)}\in\mathbb{R}^{\,n^{(l-1)}\times n^{(l)}}\) and a bias vector
\(\mathbf{b}^{(l)}\in\mathbb{R}^{\,n^{(l)}}\).
Setting \(\mathbf{H}^{(0)}=\mathbf{X}\), the pre-activations and
activations are computed recursively as
\[
\mathbf{Z}^{(l)} = \mathbf{H}^{(l-1)}\boldsymbol{\Phi}^{(l)} 
                  + \mathbf{1}_{n_b}\mathbf{b}^{(l)\!\top},
\qquad l = 1,\dots,\ell,
\]
\[
\mathbf{H}^{(l)} =
\begin{cases}
\sigma^{(l)}\!\bigl(\mathbf{Z}^{(l)}\bigr), & l = 1,\dots,\ell-1,\\[6pt]
\kappa\,\sigma^{(\ell)}\!\bigl(\mathbf{Z}^{(\ell)}\bigr), & l = \ell,
\end{cases}
\]
where \(\sigma^{(l)}\) is the element-wise activation function at layer \(l\),
\(\mathbf{1}_{n_b}\) is the column vector of ones of length \(n_b\), and
\(\kappa>0\) is a scalar that rescales the network's final output.
Throughout we set \(\kappa=1\), so no output rescaling is applied within
the networks. 
If \(\sigma^{(\ell)}\) is chosen as the identity,
\(\mathbf{H}^{(\ell)}=\kappa\,\mathbf{Z}^{(\ell)}\).
In the expanded forms below we abbreviate \(\mathbf{1}_{n_b}\mathbf{b}^{(l)\!\top}\) to \(\mathbf{b}^{(l)}\) for ease of notation; in each case the bias vector is understood to be broadcast over the \(n_b\) rows of the preceding matrix product.
\subsubsection{Density MLP}
The cell density field is represented by an MLP $\mathrm{NN}_{u}$, consisting of three hidden layers and one output layer ($\ell=4$).  
Furthermore, each hidden layer within a given MLP contains the same number of neurons denoted by  $W_{u}$, while the output layer consists of a single neuron.
\begin{align}
    &\mathrm{NN}_{u}(\mathbf{X}) =
    \sigma_{\text{soft}} \Big(
    \sigma_{\text{hidden}} \big(
    \sigma_{\text{hidden}} \big(
    \sigma_{\text{hidden}} (\mathbf{X} \boldsymbol{\Phi}_{u}^{(1)} + \mathbf{b}_{u}^{(1)})
    \boldsymbol{\Phi}_{u}^{(2)} + \mathbf{b}_{u}^{(2)}
    \big)
    \boldsymbol{\Phi}_{u}^{(3)} + \mathbf{b}_{u}^{(3)}
    \big)
    \boldsymbol{\Phi}_{u}^{(4)} + \mathbf{b}_{u}^{(4)}
    \Big),
    \nonumber
    \\[.5em]
    &\boldsymbol{\Phi}_{u}^{(1)} \in \mathbb{R}^{(d+1) \times W_{u}},
    \ \
    \boldsymbol{\Phi}_{u}^{(2)},\boldsymbol{\Phi}_{u}^{(3)} \in \mathbb{R}^{W_{u} \times W_{u}},
    \ \
    \boldsymbol{\Phi}_{u}^{(4)} \in \mathbb{R}^{W_{u}  \times 1}; 
    \quad 
    \mathbf{b}_{u}^{(l)} \in \mathbb{R}^{W_{u}},
    \ \ l=1,2,3,
    \ \
    \mathbf{b}_{u}^{(4)} \in \mathbb{R}^{1}. \nonumber
\end{align}
The expanded form is written here for the three hidden layers ($L_u =3$), but it extends straightforwardly to any number of hidden layers by inserting additional $\sigma_{\text{hidden}}(\,\cdot\,\boldsymbol{\Phi}_{u}^{(l)} + \mathbf{b}_{u}^{(l)})$ compositions.
The input comprises the $d$ spatial coordinates and time,
$\mathbf{X} \in \mathbb{R}^{n_b \times (d+1)}$.
All hidden layers share a common hidden activation $\sigma_{\text{hidden}}$ (default $\sigma_{\text{hidden}} = \texttt{SiLU} = \tfrac{x}{1+e^{-x}}$), while the output layer uses the softplus activation
$\sigma_{\text{soft}}(x)=\ln(1+e^x)$
to ensure non-negative cell density predictions. 
\subsubsection{Diffusion and growth MLPs}
The diffusion and growth coefficient functions are also represented by MLPs,
$\mathrm{NN}_{D}$ and $\mathrm{NN}_{G}$, which take $\mathrm{NN}_{u}(\mathbf{X})$ as their input and each return a scalar output.
As for $\mathrm{NN}_{u}$, each hidden layer within a given network has identical width, denoted by $W_D$ and $W_G$, respectively.
Each output layer consists of a single neuron.
\paragraph{Diffusion network}
\begin{align}
    &\mathrm{NN}_{D}(\mathrm{NN}_{u}(\mathbf{X})) =
    \sigma_{\text{soft}} \Big(
    \sigma_{\text{hidden}} \big(
    \sigma_{\text{hidden}} \big(
    \sigma_{\text{hidden}} (\mathrm{NN}_{u}(\mathbf{X}) \boldsymbol{\Phi}_{D}^{(1)} + \mathbf{b}_{D}^{(1)})
    \boldsymbol{\Phi}_{D}^{(2)} + \mathbf{b}_{D}^{(2)}
    \big) \nonumber \\
    &\qquad\qquad
    \boldsymbol{\Phi}_{D}^{(3)} + \mathbf{b}_{D}^{(3)}
    \big)
    \boldsymbol{\Phi}_{D}^{(4)} + \mathbf{b}_{D}^{(4)}
    \Big),
    \nonumber
    \\[.5em]
    &\boldsymbol{\Phi}_{D}^{(1)} \in \mathbb{R}^{1 \times W_D},
    \ \
    \boldsymbol{\Phi}_{D}^{(2)},\boldsymbol{\Phi}_{D}^{(3)} \in \mathbb{R}^{W_D \times W_D},
    \ \
    \boldsymbol{\Phi}_{D}^{(4)} \in \mathbb{R}^{W_D \times 1};
    \quad
    \mathbf{b}_{D}^{(l)} \in \mathbb{R}^{W_D},
    \ \ l=1,2,3,
    \ \
    \mathbf{b}_{D}^{(4)} \in \mathbb{R}^{1}. \nonumber
\end{align}
The softplus activation in the output layer enforces non-negative diffusivity. The expanded form is written here for the three hidden layers ($L_D =3$), but it extends straightforwardly to any number of hidden layers by inserting additional $\sigma_{\text{hidden}}(\,\cdot\,\boldsymbol{\Phi}_{D}^{(l)} + \mathbf{b}_{D}^{(l)})$ compositions.
\paragraph{Growth network}
\begin{align}
    &\mathrm{NN}_{G}(\mathrm{NN}_{u}(\mathbf{X})) =
    \sigma_{\text{hidden}} \Big(
    \sigma_{\text{hidden}} \big(
    \sigma_{\text{hidden}} (\mathrm{NN}_{u}(\mathbf{X}) \boldsymbol{\Phi}_{G}^{(1)} + \mathbf{b}_{G}^{(1)})
    \boldsymbol{\Phi}_{G}^{(2)} + \mathbf{b}_{G}^{(2)}
    \big)
    \boldsymbol{\Phi}_{G}^{(3)} + \mathbf{b}_{G}^{(3)}
    \Big)
    \boldsymbol{\Phi}_{G}^{(4)} + \mathbf{b}_{G}^{(4)},
    \nonumber
    \\[.5em]
    &\boldsymbol{\Phi}_{G}^{(1)} \in \mathbb{R}^{1 \times W_G},
    \quad
    \boldsymbol{\Phi}_{G}^{(2)},\boldsymbol{\Phi}_{G}^{(3)} \in \mathbb{R}^{W_G \times W_G},
    \quad
    \boldsymbol{\Phi}_{G}^{(4)} \in \mathbb{R}^{W_G \times 1};
    \quad
    \mathbf{b}_{G}^{(l)} \in \mathbb{R}^{W_G},
    \ \ l=1,2,3,
    \ \
    \mathbf{b}_{G}^{(4)} \in \mathbb{R}^{1}. \nonumber
\end{align}
No non-negativity constraint is imposed on the growth term, and the output layer therefore uses the identity activation. The expanded form is written here for three hidden layers ($L_G =3$), but it extends straightforwardly to any number of hidden layers by inserting additional $\sigma_{\text{hidden}}(\,\cdot\,\boldsymbol{\Phi}_{G}^{(l)} + \mathbf{b}_{G}^{(l)})$ compositions.
\subsection{Early-stopping algorithm} \label{sec:SM:ES}

Here, the primary hyperparameters are the early-stopping patience ($ES$) and the 
widths of the density, diffusion, and growth MLPs, denoted by $W_u$, $W_D$, 
and $W_G$, respectively. Without loss of 
generality, we define the BINN configuration with respect to these parameters
$
\mathcal{A} = (W_u, W_D, W_G, ES).
$
The complete early-stopping procedure is given in Algorithm~\ref{alg:early_stopping}.

\newcommand{\LossHist}{\mathsf{L}}
\newcommand{\DiffHist}{\mathsf{D}}
\newcommand{\GrowHist}{\mathsf{G}}
\SetKwInput{KwInit}{Initialisation}

\begin{algorithm}[H]
\caption{Validation-loss-based early stopping for experimental and
simulated data, with optional recording of diffusion and growth MSEs}
\label{alg:early_stopping}
\DontPrintSemicolon

\KwIn{
BINN configuration
$\mathcal{A}=(W_u,W_D,W_G,ES)$;
training data $\mathcal{T}$ and validation data $\mathcal{V}$;
minibatch size $n_b$;
density grid
$\{\tilde{u}_q\}_{q=1}^{n_{\mathrm{MSE}}}
\subseteq
[\tilde{u}_{\min},\tilde{u}_{\max}]$;
characteristic diffusion and growth scales
$d_{\mathrm{scale}}$ and $g_{\mathrm{scale}}$;
large maximum epoch cap $e_{\max}$;
$\mathrm{mode}\in
\{\mathrm{simulated},\mathrm{experimental}\}$;
relative update threshold $\delta_{\mathrm{upd}}$
(default $0.05$);
validation frequency $\nu$ (default $1$);
and, for simulated data only, ground-truth diffusion
$D_{\mathrm{true}}$, ground-truth growth
$G_{\mathrm{true}}$, MSE sampling interval $\rho=10$,
and number of MSE points $n_{\mathrm{MSE}}=20$.
}

\KwInit{
$\mathcal{L}_{\mathrm{val}}^\star
\leftarrow +\infty$
\tcp*[r]{Best validation loss for updates}

$e^\star \leftarrow 0$
\tcp*[r]{Epoch of most recent qualifying update}

$e_{\mathrm{sav}} \leftarrow 0$
\tcp*[r]{Epoch of most recently saved checkpoint}

$\LossHist \leftarrow \emptyset$
\tcp*[r]{Validation-loss history}

$\DiffHist \leftarrow \emptyset$
\tcp*[r]{Diffusion-MSE history}

$\GrowHist \leftarrow \emptyset$
\tcp*[r]{Growth-MSE history}
}

\For{$e \leftarrow 0$ \KwTo $e_{\max}-1$}{

Shuffle $\mathcal{T}$ and divide it into
$
N_{\mathrm{tr,batch}}
=
\left\lceil
\frac{|\mathcal{T}|}{n_b}
\right\rceil
$
training minibatches, each containing at most $n_b$ observations;

\ForEach{training minibatch
$\mathcal{B}_{\mathrm{tr}}\subseteq\mathcal{T}$}{

Compute the total training loss on
$\mathcal{B}_{\mathrm{tr}}$;

Backpropagate the total loss and update the model parameters
using the optimiser;
}

\If{$e \bmod \nu = 0$}{

Divide $\mathcal{V}$ into
$
N_{\mathrm{val,batch}}
=
\left\lceil
\frac{|\mathcal{V}|}{n_b}
\right\rceil
$
validation minibatches, each containing at most $n_b$ observations;

\ForEach{validation minibatch
$\mathcal{B}_{\mathrm{val}}\subseteq\mathcal{V}$}{

Compute the validation loss
$\mathcal{L}_{\mathrm{val}}
(\mathcal{B}_{\mathrm{val}};e)$;
}

Average the validation-minibatch losses:
$
\mathcal{L}_{\mathrm{val}}[e]
\leftarrow
\frac{1}{N_{\mathrm{val,batch}}}
\sum_{\mathcal{B}_{\mathrm{val}}}
\mathcal{L}_{\mathrm{val}}
(\mathcal{B}_{\mathrm{val}};e);
$

Append $\mathcal{L}_{\mathrm{val}}[e]$
to $\LossHist$;

$\mathcal{L}_{\mathrm{best,prev}}^\star
\leftarrow
\mathcal{L}_{\mathrm{val}}^\star$;

\If{
$\mathcal{L}_{\mathrm{val}}[e]
<
\mathcal{L}_{\mathrm{best,prev}}^\star
(1-\delta_{\mathrm{upd}})$
}{

$\mathcal{L}_{\mathrm{val}}^\star
\leftarrow
\mathcal{L}_{\mathrm{val}}[e]$;

$e^\star \leftarrow e$;

Save the trained model as
\texttt{configuration\_best};

$e_{\mathrm{sav}} \leftarrow e$;
}
}

\If{
$\mathrm{mode}=\mathrm{simulated}$
\textbf{and}
$e \bmod \rho = 0$
}{

Compute
$
\operatorname{MSE}(\hat{D};e)
=
\frac{1}{n_{\mathrm{MSE}}}
\sum_{q=1}^{n_{\mathrm{MSE}}}
\left(
d_{\mathrm{scale}}\hat{D}(\tilde{u}_q)
-
D_{\mathrm{true}}(\tilde{u}_q)
\right)^2;
$

Compute
$
\operatorname{MSE}(\hat{G};e)
=
\frac{1}{n_{\mathrm{MSE}}}
\sum_{q=1}^{n_{\mathrm{MSE}}}
\left(
g_{\mathrm{scale}}\hat{G}(\tilde{u}_q)
-
G_{\mathrm{true}}(\tilde{u}_q)
\right)^2;
$

Append $\operatorname{MSE}(\hat{D};e)$
to $\DiffHist$;

Append $\operatorname{MSE}(\hat{G};e)$
to $\GrowHist$;
}

\If{$e-e^\star \geq ES$}{
\textbf{break};
}
}

\KwOut{
saved model \texttt{configuration\_best} (referred to as the best model) and its epoch
$e_{\mathrm{sav}}$;
best update epoch $e^\star$;
validation-loss history $\LossHist$;
and, for simulated data only, diffusion- and growth-MSE
histories $\DiffHist$ and $\GrowHist$.
}

\end{algorithm}

\end{document}